\documentclass[aps,a4paper,12pt,tightenlines,nofootinbib,longbibliography,notitlepage]{revtex4-1}

\usepackage{mathptmx}
\usepackage[utf8]{inputenc}
\usepackage[english]{babel}
\usepackage{latexsym}
\usepackage{amsmath}
\usepackage{amsfonts}
\usepackage{graphicx}

\usepackage{lipsum}
\usepackage{colortbl}
\usepackage{xcolor}
\usepackage{booktabs}
\usepackage{rotating}
\usepackage{relsize}

\usepackage{overpic}
\usepackage{hyperref}

\usepackage{setspace}
\usepackage[multiple]{footmisc}

\newcommand{\avg}[1]{{\left<#1\right>}}

\newcommand{\dd}{\mathrm{d}}

\newcommand{\A}{\bm{A}}
\newcommand{\bb}{\bm{b}}
\newcommand{\e}{\bm{e}}
\newcommand{\Ao}{\bm{A^O}}
\newcommand{\dA}{\delta\A}
\newcommand{\G}{\bm{G}}

\usepackage{mathtools}
\usepackage{bm}
\def\multiset#1#2{\ensuremath{\left(\kern-.3em\left(\genfrac{}{}{0pt}{}{#1}{#2}\right)\kern-.3em\right)}}

\renewcommand{\index}[2]{}

\makeatletter 

\begin{document}

\title{Bayesian stochastic blockmodeling\footnote{To appear in
``Advances in Network Clustering and Blockmodeling,'' edited by P. Doreian,
V. Batagelj, A. Ferligoj, (Wiley, New York, 2019 [forthcoming]).}}
\author{Tiago P. Peixoto}
\email{t.peixoto@bath.ac.uk}
\affiliation{Department of Mathematical
Sciences and Centre for Networks and Collective Behaviour, University of
Bath, United Kingdom}
\affiliation{ISI Foundation, Turin, Italy}

\begin{abstract}
  This chapter provides a self-contained introduction to the use of
  Bayesian inference to extract large-scale modular structures from
  network data, based on the stochastic blockmodel (SBM), as well as
  its degree-corrected and overlapping generalizations. We focus on
  nonparametric formulations that allow their inference in a manner that
  prevents overfitting, and enables model selection. We discuss aspects
  of the choice of priors, in particular how to avoid underfitting via
  increased Bayesian hierarchies, and we contrast the task of sampling
  network partitions from the posterior distribution with finding the
  single point estimate that maximizes it, while describing efficient
  algorithms to perform either one. We also show how inferring the SBM
  can be used to predict missing and spurious links, and shed light on
  the fundamental limitations of the detectability of modular structures
  in networks.

\end{abstract}

\maketitle
\newpage
\tableofcontents

\makeatletter
\let\toc@pre\relax
\let\toc@post\relax
\makeatother

\newpage

\section{Introduction}

Since the past decade and a half there has been an ever-increasing
demand to analyze network data, in particular those stemming from
social, biological and technological systems. Often these systems are
very large, comprising millions of even billions of nodes and edges,
such as the World Wide Web, and the global-level social interactions
among humans. A particular challenge that arises is how to describe the
large-scale structures of these systems, in a way that abstracts away
from low-level details, allowing us to focus instead on ``the big
picture.'' Differently from systems that are naturally embedded in some
low-dimensional space
--- such as the population density of cities or the physiology of
organisms ---- we are unable just to ``look'' at a network and readily
extract its most salient features. This has prompted a fury of activity
in developing algorithmic approaches to extract such global information
in a well-defined manner, many of which are described in the remaining
chapters of this book. Most of them operate on a rather simple ansatz,
where we try to divide the network into ``building blocks,'' which then
can be described at an aggregate level in a simplified manner. The
majority of such methods go under the name ``community detection,''
``network clustering'' or ``blockmodeling.'' In this chapter we
consider the situation where the ultimate objective when analyzing
network data in this way is to \emph{model} it, i.e. we want to make
statements about possible generative mechanisms that are responsible for
the network formation. This overall aim sets us in a well-defined path,
where we get to formulate probabilistic models for network structure,
and use principled and robust methods of statistical inference to fit
our models to data. Central to this approach is the ability to
distinguish structure from randomness, so that we do not fool ourselves
into believing that there are elaborate structures in our data which are
in fact just the outcome of stochastic fluctuations --- which tends to
be the Achilles' heel of alternative nonstatistical approaches. In
addition to providing a description of the data, the models we infer can
also be used to generalize from observations, and make statements about
what has \emph{not} yet been observed, yielding something more tangible
than mere interpretations. In what follows we will give an introduction
to this inference approach, which includes recent developments that
allow us to perform it in a consistent, versatile and efficient manner.

\section{Structure versus randomness in networks}\label{sec:noise}

If we observe a random string of characters we will eventually encounter
every possible substring, provided the string is long enough. This leads
to the famous thought experiment of a large number of monkeys with
typewriters: Assuming that they type randomly, for a sufficiently large
number of monkeys any output can be observed, including, for example,
the very text you are reading. Therefore, if we are ever faced with this
situation, we should not be surprised if a such a text is in fact
produced, and most importantly, we should not offer its simian author a
place in a university department, as this occurrence is unlikely to be
repeated. However, this example is of little practical relevance, as the
number of monkeys necessary to type the text ``blockmodeling'' by chance
is already of the order of $10^{18}$, and there are simply not that many
monkeys.

\begin{figure}
  \includegraphics[width=.32\textwidth,trim=.4cm .4cm .3cm 0,clip]{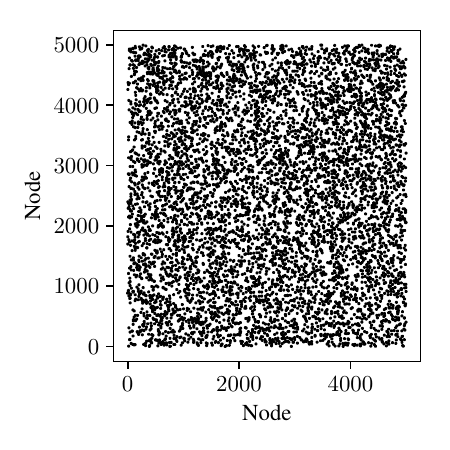}
  \includegraphics[width=.32\textwidth,trim=.4cm .4cm .3cm 0,clip]{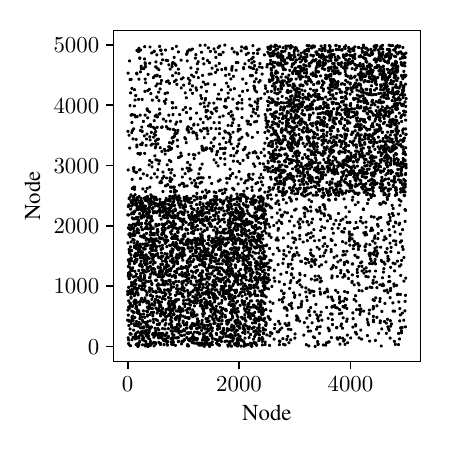}
  \includegraphics[width=.32\textwidth,trim=.4cm .4cm .3cm 0,clip]{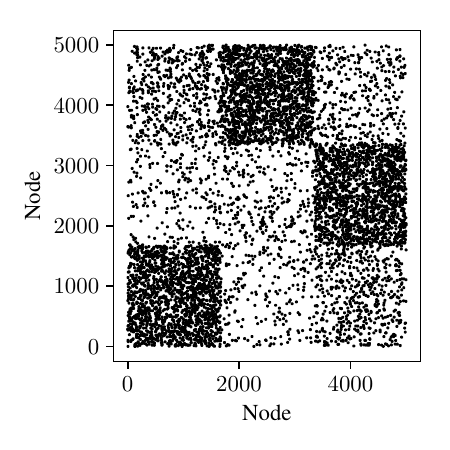}

  \caption{The three panels show the same adjacency matrix, with the
  only difference between them being the ordering of the nodes. The
  different orderings show seemingly clear, albeit very distinct
  patterns of modular structure. However, the adjacency matrix in
  question corresponds to an instance of a fully random Erd\H{o}s-Rényi
  model, where each edge has the same probability $p=\avg{k}/(N-1)$ of
  occurring, with $\avg{k}=3$. Although the patterns seen in the second
  and third panels are not mere fabrications --- as they are really
  there in the network
  --- they are also not meaningful descriptions of this network, since
  they arise purely out of random fluctuations. Therefore, the node
  groups that are identified via these patterns bear no relation to the
  generative process that produced the data.  In other words, the second
  and third panels correspond each to an \emph{overfit} of the data,
  where stochastic fluctuations are misrepresented as underlying
  structure. This pitfall can lead to misleading interpretations of
  results from clustering methods that do not account for statistical
  significance.\label{fig:random}}
\end{figure}

Networks, however, are different from random strings. The network
analogue of a random string is an Erd\H{o}s-Rényi random
graph~\cite{erdos_random_1959}\index{author}{Paul
Erd\H{o}s}\index{author}{Alfréd Rényi} where each possible edge can
occur with the same probability. But differently from a random string, a
random graph can contain a wealth of structure before it becomes
astronomically large --- specially if we \emph{search} for it. An
example of this is shown in Fig.~\ref{fig:random} for a modest network
of $5,000$ nodes, where its adjacency matrix is visualized using three
different node orderings. Two of the orderings seem to reveal patterns
of large-scale connections that are tantalizingly clear, and indeed
would be eagerly captured my many network clustering
methods~\cite{guimera_modularity_2004}. In particular, they seem to show
groupings of nodes that have distinct probabilities of connections to
each other --- in direct contradiction to actual process that generated
the network, where all connections had the same probability of
occurring. What makes matters even worse is that in
Fig.~\ref{fig:random} is shown only a very small subset of all orderings
that have similar patterns, but are otherwise very distinct from each
other. Naturally, in the same way we should not confuse a monkey with a
proper scientist in our previous example, we should not use any of these
node groupings to explain why the network has its structure. Doing so
should be considering \emph{overfitting}\index{topic}{overfitting} it,
i.e. mistaking random fluctuations for generative structure, yielding an
overly complicated and ultimately wrong explanation for the data.

The remedy to this problem is to think probabilistically. We need to
ascribe to each possible explanation of the data a probability that it
is correct, which takes into account modeling assumptions, the
statistical evidence available in the data, as well any source of prior
information we may have. Imbued in the whole procedure must be the
principle of parsimony --- or Occam's razor\index{topic}{Occam's razor}
--- where a simpler model is preferred if the evidence is not sufficient
to justify a more complicated one.

In order to follow this path, before we look at any network data, we
must first look in the ``forward'' direction, and decide on which
mechanisms generate networks in the first place. Based on this, we will
finally be able to look ``backwards,'' and tell which particular
mechanism generated a given observed network.

\section{The stochastic blockmodel (SBM)}\index{topic}{stochastic blockmodel}

\begin{figure}
  \begin{tabular}{cc}
    {\begin{overpic}[width=.48\textwidth]{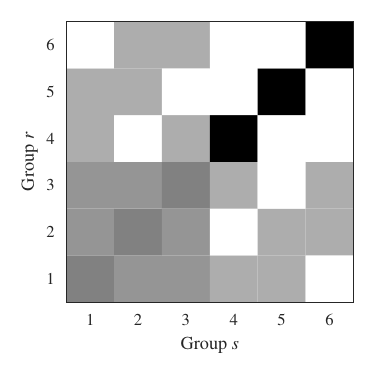}\put(0,0){(a)}\end{overpic}} &
    {\begin{overpic}[width=.48\textwidth]{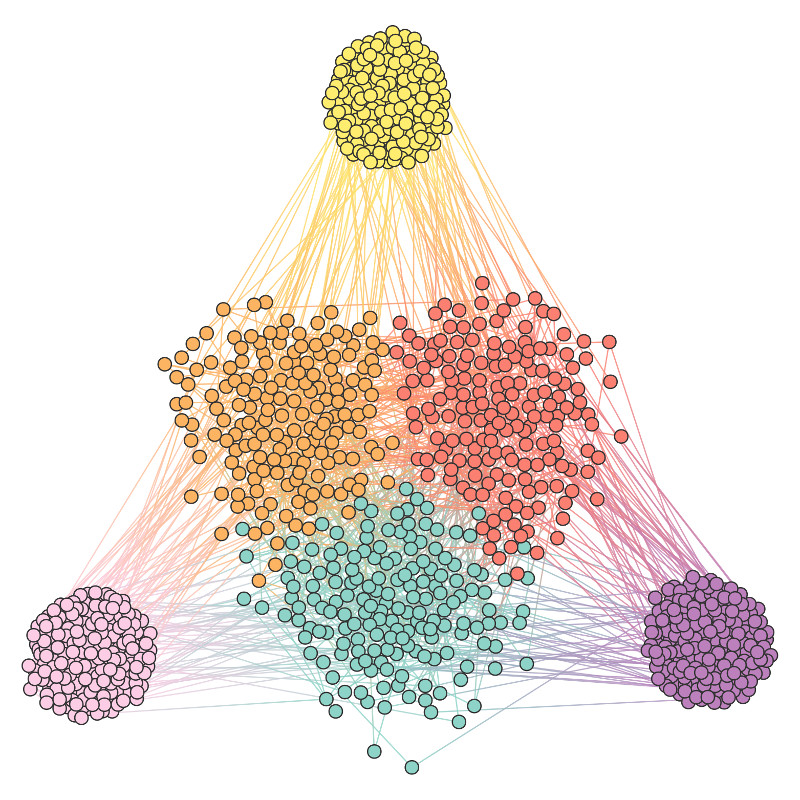}\put(0,0){(b)}\end{overpic}}
  \end{tabular} \caption{The stochastic blockmodel (SBM): (a) The
  matrix of probabilities between groups $p_{rs}$ defines the
  large-scale structure of generated networks; (b) a sampled network
  corresponding to (a), where the node colors indicate the group
  membership.\label{fig:sbm}}
\end{figure}

As mentioned in the introduction, we wish to decompose networks into
``building blocks,'' by grouping together nodes that have a similar role
in the network. From a generative point of view, we wish to work with
models that are based on a partition of $N$ nodes into $B$ such building
blocks, given by the vector $\bb$ with entries
\begin{equation*}
  b_i \in \{1,\dots, B\},
\end{equation*}
specifying the group membership of node $i$. We wish to construct a
generative model that takes this division of the nodes as parameters,
and generates networks with a probability
\begin{equation*}
  P(\A|\bb),
\end{equation*}
where where $\A=\{A_{ij}\}$ is the adjacency matrix. But what shape
should $P(\A|\bb)$ have? If we wish to impose that nodes that belong to
the same group are statistically indistinguishable, our ensemble of
networks should be fully characterized by the number of edges that
connects nodes of two groups $r$ and $s$,
\begin{equation}\label{eq:edge_counts}
  e_{rs} = \sum_{ij}A_{ij}\delta_{b_i,r}\delta_{b_j,s},
\end{equation}
or twice that number if $r=s$. If we take these as conserved
quantities, the ensemble that reflects our maximal indifference towards
any other aspect is the one that maximizes the entropy~\cite{jaynes_probability_2003}
\begin{align}
  S = -\sum_{\A}P(\A|\bb)\ln P(\A|\bb)
\end{align}
subject to the constraint of Eq.~\ref{eq:edge_counts}. If we relax
somewhat our requirements, such that Eq.~\ref{eq:edge_counts} is obeyed
only on expectation, then we can obtain our model using the method of
Lagrange multipliers, using the Lagrangian function
\begin{align}
  F = S - \sum_{r\le s}\mu_{rs}\left(\sum_{\A}P(\A|\bb)\sum_{i<j}A_{ij}\delta_{b_i,r}\delta_{b_j,s} - \avg{e_{rs}}\right) - \lambda\left(\sum_{\A}P(\A|\bb) - 1\right)
\end{align}
where $\avg{e_{rs}}$ are constants
independent of $P(\A|\bb)$, and $\bm{\mu}$ and $\lambda$ are
multipliers that enforce our desired constraints and normalization,
respectively. Obtaining the saddle point $\partial F/\partial
P(\A|\bb)=0$, $\partial F/\partial\mu_{rs}=0$ and $\partial
F/\partial\lambda=0$ gives us the maximum entropy ensemble with the
desired properties. If we constrain ourselves to simple graphs, i.e.
$A_{ij}\in\{0,1\}$, without self-loops,
we have as our maximum entropy model\index{topic}{stochastic blockmodel!Bernoulli}
\begin{equation}\label{eq:bernoulli}
  P(\A|\bm{p},\bb) = \prod_{i<j}p_{b_i,b_j}^{A_{ij}}(1-p_{b_i,b_j})^{1-A_{ij}}.
\end{equation}
with $p_{rs}=e^{-\mu_{rs}}/(1+e^{-\mu_{rs}})$ being the probability of
an edge existing between any two nodes belonging to group $r$ and
$s$. This model is called the {\bfseries stochastic blockmodel} (SBM),
and has its roots in the social sciences and
statistics~\cite{holland_stochastic_1983, wang_stochastic_1987,
snijders_estimation_1997, nowicki_estimation_2001}, but has appeared
repeatedly in the literature under a variety of different
names~\cite{soderberg_general_2002,bollobas_phase_2007,condon_algorithms_2001,
  boguna_class_2003,daudin_mixture_2008,bianconi_assessing_2009}. By
selecting the probabilities $\bm{p}=\{p_{rs}\}$ appropriately, we can
achieve arbitrary mixing patterns between the groups of nodes, as
illustrated in Fig.~\ref{fig:sbm}. We stress that while the SBM can
perfectly accommodate the usual ``community structure''
pattern~\cite{fortunato_community_2010},
i.e. when the diagonal entries of $\bm{p}$ are dominant, it can equally
well describe a large variety of other patterns, such as bipartiteness,
core-periphery, and many others.

Instead of simple graphs, we may consider \emph{multigraphs} by allowing
multiple edges between nodes, i.e. $A_{ij} \in \mathbb{N}$. Repeating
the same procedure, we obtain in this case\index{topic}{stochastic blockmodel!geometric}
\begin{equation}\label{eq:geometric}
  P(\A|\bm{\lambda},\bb) = \prod_{i<j}\frac{\lambda_{b_i, b_j}^{A_{ij}}}{(\lambda_{b_i, b_j}+1)^{A_{ij}+1}},
\end{equation}
with $\lambda_{rs}=e^{-\mu_{rs}}/(1-e^{-\mu_{rs}})$ being the
\emph{average number} of edges existing between any two nodes belonging
to group $r$ and $s$. Whereas the placement of edges in
Eq.~\ref{eq:bernoulli} is given by a Bernoulli distribution, in
Eq.~\ref{eq:geometric} it is given by a geometric distribution,
reflecting the different nature of both kinds of networks. Although
these models are not the same, there is in fact little difference
between the networks they generate in the \emph{sparse limit} given by
$p_{rs}=\lambda_{rs}=O(1/N)$ with $N \gg 1$. We see this by noticing how
their log-probabilities become asymptotically identical in this limit,
i.e.
\begin{align}
  \ln P(\A|\bm{p},\bb) &\approx \frac{1}{2}\sum_{rs}e_{rs}\ln p_{rs} - n_rn_sp_{rs} + O(1),\\
  \ln P(\A|\bm{\lambda},\bb) &\approx \frac{1}{2}\sum_{rs}e_{rs}\ln \lambda_{rs} - n_rn_s\lambda_{rs} + O(1)\label{eq:geom_sparse}.
\end{align}
Therefore, since most networks that we are likely to encounter are
sparse~\cite{newman_networks:_2010}, it does not matter which model we
use, and we may prefer whatever is more convenient for our
calculations. With this in mind, we may consider yet another variant,
which uses instead a Poisson distribution to sample
edges~\cite{karrer_stochastic_2011},\index{topic}{stochastic blockmodel!Poisson}
\begin{equation}\label{eq:poisson-sbm}
  P(\A|\bm{\lambda},\bb) = \prod_{i< j}\frac{e^{-\lambda_{b_i,b_j}}\lambda_{b_i,b_j}^{A_{ij}}}{A_{ij}!}\times\prod_i\frac{e^{-\lambda_{b_i,b_i}/2}(\lambda_{b_i,b_i}/2)^{A_{ii}/2}}{(A_{ii}/2)!},
\end{equation}
where now we also allow for self-loops. Like the geometric model, the
Poisson model generates multigraphs, and it is easy to verify that
it also leads to Eq.~\ref{eq:geom_sparse} in the sparse limit. This
model is easier to use in some of the calculations that we are going to
make, in particular when we consider important extensions of the SBM,
therefore we will focus on it.\footnote{Although the Poisson model is not
strictly a maximum entropy ensemble, the generative process behind it is
easy to justify. We can imagine it as the random placement of exactly
$E$ edges into the $N(N-1)/2$ entries of the matrix $\A$, each with a
probability $q_{ij}$ of attracting an edge, with $\sum_{i<j}q_{ij}=1$,
yielding a multinomial distribution
$P(\A|\bm{q},E)=E!\prod_{i<j}q_{ij}^{A_{ij}}/A_{ij}!$
--- where, differently from Eq.~\ref{eq:poisson-sbm}, the
edge placements are not conditionally independent. But if we now sample
the total number of edges $E$ from a Poisson distribution $P(E|\bar{E})$
with average $\bar{E}$, by exploiting the relationship between the
multinomial and Poisson distributions, we have
$P(\A|\bm{q})=\sum_EP(\A|\bm{q},E)P(E|\bar{E})=\prod_{i<j}e^{-\omega_{ij}}\omega_{ij}^{A_{ij}}/A_{ij}!$,
where $\omega_{ij}=q_{ij}/\bar{E}$, which does amount to conditionally
independent edge placements. Making $q_{ij}=\bar{E}\lambda_{b_i,b_j}$,
and allowing self-loops, we arrive at Eq.~\ref{eq:poisson-sbm}.}

The model above generates undirected networks. It can be very easily
modified to generate directed networks instead, by making $\lambda_{rs}$
an asymmetric matrix, and adjusting the model likelihood
accordingly. The same is true for all model variations that are going to
be used in the following sections. However, for the sake of conciseness
we will focus only on the undirected case. We point out that the
corresponding expressions for the directed case are readily available in
the literature
(e.g. Refs.~\cite{peixoto_entropy_2012,peixoto_model_2015,peixoto_nonparametric_2017}).

Now that we have defined how networks with prescribed modular structure
are generated, we need to develop the reverse procedure, i.e. how to
infer the modular structure from data.

\section{Bayesian inference: the posterior probability of partitions}\index{topic}{Bayesian inference}

Instead of generating networks, our nominal task is to determine which
partition $\bb$ generated an observed network $\A$, assuming this was
done via the SBM. In other words, we want to obtain the probability
$P(\bb|\A)$ that a node partition $\bb$ was responsible for a network
$\A$. By evoking elementary properties of conditional probabilities, we
can write this probability as\index{topic}{Bayesian inference!posterior
distribution}
\begin{equation}\label{eq:b-posterior}
  P(\bb|\A) = \frac{P(\A|\bb)P(\bb)}{P(\A)}
\end{equation}
with
\begin{equation}\label{eq:marginal}
  P(\A|\bb) = \int P(\A|\bm{\lambda},\bb)P(\bm{\lambda}|\bb)\,\dd\bm{\lambda}
\end{equation}
being the \emph{marginal likelihood} integrated over the remaining model
parameters, and\index{topic}{Bayesian inference!evidence}
\begin{equation}\label{eq:evidence}
  P(\A) = \sum_{\bb}P(\A|\bb)P(\bb)
\end{equation}
is called the \emph{evidence}, i.e. the total probability of the data
under the model, which serves as a normalization constant in
Eq.~\ref{eq:b-posterior}. Eq.~\ref{eq:b-posterior} is known as
\emph{\bfseries Bayes' rule}\index{author}{Thomas Bayes}, and far from
being only a simple mathematical step, it encodes how our prior
beliefs\index{topic}{Bayesian inference!prior probabilities} about the
model, i.e. before we observe any data --- in the above represented by
the \emph{prior distributions} $P(\bb)$ and $P(\bm{\lambda}|\bb)$ ---
are affected by the observation, yielding the so-called \emph{posterior
distribution} $P(\bb|\A)$. The overall approach outlined above has been
proposed to the problem of network inference by several
authors~\cite{hastings_community_2006,kemp_learning_2006,rosvall_information-theoretic_2007,
  hofman_bayesian_2008,airoldi_mixed_2008,
  guimera_missing_2009,morup_infinite_2011,reichardt_interplay_2011,morup_bayesian_2012,peixoto_parsimonious_2013,
  schmidt_nonparametric_2013,come_model_2015,peixoto_model_2015,
  yan_bayesian_2016,newman_estimating_2016,peixoto_nonparametric_2017},
with different implementations that vary in some superficial details in
the model specification, approximations used, and in particular in the
choice of priors. Here we will not review or compare all approaches in
detail, but rather focus on the most important aspects, while choosing a
particular path that makes exact calculations possible.

The prior probabilities are a crucial element of the inference
procedure, as they will affect the shape of the posterior distribution,
and ultimately, our inference results. In more traditional scenarios,
the choice of priors would be guided by previous observations of data
that are believed to come from the same model. However, this is not an
applicable scenario when considering networks, which are typically
\emph{singletons}, i.e. they are unique objects, instead of coming from
a population (e.g. there is only one internet, one network of trade
between countries, etc).\footnote{One could argue that most networks
change in time, and hence belong to a time series, thus possibly
allowing priors to be selected from earlier observations of the same
network. This is a potentially useful way to proceed, but also opens a
Pandora's box of dynamical network models, where simplistic notions of
statistical stationarity are likely to be contradicted by data. Some
recent progress has been made on the inference of dynamic
networks~\cite{fu_dynamic_2009,xu_dynamic_2014,peixoto_modelling_2017,
peel_detecting_2015,ghasemian_detectability_2016,zhang_random_2017,corneli_exact_2016,matias_catherine_statistical_2016},
but this field is still in relative infancy.} In the absence of such
empirical prior information, we should try as much as possible to be
guided by well defined principles and reasonable assumptions about our
data, rather than \emph{ad hoc} choices. A central proposition we will
be using is the \emph{principle of maximum indifference} about the model
before we observe any data. This will lead us to so-called
\emph{uninformative} priors,\footnote{The name ``uninformative'' is
something of a misnomer, as it is not really possible for priors to
truly carry ``no information'' to the posterior distribution. In our
context, the term is used simply to refer to \emph{maximum entropy
priors}, conditioned on specific constraints.} that are maximum entropy
distributions that ascribe the same probability to each possible
parameter combination~\cite{jaynes_probability_2003}. These priors have
the property that they do not bias the posterior distribution in any
particular way, and thus let the data ``speak for itself.'' But as we
will see in the following, the naive application of this principle will
lead to adverse effects in many cases, and upon closer inspection we
will often be able to identify aspects of the model that we should not
be agnostic about. Instead, a more meaningful approach will be to
describe higher-order aspects of the model with their own models. This
can be done in a manner that preserves the unbiased nature of our
results, while being able to provide a more faithful representation of
the data.

We begin by choosing the prior for the partition, $\bb$. The most direct
uninformative prior is the ``flat'' distribution where all partitions
into at most $B=N$ groups are equally likely, namely
\begin{equation}\label{eq:flat_b}
  P(\bb) = \frac{1}{\sum_{\bb'}1} = \frac{1}{a_N}
\end{equation}
where $a_N$ are the ordered Bell
numbers~\cite{sloane_-line_2003-1}, given by
\begin{equation}
a_N = \sum_{B=1}^N\left\{ {N \atop B} \right\}B!
\end{equation}
where $\left\{N \atop B\right\}$ are the Stirling numbers of the second
kind~\cite{sloane_-line_2003}, which count the number of ways to
partition a set of size $N$ into $B$ indistinguishable and nonempty
groups (the $B!$ in the above equation recovers the distinguishability
of the groups, which we require). However, upon closer inspection we
often find that such flat distributions are not a good choice. In this
particular case, since there are many more partitions into $B+1$ groups
than there are into $B$ groups (if $B$ is sufficiently smaller than
$N$), Eq.~\ref{eq:flat_b} will typically prefer partitions with a number
of groups that is comparable to the number of nodes. Therefore, this
uniform assumption seems to betray the principle of parsimony that we
stated in the introduction, since it favors large models with many
groups, before we even observe the data.\footnote{Using constant priors
such as Eq.~\ref{eq:flat_b} makes the posterior distribution
proportional to the likelihood. Maximizing such a posterior distribution
is therefore entirely equivalent to a ``non-Bayesian'' maximum
likelihood approach, and nullifies our attempt to prevent overfitting.}
Instead, we may wish to be agnostic about the \emph{number of groups
itself}, by first sampling it from its own uninformative distribution
$P(B)=1/N$, and then sampling the partition conditioned on it
\begin{equation}\label{eq:flat_b_B}
  P(\bb|B) = \frac{1}{\left\{ {N \atop B} \right\}B!},
\end{equation}
since $\left\{ {N \atop B} \right\}B!$ is the number of ways to
partition $N$ nodes into $B$ labelled groups.\footnote{We could have
used simply $P(\bb|B)=1/B^N$, since $B^N$ is the number of partitions of
$N$ nodes into $B$ groups, which are allowed to be empty. However, this
would force us to distinguish between the nominal and the actual number
of groups (discounting empty ones) during
inference~\cite{newman_estimating_2016}, which becomes unnecessary if we
simply forbid empty groups in our prior.} Since $\bb$ is a parameter of
our model, the number of groups $B$ is a called a \emph{hyperparameter},
and its distribution $P(B)$ is called
a \emph{hyperprior}\index{topic}{Bayesian inference!hyperprior}. But
once more, upon closer inspection we can identify further problems: If
we sample from Eq.~\ref{eq:flat_b_B}, most partitions of the nodes will
occupy all the groups approximately equally, i.e. all group sizes will
be the approximately the same. Is this something we want to assume
before observing any data? Instead, we may wish to be agnostic about
this aspect as well, and choose to sample first the distribution of
group sizes $\bm{n}=\{n_r\}$, where $n_r$ is the number of nodes in
group $r$, forbidding empty groups,
\begin{equation}
  P(\bm{n}|B) = {N-1\choose B-1}^{-1},
\end{equation}
since ${N-1\choose B-1}$ is the number of ways to divide $N$ nonzero
counts into $B$ nonempty bins. Given these randomly sampled sizes as a
constraint, we sample the partition with a uniform probability
\begin{equation}
  P(\bb|\bm{n}) = \frac{\prod_rn_r!}{N!}.
\end{equation}
This gives us finally
\begin{equation}\label{eq:partition-prior}
  P(\bb) = P(\bb|\bm{n})P(\bm{n}|B)P(B)=\frac{\prod_rn_r!}{N!}{N-1\choose B-1}^{-1}N^{-1}.
\end{equation}
At this point the reader may wonder if there is any particular reason to
stop here. Certainly we can find some higher-order aspect of the group
sizes $\bm{n}$ that we may wish to be agnostic about, and introduce a
``\emph{hyperhyperprior}'', and so on, indefinitely. The reason why we
should not keep recursively being more and more agnostic about
higher-order aspects of our model is that it brings increasingly
diminishing returns. In this particular case, if we assume that the
individual group sizes are sufficiently large, we obtain asymptotically
\begin{equation}
  \ln P(\bb) \approx -NH(\bm{n}) + O(\ln N)
\end{equation}
where $H(\bm{n})=-\sum_r(n_r/N)\ln(n_r/N)$ is the entropy of the group
size distribution. The value $\ln P(\bb) \to -NH(\bm{n})$ is an
information-theoretical limit that cannot be surpassed, regardless of
how we choose $P(\bm{n}|B)$. Therefore, the most we can optimize by
being more refined is a marginal factor $O(\ln N)$ in the
log-probability, which would amount to little practical difference in
most cases.

In the above, we went from a purely flat uninformative prior
distribution for $\bb$, to a Bayesian hierarchy with three levels, where
we sample first the number of groups, the groups sizes, and then finally
the partition. In each of the levels we used maximum entropy
distributions that are constrained by parameters that are themselves
sampled from their own distributions at a higher level. In doing so, we
removed some intrinsic assumptions about the model (in this case, number
and sizes of groups), thereby postponing any decision on them until we
observe the data. This will be a general strategy we will use for the
remaining model parameters.

Having dealt with $P(\bb)$, this leaves us with the prior for the group
to group connections, $\bm{\lambda}$. A good starting point is an
uninformative prior conditioned on a global average, $\bar{\lambda}$,
which will determine the expected density of the network. For a
continuous variable $x$, the maximum entropy distribution with a
constrained average $\bar x$ is the exponential, $P(x)=e^{-x/\bar
x}/\bar{x}$. Therefore, for $\bm{\lambda}$ we have
\begin{equation}\label{eq:prior-lambda}
  P(\bm{\lambda}|\bb) = \prod_{r\le s}e^{-n_rn_s\lambda_{rs}/(1+\delta_{rs})\bar{\lambda}}n_rn_s/(1+\delta_{rs})\bar{\lambda},
\end{equation}
with $\bar{\lambda}=2E/B(B+1)$ determining the expected total number of
edges,\footnote{More strictly, we should treat $\bar{\lambda}$ just as
another hyperparameter and integrate over its own distribution. But
since this is just a global parameter, not affected by the dimension of
the model, we can get away with setting its value directly from the
data. It means we are pretending we know precisely the density of the
network we are observing, which is not a very strong
assumption. Nevertheless, readers that are uneasy with this procedure
can rest assured that this can be completely amended once we move to
microcanonical models in Sec.~\ref{sec:mdl} (see
footnote~\ref{foot:micro-edges}).} where we have assumed the local
average $\avg{\lambda_{rs}}=\bar{\lambda}(1+\delta_{rs})/n_rn_s$, such
that that the expected number of edges
$e_{rs}=\lambda_{rs}n_rn_s/(1+\delta_{rs})$ will be equal to
$\bar{\lambda}$, irrespective of the group sizes $n_r$ and
$n_s$~\cite{peixoto_nonparametric_2017}. Combining this with
Eq.~\ref{eq:poisson-sbm}, we can compute the integrated marginal
likelihood of Eq.~\ref{eq:marginal} as
\begin{equation}\label{eq:sbm-marginal}
  P(\A|\bb) = \frac{\bar\lambda^E}{(\bar{\lambda}+1)^{E+B(B+1)/2}}\times
  \frac{\prod_{r<s}e_{rs}!\prod_re_{rr}!!}{\prod_r n_r^{e_r}\prod_{i<j}A_{ij}!\prod_iA_{ii}!!}.
\end{equation}
Just as with the node partition, the uninformative assumption of
Eq.~\ref{eq:prior-lambda} also leads to its own problems, but we
postpone dealing with them to Sec.~\ref{sec:underfit}. For now, we have
everything we need to write the posterior distribution, with the
exception of the model evidence $P(\A)$ given by
Eq.~\ref{eq:evidence}. Unfortunately, since it involves a sum over all
possible partitions, it is not tractable to compute the evidence
exactly. However, since it is just a normalization constant, we will not
need to determine it when optimizing or sampling from the posterior, as
we will see in Sec.~\ref{sec:mcmc}. The numerator of
Eq.~\ref{eq:b-posterior}, which is comprised of the terms that we can
compute exactly, already contains all the information we need to proceed
with the inference, and also has a special interpretation, as we will
see in the next section.

The posterior of Eq.~\ref{eq:b-posterior} will put low probabilities on
partitions that are not backed by sufficient statistical evidence in the
network structure, and it will not lead us to spurious partitions such
as those depicted in Fig.~\ref{fig:random}. Inferring partitions from
this posterior amounts to a so-called \emph{nonparametric} approach; not
because it lacks the estimation of parameters, but because the number of
parameters itself, a.k.a. the \emph{order} or \emph{dimension} of the
model, will be inferred as well. More specifically, the number of groups
$B$ itself will be an outcome of the inference procedure, which will be
chosen in order to accommodate the structure in the data, without
overfitting. The precise reasons why the latter is guaranteed might not
be immediately obvious for those unfamiliar with Bayesian inference. In
the following section we will provide an explanation by making a
straightforward connection with information theory. The connection is
based on a different interpretation of our model, which allow us to
introduce some important improvements.

\section{Microcanonical models and the minimum description length principle (MDL)}\label{sec:mdl}
We can re-interpret the integrated marginal likelihood of
Eq.~\ref{eq:sbm-marginal} as the joint likelihood of a
\emph{microcanonical}\index{topic}{stochastic blockmodel!microcanonical} model given by\footnote{Some readers may wonder
  why Eq.~\ref{eq:ensemble-equivalence} should not contain a sum,
  i.e. $P(\A|\bb) = \sum_{\e}P(\A|\e,\bb)P(\e|\bb)$. Indeed, that is the
  proper way to write a marginal likelihood. However, for the
  microcanonical model there is only one element of the sum that
  fulfills the constraint of Eq.~\ref{eq:edge_counts}, and thus yields a
  nonzero probability, making the marginal likelihood identical to the
  joint, as expressed in Eq.~\ref{eq:ensemble-equivalence}. The same is
  true for the partition prior of Eq.~\ref{eq:partition-prior}. We will
  use this fact in our notation throughout, and omit sums when they are
  unnecessary.}
\begin{equation}\label{eq:ensemble-equivalence}\index{topic}{ensemble equivalence}
  P(\A|\bb) = P(\A|\e,\bb)P(\e|\bb),
\end{equation}
where
\begin{align}
  P(\A|\e,\bb) &= \frac{\prod_{r<s}e_{rs}!\prod_re_{rr}!!}{\prod_r n_r^{e_r}\prod_{i<j}A_{ij}!\prod_iA_{ii}!!},\label{eq:micro-sbm}\\
  P(\e|\bb) &= \prod_{r<s}\frac{\bar\lambda^{e_{rs}}}{(\bar\lambda+1)^{e_{rs}+1}}\prod_r\frac{\bar\lambda^{e_{rs}/2}}{(\bar\lambda+1)^{e_{rs}/2+1}}=\frac{\bar\lambda^E}{(\bar{\lambda}+1)^{E+B(B+1)/2}}\label{eq:prior-edges},
\end{align}
and $\e=\{e_{rs}\}$ is the matrix of edge counts between groups.  The
term ``microcanonical'' --- borrowed from statistical physics --- means
that model parameters correspond to ``hard'' constraints that are
\emph{strictly} imposed on the ensemble, as opposed to ``soft''
constraints that are obeyed only on average. In the particular case
above, $P(\A|\e,\bb)$ is the probability of generating a multigraph $\A$
where Eq.~\ref{eq:edge_counts} is always fulfilled,
i.e. the total number of edges between groups $r$ and $s$ is always
exactly $e_{rs}$, without any fluctuation allowed between samples (see
Ref.~\cite{peixoto_nonparametric_2017} for a combinatorial
derivation). This contrasts with the parameter $\lambda_{rs}$ in
Eq.~\ref{eq:poisson-sbm}, which determines only the \emph{average}
number of edges between groups, which fluctuates between samples.
Conversely, the prior for the edge counts $P(\e|\bb)$ is a mixture of
geometric distributions with average $\bar{\lambda}$, which does allow
the edge counts to fluctuate, guaranteeing the overall equivalence. The
fact that Eq.~\ref{eq:ensemble-equivalence} holds is rather remarkable,
since it means that --- at least for the basic priors we used --- these two
kinds of model (``canonical'' and microcanonical) cannot be
distinguished from data, since their marginal likelihoods (and hence the
posterior probability) are identical\footnote{This equivalence occurs
for a variety of Bayesian models. For
  instance, if we flip a coin with a probability $p$ of coming up heads,
  the integrated likelihood under a uniform prior after $N$ trials in
  which $m$ heads were observed is $P(\bm{x})=\int_0^1
  p^m(1-p)^{N-m}\,\dd p=(N-m)!m!/(N+1)!$. This is the same as the
  ``microcanonical'' model $P(\bm{x})=P(\bm{x}|m)P(m)$ with
  $P(\bm{x}|m)={N\choose m}^{-1}$ and $P(m)=1/(N+1)$, i.e. the number of
  heads is sampled from a uniform distribution, and the coin flips are
  sampled randomly among those that have that exact number of heads.}.

With this microcanonical interpretation in mind, we may frame the
posterior probability in an information-theoretical manner as
follows. If a discrete variable $x$ occurs with a probability mass
$P(x)$, the asymptotic amount of information necessary to describe it is
$-\log_2P(x)$ (if we choose bits as the unit of measurement), by using
an optimal lossless coding scheme such as Huffman's
algorithm~\cite{mackay_information_2003}. With this in mind, we may
write the numerator of the posterior distribution in
Eq.~\ref{eq:b-posterior} as
\begin{equation}\label{eq:dl}
  P(\A|\bb)P(\bb) = P(\A|\e,\bb)P(\e,\bb) = 2^{-\Sigma},
\end{equation}
where the quantity
\begin{align}
  \Sigma &= -\log_2P(\A,\e,\bb)\\
  &=-\log_2P(\A|\e,\bb)-\log_2P(\e,\bb)\label{eq:sbm-dl-two-part}
\end{align}
is called the {\bfseries\emph{description length}} of the
data~\cite{rissanen_modeling_1978,grunwald_minimum_2007}. It corresponds
to the asymptotic amount of information necessary to encode the data $\A$
\emph{together} with the model parameters $\e$ and $\bb$. Therefore, if
we find a network partition that maximizes the posterior distribution of
Eq.~\ref{eq:sbm-marginal}, we are also automatically finding one which
minimizes the description length.\footnote{Sometimes the minimum
description length principle (MDL) is
  considered as an alternative method to Bayesian inference. Although it
  is possible to apply MDL in a manner that makes the connection with
  Bayesian inference difficult, as for example with the normalized
  maximum likelihood
  scheme~\cite{shtarkov_universal_1987,grunwald_tutorial_2004}, in its
  more direct and tractable form it is fully equivalent to the Bayesian
  approach~\cite{grunwald_minimum_2007}. Note also that we do not in
  fact require the connection with microcanonical models made here, as
  the description length can be defined directly as $\Sigma =
  -\log_2P(\A,\bb)$, without referring explicitly to internal model
  parameters.}\index{topic}{minimum description length (MDL)} With
this, we can see how the Bayesian approach outlined above prevents
overfitting: As the size of the model increases (via a larger number of
occupied groups), it will constrain itself better to the data, and the
amount of information necessary to describe it when the model is known,
$-\log_2P(\A|\e,\bb)$, will decrease. At the same time, the amount of
information necessary to describe the model itself, $-\log_2P(\e,\bb)$,
will increase as it becomes more complex. Therefore, the latter will
function as a \emph{penalty}\footnote{Some readers may notice the
similarity between Eq.~\ref{eq:sbm-dl-two-part} and other penalty-based
criteria, such as BIC~\cite{schwarz_estimating_1978} and
AIC~\cite{akaike_new_1974}\index{topic}{Bayesian inference!model
selection!BIC}\index{topic}{Bayesian inference!model
selection!AIC}. Although all these criteria share the same overall
interpretation, BIC and AIC rely on specific assumptions about the
asymptotic shape of the model likelihood, which are known to be invalid
for the SBM~\cite{yan_model_2014}, unlike Eq.~\ref{eq:sbm-dl-two-part}
which is exact.} that prevents the model from becoming overly complex,
and the optimal choice will amount to a proper balance between both
terms.\footnote{An important result in information theory states that
compressing random data is asymptotically
impossible~\cite{cover_elements_1991}. This lies at the heart of the
effectiveness of the MDL approach in preventing overfitting, as
incorporating randomness into the model description cannot be used to
reduce the description length.} Among other things, this approach will
allow us to properly estimate the dimension of the model
--- represented by the number of groups $B$ --- in a parsimonious way.

\begin{figure}
  \begin{tabular}{cc}
    {\begin{overpic}[width=.4\textwidth, trim=0 -1cm 0 0, clip]{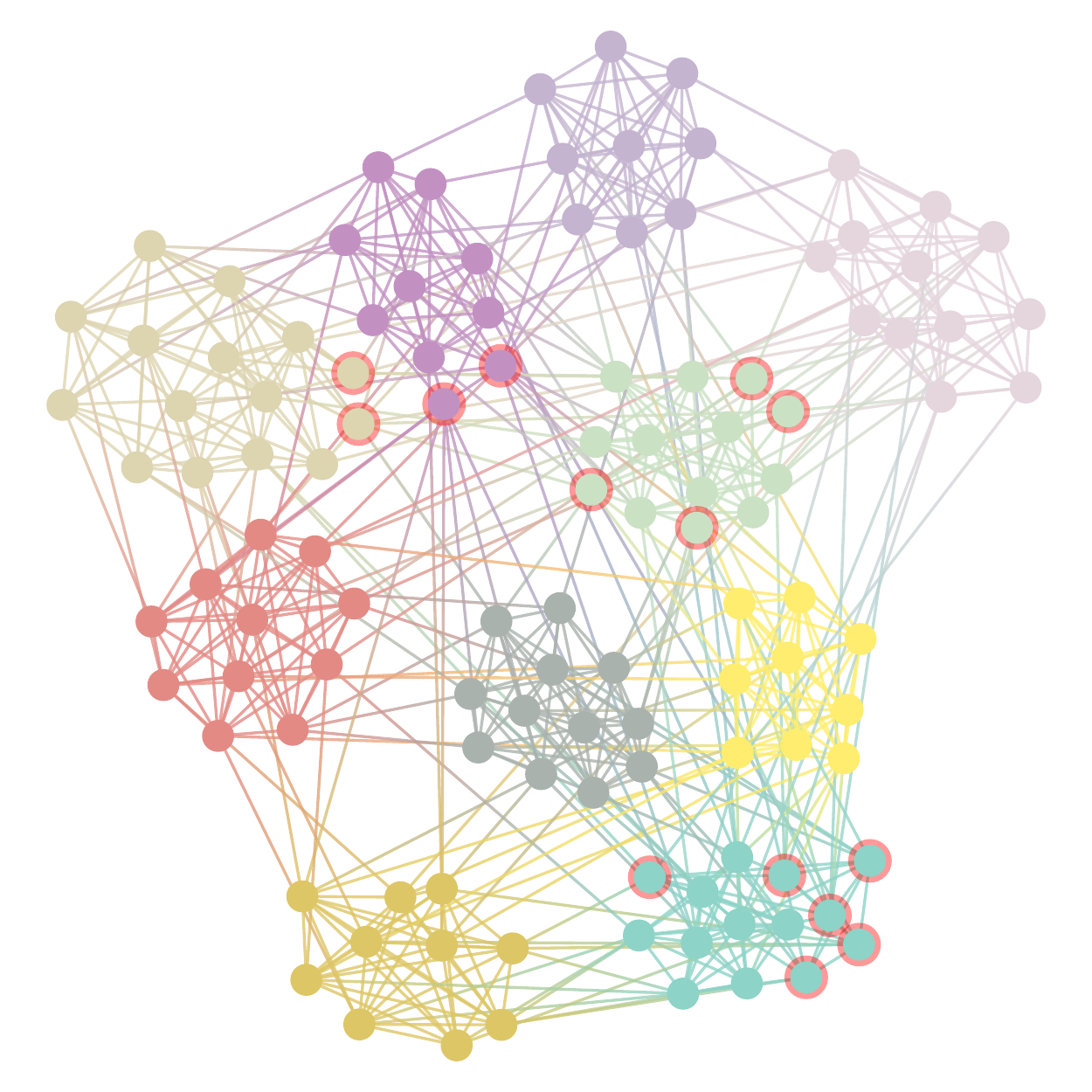}\put(0,0){(a)}\end{overpic}} &
    {\begin{overpic}[width=.55\textwidth,trim=.2cm .3cm .2cm .2cm, clip]{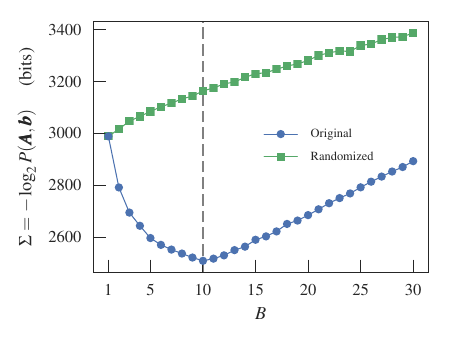}\put(0,0){(b)}\end{overpic}}
  \end{tabular}

  \caption{Bayesian inference of the SBM for a network of American
    college football teams~\cite{girvan_community_2002}: (a) The
    partition that maximizes the posterior probability of
    Eq.~\ref{eq:b-posterior}, or equivalently, minimizes the description
    length of Eq.~\ref{eq:dl}. Nodes marked in red are not classified
    according to the known division into ``conferences.'' (b)
    Description length as a function of the number of groups of the
    corresponding optimal partition, both for the original and
    randomized data.\label{fig:sbm-random}}
\end{figure}

We now illustrate this approach with a real-world dataset of American
college football teams~\cite{girvan_community_2002}, where a node is a
team and an edge exists if two teams played against each other in a
season. If we find the partition that maximizes the posterior
distribution, we uncover $B=10$ groups, as can be seen in
Fig.~\ref{fig:sbm-random}a. If we compare this partition with the known
division of the teams into
``conferences''~\cite{evans_clique_2010,evans_american_2012}, we find
that they match with a high degree of precision, with the exception of
only a few nodes.\footnote{Care should be taken when comparing with
``known'' divisions in this manner, as there is no guarantee that the
available metadata is in fact relevant for the network structure. See
Refs.~\cite{peel_ground_2017,newman_structure_2016,hric_network_2016}
for more detailed discussions.} In Fig.~\ref{fig:sbm-random}b we show
the description length of the optimal partitions if we constrain them to
have a pre-specified number of groups, which allows us to see how the
approach penalizes both too simple and too complex models, with a global
minimum at $B=10$ --- corresponding to the most compressive
partition. Importantly, if we now \emph{randomize} the network, by
placing all its edges in a completely random fashion, we obtain instead
a trivial partition into $B=1$ group --- indicating that the best model
for this data is indeed a fully random graph. Hence, we see that this
approach completely avoids the pitfall discussed in Sec.~\ref{sec:noise}
and does not identify groups in fully random networks, and that the
division shown in Fig.~\ref{fig:sbm-random}a points to a statistically
significant structure in the data, that cannot be explained simply by
random fluctuations.

\section{The ``resolution limit'' underfitting problem, and the nested SBM}\label{sec:underfit}

\begin{figure}[t]
  \begin{tabular}{cc}
    {\begin{overpic}[width=.4\textwidth,trim=0 -1cm 0 0, clip]{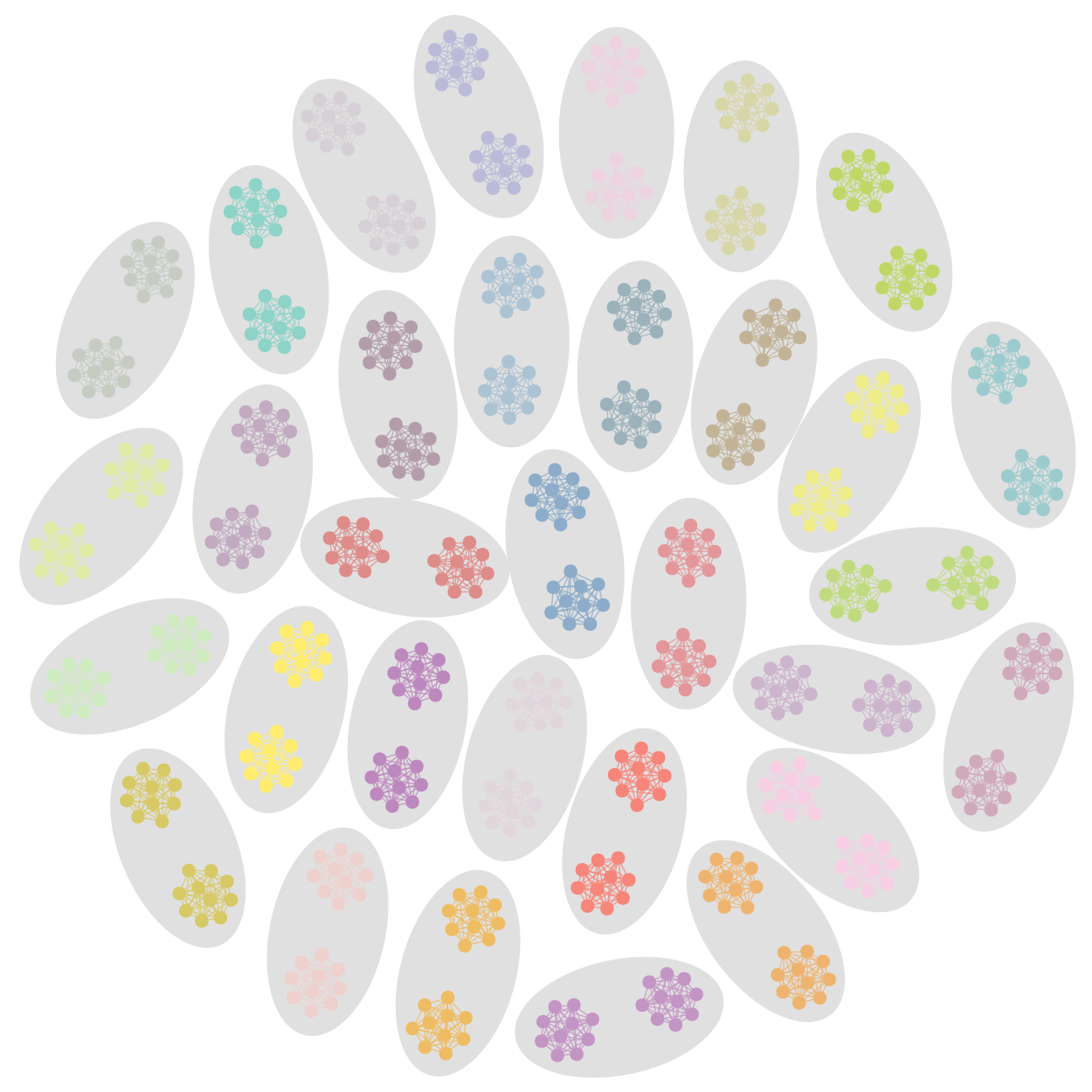}\put(0,0){(a)}\end{overpic}} &
    {\begin{overpic}[width=.55\textwidth,trim=.2cm .3cm .2cm .2cm, clip]{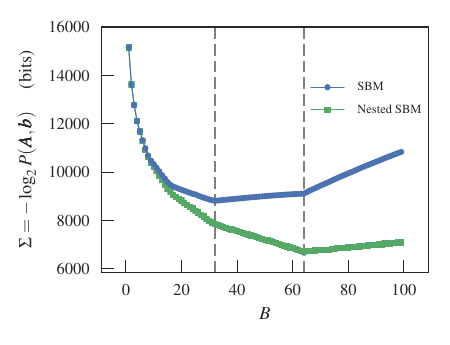}\put(0,0){(b)}\end{overpic}}
  \end{tabular}

  \caption{Inference of the SBM on a simple artificial network composed
  of 64 cliques of size 10, illustrating the underfitting problem: (a)
  The partition that maximizes the posterior probability of
  Eq.~\ref{eq:b-posterior}, or equivalently, minimizes the description
  length of Eq.~\ref{eq:dl}. The 64 cliques are grouped into 32 groups
  composed of two cliques each.
  (b) Minimum description length as a function of the number of groups
  of the corresponding partition, both for the SBM and its nested
  variant, which is less susceptible to underfitting, and puts all 64
  cliques in their own groups.\label{fig:underfit}}
\end{figure}
Although the Bayesian approach outlined above is in general protected
against overfitting, it is still susceptible to
\emph{underfitting},\index{topic}{underfitting} i.e. when we mistake
statistically significant structure for randomness, resulting in the
inference of an overly simplistic model. This happens whenever there is
a large discrepancy between our prior assumptions and what is observed
in the data. We illustrate this problem with a simple example: Consider
a network formed of 64 isolated cliques of size 10, as shown in
Fig.~\ref{fig:underfit}a. If we employ the approach described in the
previous section, and maximize the posterior of
Eq.~\ref{eq:b-posterior}, we obtain a partition into $B=32$ groups,
where each group is composed of two cliques. This is a fairly
unsatisfying characterization of this network, and also somewhat
perplexing, since the probability that the inferred SBM will generate
the observed network
--- i.e. each of the 32 groups will simultaneously and spontaneously
split in two disjoint cliques --- is vanishingly small. Indeed,
intuitively it seems we should do significantly better with this rather
obvious example, and that the best fit would be to put each of the
cliques in their own group. In order to see what went wrong, we need to
revisit our prior assumptions, in particular our choice for
$P(\bm{\lambda}|\bb)$ in Eq.~\ref{eq:prior-lambda}, or equivalently, our
choice of $P(\e|\bb)$ in Eq.~\ref{eq:prior-edges} for the microcanonical
formulation. In both cases, they correspond to uninformative priors,
which put approximately equal weight on all allowed types of large-scale
structures. As argued before, this seems reasonable at first, since we
should not bias our model before we observe the data. However, the
implication of this choice is that we expect \emph{a priori} the
structure of the network at the aggregate group level, i.e. considering
only the groups and the edges between them (not the individual nodes),
to be fully random. This is indeed not the case in the simple example of
Fig.~\ref{fig:underfit}, and in fact it is unlikely to be the case for
most networks that we encounter, which will probably be structured at a
higher level as well. The unfavorable outcome of the uninformative
assumption can also be seen by inspecting its effect on the description
length of Eq.~\ref{eq:dl}. If we revisit our simple model with $C$
cliques of size $m$, grouped uniformly into $B$ groups of size $C/B$,
and we assume that these values are sufficiently large so that
Stirling's factorial approximation $\ln x!\approx x\ln x-x$ can be used,
the description length becomes
\begin{equation}\label{eq:dl_assympt}
  \Sigma \approx -(E-N)\log_2 B + \frac{B(B+1)}{2}\log_2E,
\end{equation}
where $N=Cm$ is the total number of nodes and $E=C{m \choose 2}$ is the
total number of edges, and we have omitted terms that do not depend on
$B$. From this, we see that if we increase the number of groups $B$,
this incurs a quadratic penalty in the description length given by the
second term of Eq.~\ref{eq:dl_assympt}, which originates precisely from
our expression of $P(\e|\bb)$: It corresponds to the amount of
information necessary to describe all entries of a symmetric $B\times B$
matrix that takes independent values between $0$ and $E$. Indeed, a
slightly more careful analysis of the scaling of the description
length~\cite{peixoto_parsimonious_2013,peixoto_nonparametric_2017}
reveals that this approach is unable to uncover a number of groups that
is larger than $B_{\text{max}}\propto\sqrt{N}$, even if their existence
is obvious, as in our example of Fig.~\ref{fig:underfit}.\footnote{This
  same problem occurs for slight variations of the SBM and corresponding
  priors, provided they are uninformative, such as those in
  Refs.~\cite{schmidt_nonparametric_2013,come_model_2015,newman_estimating_2016},
  and also with other penalty based approaches that rely on a functional
  form similar to
  Eq.~\ref{eq:dl_assympt}~\cite{wang_likelihood-based_2017}. Furthermore,
  this limitation is conspicuously similar to the ``resolution limit''
  present in the popular heuristic of modularity
  maximization~\cite{fortunato_resolution_2007}, although it is not yet
  clear if a deeper connection exists between both
  phenomena.}\index{topic}{resolution limit}

Trying to avoid this limitation might seem like a conundrum, since
replacing the uninformative prior for $P(\e|\bb)$ amounts to making a more
definite statement on the most likely large-scale structures that we
expect to find, which we might hesitate to stipulate, as this is
precisely what we want to discover from the data in the first place, and
we want to remain unbiased. Luckily, there is in fact a general approach
available to us to deal with this problem: We postpone our decision
about the higher-order aspects of the model until we observe the
data. In fact, we already saw this approach in action when we decided on
the prior for the partitions; We do so by replacing the uninformative
prior with a \emph{parametric} distribution, whose parameters are in
turn modelled by a another distribution, i.e. a \emph{hyperprior}. The
parameters of the prior then become \emph{latent variables} that are
learned from data, allowing us to uncover further structures, while
remaining unbiased.

\begin{figure}[h!]
  \begin{tabular}{cc}
    {\begin{overpic}[width=.52\textwidth]{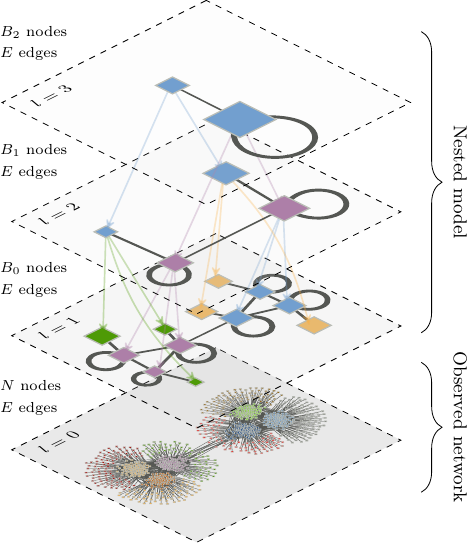}\put(0,0){(a)}\end{overpic}} &
    {\begin{tabular}[b]{c}
      {\begin{overpic}[width=.4\textwidth]{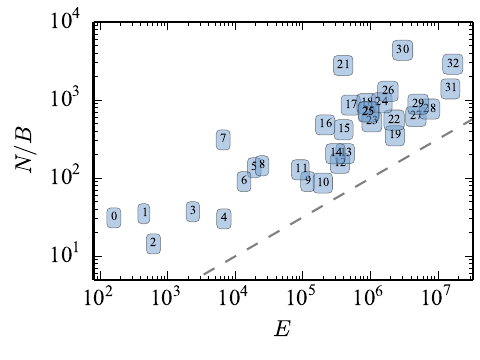}\put(0,0){(b)}\end{overpic}}\\
      {\begin{overpic}[width=.4\textwidth]{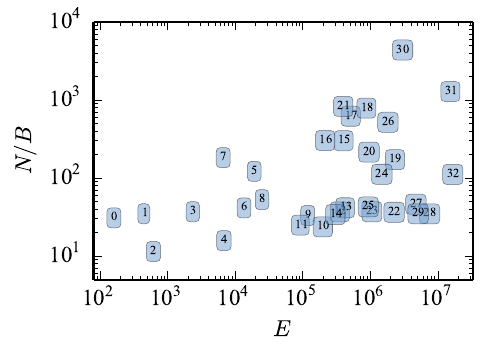}\put(0,0){(c)}\end{overpic}}
    \end{tabular}}
  \end{tabular}

  \caption{\label{fig:nested-sbm}(a) Diagrammatic representation of the
    nested SBM described in the text, with $L=3$ levels, adapted from
    Ref.~\cite{peixoto_hierarchical_2014}. (b) Average group sizes $N/B$
    obtained with the SBM using uninformative priors, for a variety of
    empirical networks, listed in
    Ref.~\cite{peixoto_hierarchical_2014}. The dashed line shows a slope
    $\sqrt{E}$, highlighting the systematic underfitting problem. (c)
    The same as in (b), but using the nested SBM, where the underfitting
    has virtually disappeared, with datasets randomly scattered in the
    allowed range.}
\end{figure}

The microcanonical formulation allows us to proceed in this direction in
a straightforward manner, as we can interpret the matrix of edge counts
$\e$ as the adjacency matrix of a multigraph where each of the groups is
represented as a single node. Within this interpretation, an elegant
solution presents itself, where we describe the matrix $\e$ with
\emph{another} SBM,
i.e. we partition each of the groups into meta-groups, and the edges
between groups are placed according to the edge counts between
meta-groups. For this second SBM, we can proceed in the same manner, and
model it by a third SBM, and so on, forming a nested hierarchy, as
illustrated in
Fig.~\ref{fig:nested-sbm}~\cite{peixoto_hierarchical_2014}\index{topic}{stochastic
blockmodel!nested}. More precisely, if we denote by $B_l$, $\bb_l$ and
$\e_l$ the number of groups, the partition and the matrix of edge counts
at level $l \in \{0,\dots,L\}$, we have
\begin{equation}\label{eq:multi-sbm}
  P(\e_l|\bb_{l-1},\e_{l+1},\bb_l)=\prod_{r<s}\multiset{n^l_rn^l_s}{e_{rs}^{l+1}}^{-1}
  \prod_r\multiset{n^l_r(n^l_r+1)/2}{e_{rs}^{l+1}/2}^{-1},
\end{equation}
with $\multiset{n}{m}={n+m-1\choose m}$ counting the number of
$m$-combinations with repetitions from a set of size
$n$. Eq.~\ref{eq:multi-sbm} is the likelihood of a maximum-entropy
multigraph SBM, i.e. every multigraph occurs with the same probability,
provided they fulfill the imposed constraints\footnote{Note that we
cannot use in the upper levels exactly the same model we use in the
bottom level, given by Eq.~\ref{eq:micro-sbm}, as most terms in the
subsequent levels will cancel out. This happens because the model in
Eq.~\ref{eq:micro-sbm} is based on the uniform generation of
configurations, not
multigraphs~\cite{peixoto_nonparametric_2017}. However, we are free to
use Eq.~\ref{eq:multi-sbm} in the bottom level as
well.}~\cite{peixoto_entropy_2012}. The prior for the partitions is
again given by Eq.~\ref{eq:partition-prior},
\begin{equation}
  P(\bb_l) = \frac{\prod_rn_r^l!}{B_{l-1}!}{B_{l-1}-1\choose B_l-1}^{-1}B_{l-1}^{-1},
\end{equation}
with $B_{-1}=N$, so that the joint probability of the data, edge counts
and the hierarchical partition $\{\bb_l\}$ becomes
\begin{equation}\label{eq:prior-nested}
  P(\A,\{\e_l\},\{\bb_l\}|L)=P(\A|\e_1,\bb_0)P(\bb_0)\prod_{l=1}^{L}P(\e_l|\bb_{l-1},\e_{l+1},\bb_l)P(\bb_l),
\end{equation}
where we impose the boundary conditions $B_L=1$ and $P(\bb_L)=1$. We can
treat the hierarchy depth $L$ as a latent variable as well, by placing a
prior on it $P(L)=1/L_{\text{max}}$, where $L_{\text{max}}$ is the
maximum value allowed. But since this only contributes to an overall
multiplicative constant, it has no effect on the posterior distribution,
and thus can be omitted. If we impose $L=1$, we recover the
uninformative prior for $\e=\e_1$,
\begin{equation}\label{eq:prior-edges-micro}
  P(\e|\bb_0) = \multiset{B(B+1)/2}{E}^{-1},
\end{equation}
which is different from Eq.~\ref{eq:prior-edges} only in that the number
of edges $E$ is not allowed to fluctuate.\footnote{The prior of
Eq.~\ref{eq:prior-edges-micro} and the hierarchy in
Eq.~\ref{eq:prior-nested} are conditioned on the total number of edges
$E$, which is typically unknown before we observe the data. Similarly to
the parameter $\bar\lambda$ in the canonical model formulation, the
strictly correct approach would be to consider this quantity as an
additional model parameter, with its prior distribution $P(E)$. However,
in the microcanonical model there is no integration involved, and $P(E)$
--- regardless of how we specify it --- would contribute to an overall
multiplicative constant that disappears from the posterior distribution
after normalization. Therefore we can simply omit
it.\label{foot:micro-edges}} The inference of this model is done in the
same manner as the uninformative one, by obtaining the posterior
distribution of the hierarchical partition
\begin{equation}
  P(\{\bb_l\}|\A) = \frac{P(\A,\{\bb_l\})}{P(\A)} = \frac{P(\A,\{\e_l\},\{\bb_l\})}{P(\A)},
\end{equation}
and the description length is given analogously by
\begin{equation}
  \Sigma = -\log_2P(\A|\{\e_l\},\{\bb_l\}) -\log_2P(\{\e_l\},\{\bb_l\}).
\end{equation}

\noindent
This approach has a series of advantages; in particular, we remain
\emph{a priori} agnostic with respect to what kind of large-scale
structure is present in the network, having constrained ourselves simply
in that it can be represented as a SBM at a higher level, and with the
uninformative prior as a special case. Despite this, we are able to
overcome the underfitting problem encountered with the uninformative
approach: If we apply this model to the example of
Fig.~\ref{fig:underfit}, we can successfully distinguish all 64 cliques,
and provide a lower overall description length for the data, as can be
seen in Fig.~\ref{fig:underfit}b. More generally, by investigating the
properties of the model likelihood, it is possible to show that the
maximum number of groups that can be uncovered with this model scales as
$B_{\text{max}}\propto N/\log N$, which is significantly larger than the
limit with uninformative
priors~\cite{peixoto_hierarchical_2014,peixoto_nonparametric_2017}.  The
difference between both approaches manifests itself very often in
practice, as shown in Fig.~\ref{fig:nested-sbm}b, where systematic
underfitting is observed for a wide variety of network datasets, which
disappears with the nested model, as seen in
Fig.~\ref{fig:nested-sbm}c. Crucially, we achieve this decreased
tendency to underfit without sacrificing our protection against
overfitting: Despite the more elaborate model specification, the
inference of the nested SBM is completely nonparametric, and the same
Bayesian and information-theoretical principles still hold. Furthermore,
as we already mentioned, the uninformative case is a special case of the
nested SBM,
i.e. when $L=1$, and hence it can only improve the inference (e.g. by
reducing the description length), with no drawbacks. We stress that the
number of hierarchy levels, as with any other dimension of the model,
such as the number of groups in each level, is inferred from data, and
does not need to be determined a priori.
\begin{figure}[t!]
  \includegraphics[width=\textwidth]{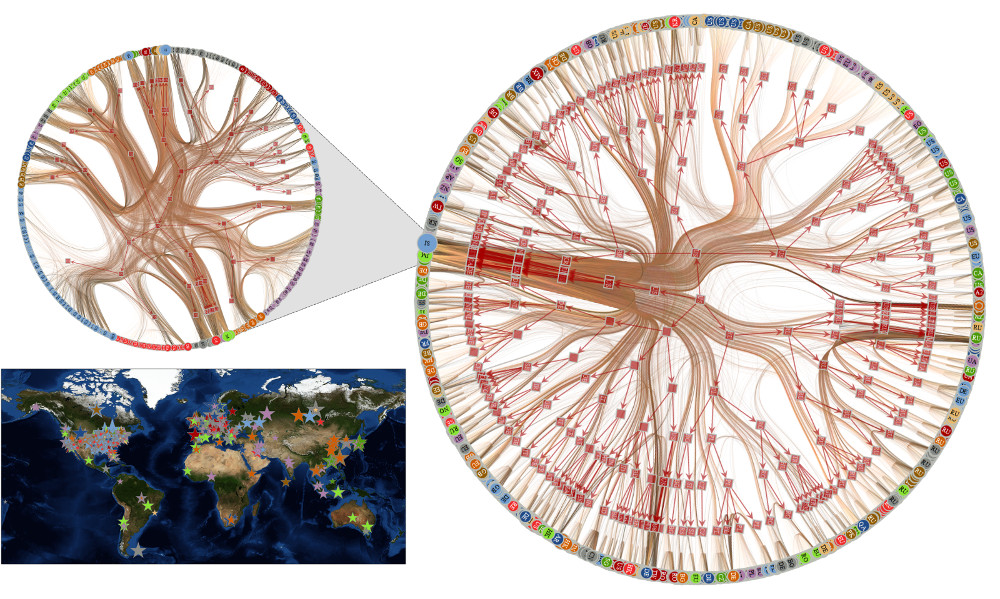}

  \caption{Fit of the (degree-corrected) nested SBM for the internet
  topology at the autonomous systems level, adapted from
  Ref.~\cite{peixoto_hierarchical_2014}. The hierarchical division
  reveals a core-periphery organization at the higher levels, where most
  routes go through a relatively small number of nodes (shown in the
  inset and in the map). The lower levels reveal a more detailed
  picture, where a large number of groups of nodes are identified
  according to their routing patterns (amounting largely to distinct
  geographical regions). The layout is obtained with an edge bundling
  algorithm by Holten~\cite{holten_hierarchical_2006}, which uses the
  hierarchical partition to route the edges.\label{fig:internet}}
\end{figure}

In addition to the above, the nested model also gives us the capacity of
describing the data at multiple scales, which could potentially exhibit
different mixing patterns. This is particularly useful for large
networks, where the SBM might still give us a very complex description,
which becomes easier to interpret if we concentrate first on the upper
levels of the hierarchy. A good example is the result obtained for the
internet topology at the autonomous systems level, shown in
Fig.~\ref{fig:internet}. The lowest level of the hierarchy shows a
division into a large number of groups, with a fairly complicated
structure, whereas the higher levels show an increasingly simplified
picture, culminating in a core-periphery organization as the dominating
pattern.

\section{Model variations}

Varying the number of groups and building hierarchies are not the only
ways we have of adapting the complexity of the model to the data. We may
also change the internal structure of the model, and how the division
into groups affect the placement of edges. In fact, the basic ansatz of
the SBM is very versatile, and many variations have been proposed in the
literature. In this section we review two important ones --- SBMs with
degree correction and group overlap --- and review other model flavors
in a summarized manner.

Before we go further into the model variations, we point out that the
multiplicity of models is a strength of the inference approach. This is
different from the broader field of network clustering, where a large
number of available algorithms often yield conflicting results for the
same data, leaving practitioners lost in how to select between
them~\cite{good_performance_2010,hric_community_2014}. Instead, within
the inference framework we can in fact compare different models in a
principled manner and select the best one according to the statistical
evidence available. We proceed with a general outline of the model
selection procedure, before following with specific model variations.

\subsection{Model selection}\index{topic}{Bayesian inference!model selection}

Suppose we define two versions of the SBM, labeled $\mathcal{C}_1$ and
$\mathcal{C}_2$, each with their own posterior distribution of
partitions, $P(\bb|\A,\mathcal{C}_1)$ and
$P(\bb|\A,\mathcal{C}_2)$. Suppose we find the most likely partitions
$\bb_1$ and $\bb_2$, according to $\mathcal{C}_1$ and $\mathcal{C}_2$,
respectively. How do we decide which partition is more representative of
the data? The consistent approach is to obtain the so-called posterior
odds ratio~\cite{jeffreys_theory_2000,jaynes_probability_2003}\index{topic}{Bayesian inference!model selection!posterior odds ratio}
\begin{equation}
  \Lambda = \frac{P(\bb_1,\mathcal{C}_1|\A)}{P(\bb_2,\mathcal{C}_2|\A)}
  = \frac{P(\A|\bb_1,\mathcal{C}_1)P(\bb_1)P(\mathcal{C}_1)}{P(\A|\bb_2,\mathcal{C}_2)P(\bb_2)P(\mathcal{C}_2)},
\end{equation}
where $P(\mathcal{C})$ is our prior belief that variant $\mathcal{C}$ is
valid. A value of $\Lambda > 1$ indicates that the choice
$(\bb_1,\mathcal{C}_1)$ is $\Lambda$ times more plausible as an
explanation for the data than the alternative,
$(\bb_2,\mathcal{C}_2)$. If we are \emph{a priori} agnostic with respect
to which model flavor is best, i.e. $P(\mathcal{C}_1)=P(\mathcal{C}_2)$,
we have then
\begin{equation}
  \Lambda = \frac{P(\A|\bb_1,\mathcal{C}_1)P(\bb_1)}{P(\A|\bb_2,\mathcal{C}_2)P(\bb_2)} = 2^{-\Delta\Sigma},
\end{equation}
where $\Delta\Sigma = \Sigma_1-\Sigma_2$ is the description length
difference between both choices. Hence, we should generally prefer the
model choice that is most compressive, i.e. with the smallest
description length. However, if the value of $\Lambda$ is close to $1$,
we should refrain from forcefully rejecting the alternative, as the
evidence in data would not be strongly decisive either way. I.e. the
actual value of $\Lambda$ gives us the confidence with which we can
choose the preferred model. The final decision, however, is subjective,
since it depends on what we might consider plausible. A value of
$\Lambda=2$, for example, typically cannot be used to forcefully reject
the alternative hypothesis, whereas a value of $\Lambda=10^{100}$ might.

An alternative test we can make is to decide which model \emph{class} is
most representative of the data, when averaged over all possible
partitions. In this case, we proceed in a an analogous way by computing
the posterior odds ratio
\begin{equation}
  \Lambda' = \frac{P(\mathcal{C}_1|\A)}{P(\mathcal{C}_2|\A)} = \frac{P(\A|\mathcal{C}_1)P(\mathcal{C}_1)}{P(\A|\mathcal{C}_2)P(\mathcal{C}_2)},
\end{equation}
where
\begin{equation}
  P(\A|\mathcal{C}) = \sum_{\bb}P(\A|\bb,\mathcal{C})P(\bb)
\end{equation}
is the model evidence. When $P(\mathcal{C}_1)=P(\mathcal{C}_2)$,
$\Lambda'$ is called the \emph{Bayes factor}\index{topic}{Bayesian
inference!model selection!Bayes factor}, with an interpretation
analogous to $\Lambda$ above, but where the statement is made with
respect to all possible partitions, not only the most likely
one. Unfortunately, as mentioned previously, the evidence
$P(\A|\mathcal{C})$ cannot be computed exactly for the models we are
interested in, making this criterion more difficult to employ in
practice (although approximations have been proposed, see
e.g. Ref~\cite{peixoto_nonparametric_2017}). We return to the issue of
when it should we optimize or sample from the posterior distribution in
Sec.~\ref{sec:sample}, and hence which of the two criteria should be
used.

\subsection{Degree correction}\index{topic}{stochastic blockmodel!degree-corrected}

\begin{figure}[b!]
  \begin{tabular}{cc}
    \begin{overpic}[width=.49\textwidth]{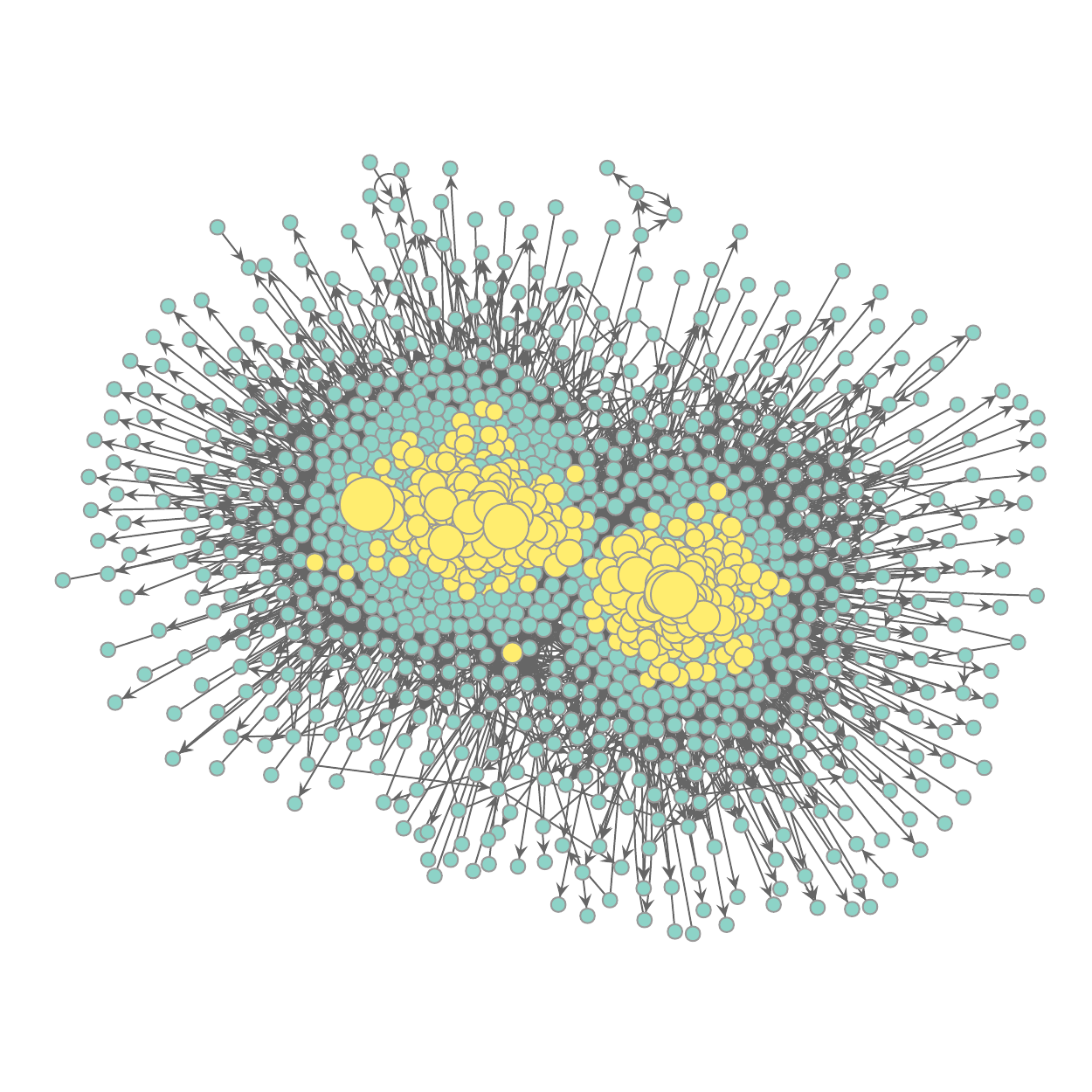}\put(0,0){(a)}\end{overpic} &
    \begin{overpic}[width=.49\textwidth]{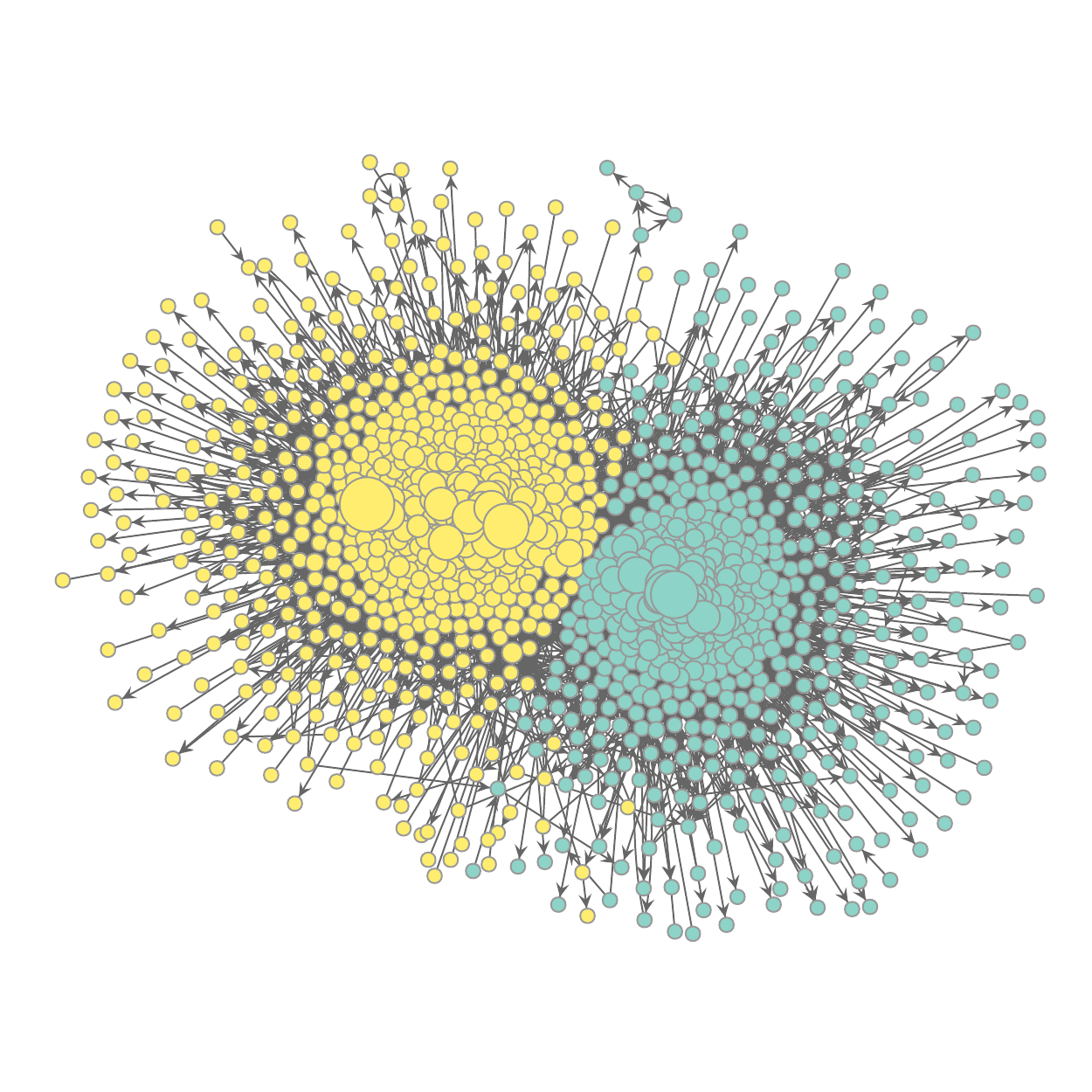}\put(0,0){(b)}\end{overpic}
  \end{tabular} \caption{Inferred partition for a network of political
  blogs~\cite{adamic_political_2005} using (a) the SBM and (b) the
  DC-SBM, in both cases forcing $B=2$ groups. The node sizes are
  proportional to the node degrees. The SBM divides the network into low
  and high-degree groups, whereas the DC-SBM prefers the division into
  political factions.\label{fig:dc-sbm}}
\end{figure}

The underlying assumption of all variants of the SBM considered so far
is that nodes that belong to the same group are statistically
equivalent. As it turns out, this fundamental aspect results in a very
unrealistic property. Namely, this generative process implies that all
nodes that belong to the same group receive on average the same number
of edges. However, a common property of many empirical networks is that
they have very heterogeneous degrees, often broadly distributed over
several orders of magnitudes~\cite{newman_networks:_2010}. Therefore, in
order for this property to be reproduced by the SBM, it is necessary to
group nodes according to their degree, which may lead to some seemingly
odd results. An example of this was given in
Ref.~\cite{karrer_stochastic_2011} and is shown in
Fig.\ref{fig:dc-sbm}a. It corresponds to a fit of the SBM to a network
of political blogs recorded during the 2004 American presidential
election campaign~\cite{adamic_political_2005}, where an edge exists
between two blogs if one links to the other. If we guide ourselves by
the layout of the figure, we identify two assortative groups, which
happen to be those aligned with the Republican and Democratic
parties. However, inside each group there is a significant variation in
degree, with a few nodes with many connections and many with very
few. Because of what just has been explained, if we perform a fit of the
SBM using only $B=2$ groups, it prefers to cluster the nodes into
high-degree and low-degree groups, completely ignoring the party
alliance.\footnote{It is possible that unexpected results of this kind
inhibited the initial adoption of SBM methods in the network science
community, which focused instead on more heuristic community detection
methods, save for a few exceptions
(e.g.~\cite{guimera_functional_2005,hastings_community_2006,
  rosvall_information-theoretic_2007, hofman_bayesian_2008,
  clauset_hierarchical_2008, guimera_missing_2009}).} Arguably, this is
a bad fit of this network, since --- similarly to the underfitting
example of Fig.~\ref{fig:underfit} --- the probability of the fitted SBM
generating a network with such a party structure is vanishingly
small. In order to solve this undesired behavior, Karrer and
Newman~\cite{karrer_stochastic_2011}\index{author}{Mark E. J.
Newman}\index{author}{Brian Karrer} proposed a modified model, which
they dubbed the degree-corrected SBM (DC-SBM). In this variation, each
node $i$ is attributed with a parameter $\theta_i$ that controls its
expected degree, independently of its group membership. Given this extra
set of parameters, a network is generated with probability
\begin{equation}\label{eq:dc-sbm}
  P(\A|\bm{\lambda},\bm{\theta},\bb) = \prod_{i< j}\frac{e^{-\theta_i\theta_j\lambda_{b_i,b_j}}(\theta_i\theta_j\lambda_{b_i,b_j})^{A_{ij}}}{A_{ij}!}\times\prod_i\frac{e^{-\theta_i^2\lambda_{b_i,b_i}/2}(\theta_i^2\lambda_{b_i,b_i}/2)^{A_{ii}/2}}{(A_{ii}/2)!},
\end{equation}
where $\lambda_{rs}$ again controls the expected number of edges between
groups $r$ and $s$. Note that since the parameters $\lambda_{rs}$ and
$\theta_i$ always appear multiplying each other in the likelihood, their
individual values may be arbitrarily scaled, provided their products
remain the same. If we choose the parametrization
$\sum_i\theta_i\delta_{b_i,r}=1$ for every group $r$, then they acquire
a simple interpretation: $\lambda_{rs}$ is the expected number of edges
between groups $r$ ans $s$, $\lambda_{rs}=\avg{e_{rs}}$, and $\theta_i$
is proportional to the expected degree of node $i$,
$\theta_i=\avg{k_i}/\sum_s\lambda_{b_i,s}$.

When inferring this model from the political blogs data --- again
forcing $B=2$ --- we obtain a much more satisfying result, where the two
political factions are neatly identified, as seen in
Fig.~\ref{fig:dc-sbm}b. As this model is capable of fully decoupling the
community structure from the degrees, which are captured separately by
the parameters $\bm{\lambda}$ and $\bm{\theta}$, respectively, the
degree heterogeneity of the network does not interfere with the
identification of the political factions.

Based on the above example, and on the knowledge that most networks
possess heterogeneous degrees, we could expect the DC-SBM to provide a
better fit for most of them. However, before we jump to this conclusion,
we must first acknowledge that the seemingly increased quality of fit
obtained with the SBM came at the expense of adding an extra set of
parameters, $\bm{\theta}$~\cite{yan_model_2014}. However intuitive we
might judge the improvement brought on by degree correction, simply
adding more parameters to a model is an almost sure recipe for
overfitting. Therefore, a more prudent approach is once more to frame
the inference problem in a Bayesian way, by focusing on the posterior
distribution $P(\bb|\A)$, and on the description length. For this, we
must include a prior for the node propensities $\bm{\theta}$. The
uninformative choice is the one which ascribes the same probability to
all possible choices,
\begin{equation}
  P(\bm{\theta}|\bb) = \prod_r(n_r-1)!\delta({\textstyle\sum_i\theta_i\delta_{b_i,r}-1}).
\end{equation}
Using again an uninformative prior for $\bm{\lambda}$,
\begin{equation}\label{eq:prior-lambda-dc}
  P(\bm{\lambda}|\bb) = \prod_{r\le s}e^{-\lambda_{rs}/(1+\delta_{rs})\bar{\lambda}}/(1+\delta_{rs})\bar{\lambda}
\end{equation}
with $\bar{\lambda}=2E/B(B+1)$, the marginal likelihood now becomes
\begin{align}
  P(\A|\bb) &= \int P(\A|\bm{\lambda},\bm{\theta},\bb)P(\bm{\lambda}|\bb)P(\bm{\theta}|\bb)\;\dd\bm{\lambda}\dd\bm{\theta}\nonumber\\
  &= \frac{\bar\lambda^E}{(\bar{\lambda}+1)^{E+B(B+1)/2}}\times\frac{\prod_{r<s}e_{rs}!\prod_re_{rr}!!}{\prod_{i<j}A_{ij}!\prod_iA_{ii}!!}
  \times \prod_r\frac{(n_r - 1)!}{(e_r + n_r-1)!} \times \prod_ik_i!,\label{eq:dc-sbm-marginal}
\end{align}
where $k_i=\sum_jA_{ij}$ is the degree of node $i$, which can be used in
the same way to obtain a posterior for $\bb$, via
Eq.~\ref{eq:b-posterior}. Once more, the model above is equivalent to a
microcanonical formulation~\cite{peixoto_nonparametric_2017}, given by
\begin{figure}
  \begin{tabular}{ccc}
    \begin{minipage}{.32\textwidth}\smaller
    \includegraphics[width=\textwidth]{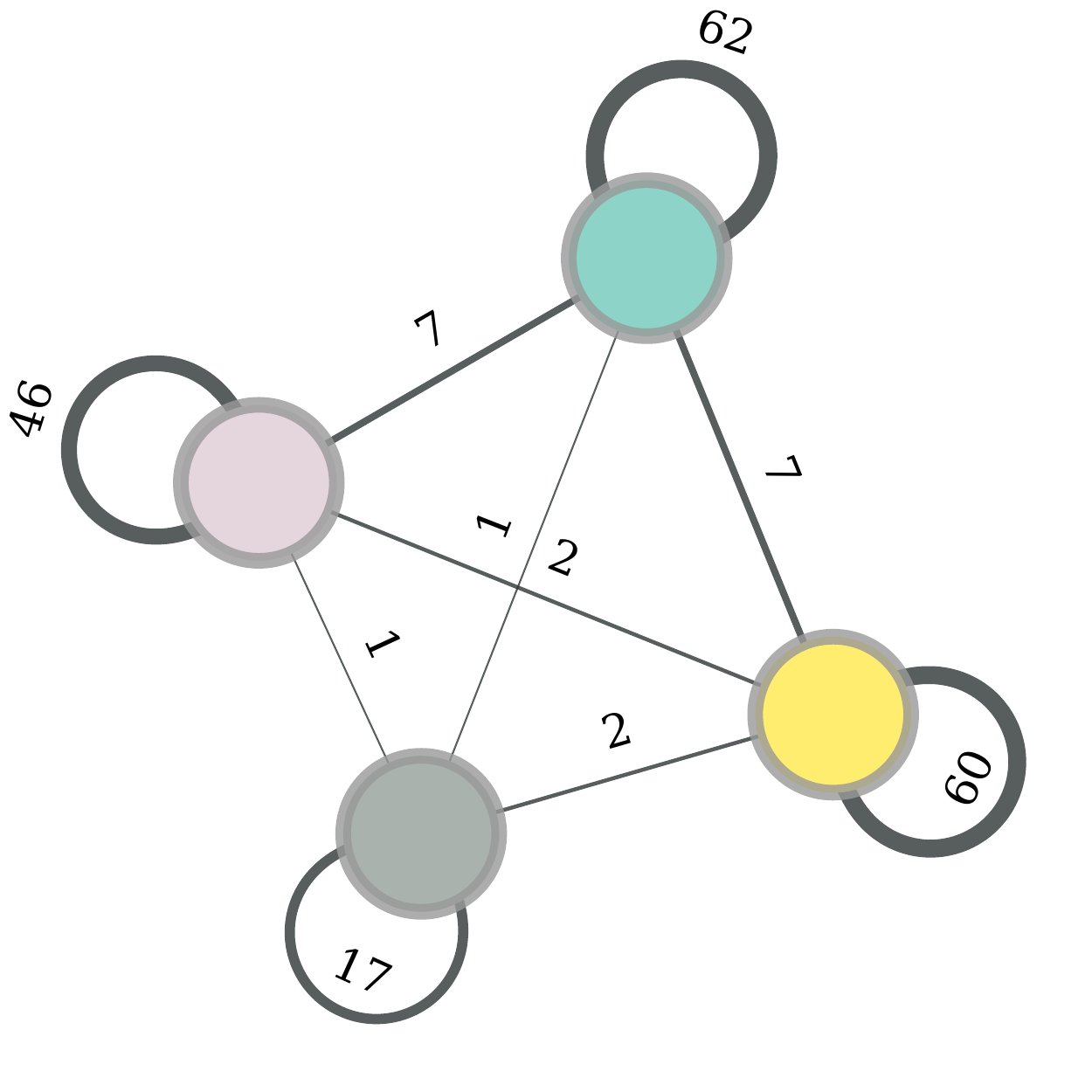}\\
    (a) Edge counts, $P(\e|\bm{b}).$
    \end{minipage}&
    \begin{minipage}{.32\textwidth}\smaller
    \includegraphics[width=\textwidth]{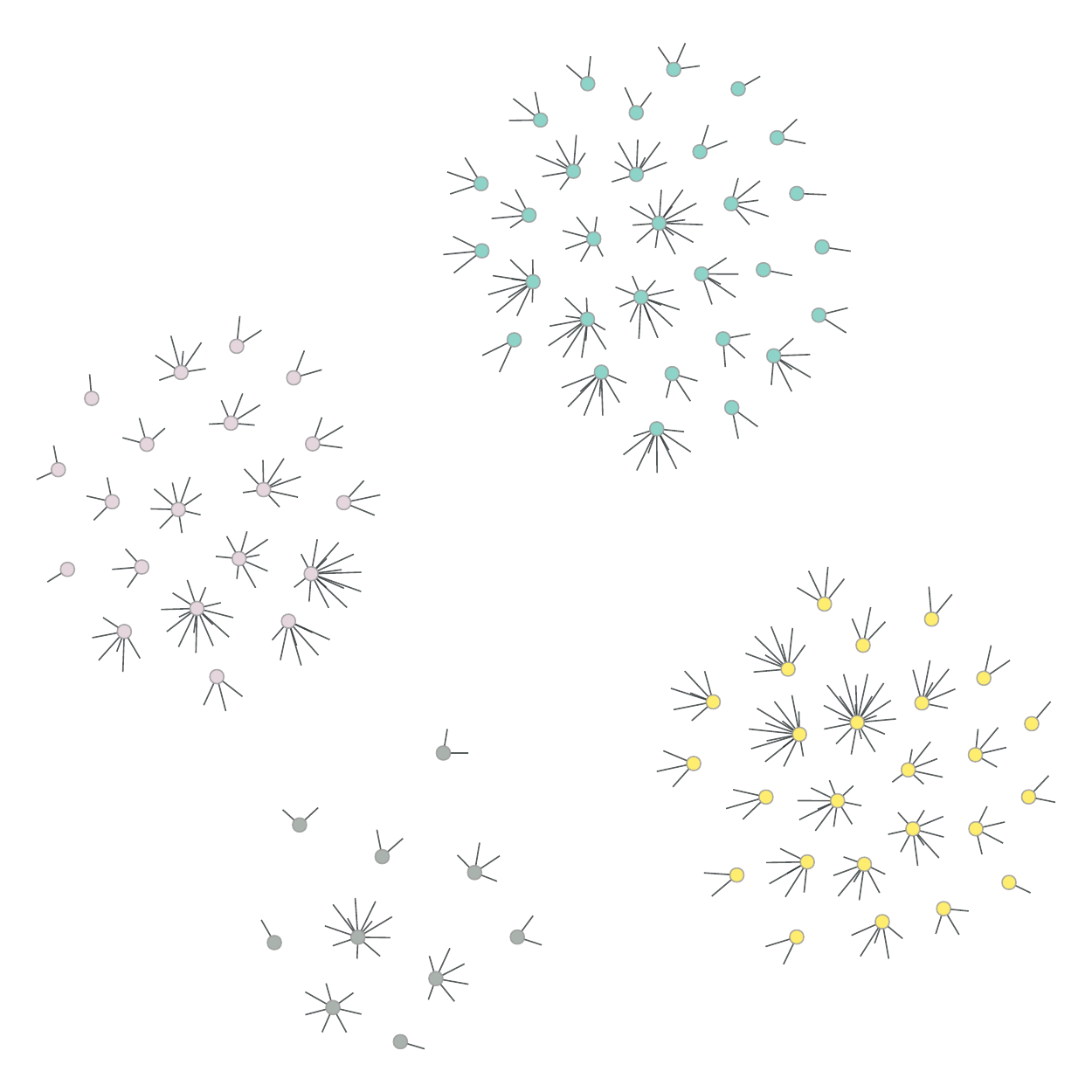}\\
    (b) Degrees, $P(\bm{k}|\e,\bm{b}).$
    \end{minipage}
    &
    \begin{minipage}{.32\textwidth}\smaller
    \includegraphics[width=\textwidth]{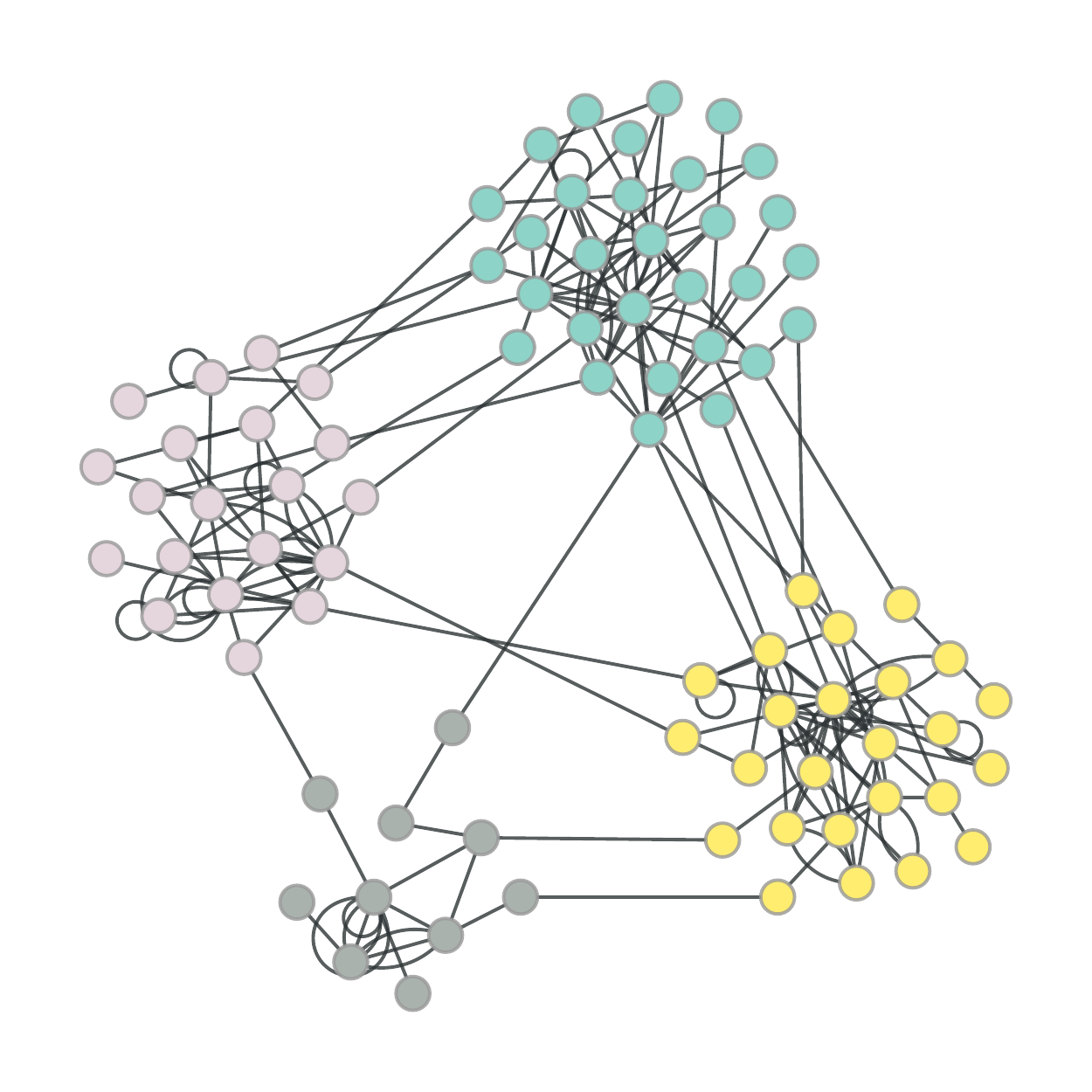}\\
    (c) Network, $P(\bm{A}|\bm{k},\e,\bm{b}).$
    \end{minipage}
  \end{tabular}

  \caption{Illustration of the generative process of the microcanonical
  DC-SBM. Given a partition of the nodes, the edge counts between groups
  are sampled (a), followed by the degrees of the nodes (b) and finally
  the network itself (c). Adapted from
  Ref.~\cite{peixoto_nonparametric_2017}.\label{fig:generative}}
\end{figure}
\begin{equation}\label{eq:dc-ensemble-equivalence}
  P(\A|\bb) = P(\A|\bm{k},\e,\bb)P(\bm{k}|\e,\bb)P(\e|\bb),
\end{equation}
with
\begin{align}
  P(\A|\bm{k},\e,\bb) &= \frac{\prod_{r<s}e_{rs}!\prod_re_{rr}!!\prod_ik_i!}{\prod_{i<j}A_{ij}!\prod_iA_{ii}!!\prod_re_r!},\label{eq:micro-dc-sbm}\\
  P(\bm{k}|\e,\bb) &= \prod_r\multiset{n_r}{e_r}^{-1},\label{eq:micro-uniform-degrees}
\end{align}
and $P(\e|\bb)$ given by Eq.~\ref{eq:prior-edges}. In the model above,
$P(\A|\bm{k},\e,\bb)$ is the probability of generating a multigraph
where the edge counts between groups \emph{as well as} the degrees
$\bm{k}$ are fixed to specific values.\footnote{The ensemble equivalence
of Eq.~\ref{eq:dc-ensemble-equivalence} is in some ways more remarkable
than for the traditional SBM. This is because a direct equivalence between
the ensembles of Eqs.~\ref{eq:dc-sbm} and \ref{eq:micro-dc-sbm} is not
satisfied even in the asymptotic limit of large
networks~\cite{peixoto_entropy_2012,garlaschelli_ensemble_2017}, which
does happen for Eqs.~\ref{eq:poisson-sbm}
and~\ref{eq:micro-sbm}. Equivalence is observed only if the individual
degrees $k_i$ also become asymptotically large. However, when the
parameters $\bm{\lambda}$ and $\bm{\theta}$ are integrated out, the
equivalence becomes exact for networks of any size.} The prior
$P(\bm{k}|\e,\bb)$ is the uniform probability of generating a degree
sequence, where all possibilities that satisfy the constraints imposed
by the edge counts $\e$, namely $\sum_ik_i\delta_{b_i,r}=e_r$, occur
with the same probability. The description length of this model is then
given by
\begin{equation}
  \Sigma = -\log_2 P(\A,\bb) = -\log_2P(\A|\bm{k},\e,\bb) - \log_2P(\bm{k},\e,\bb).
\end{equation}
Because uninformative priors were used to derive the above equations,
we are once more subject to the same underfitting problem described
previously. Luckily, from the microcanonical model we can again derive a
nested DC-SBM, by replacing $P(\e)$ by a nested sequence of SBMs,
exactly in the same was as was done
before~\cite{peixoto_hierarchical_2014,peixoto_nonparametric_2017}. We
also have the opportunity of replacing the uninformative prior for the
degrees in Eq.~\ref{eq:micro-uniform-degrees} with a more realistic
option. As was argued in Ref.~\cite{peixoto_nonparametric_2017}, degree
sequences generated by Eq.~\ref{eq:micro-uniform-degrees} result in
exponential degree distributions, which are not quite as heterogeneous
as what is often encountered in practice. A more refined approach, which
is already familiar to us at this point, is to increase the Bayesian
hierarchy, and choose a prior that is conditioned on a higher-order
aspect of the data, in this case the \emph{frequency} of degrees, i.e.
\begin{align}\label{eq:micro-distributed-degrees}
  P(\bm{k}|\e,\bb) &= P(\bm{k}|\e,\bb,\bm{\eta})P(\bm{\eta}|\e,\bb),
\end{align}
where $\bm{\eta}=\{\eta_k^r\}$, with $\eta_k^r$ being the number of
nodes of degree $k$ in group $r$. In the above, $P(\bm{\eta}|\e,\bb)$ is
a uniform distribution of frequencies, and $P(\bm{k}|\e,\bb,\bm{\eta})$
generates the degrees according to the sampled frequencies (we omit the
respective expressions for brevity, and refer to
Ref.~\cite{peixoto_nonparametric_2017} instead). Thus, this model is
capable of using regularities in the degree distribution to inform the
division into groups, and is generally capable of better fits than the
uniform model of Eq.~\ref{eq:micro-uniform-degrees}.

\begin{figure}
  \begin{tabular}{ccc}
    \begin{overpic}[width=.32\textwidth,trim=1cm 0 0 0]{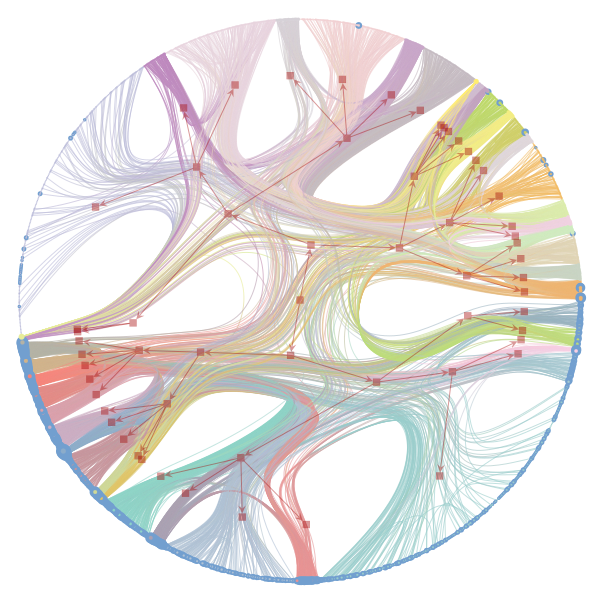}\put(0,0){(a)}\end{overpic}&
    \begin{overpic}[width=.32\textwidth,trim=1cm 0 0 0]{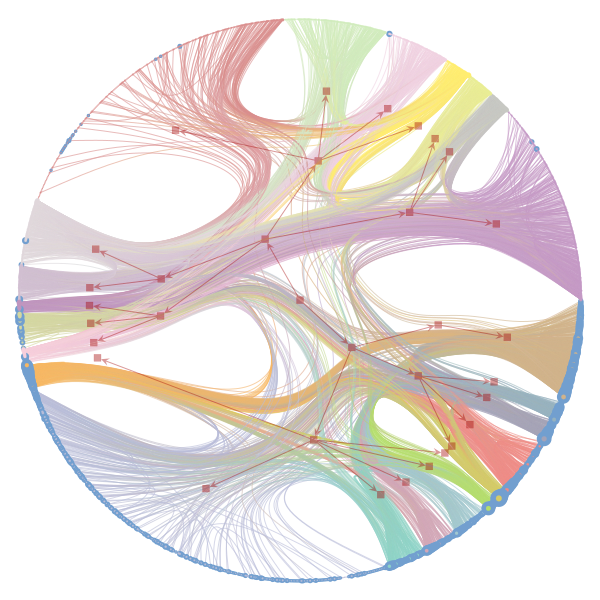}\put(0,0){(b)}\end{overpic}&
    \begin{overpic}[width=.32\textwidth,trim=1cm 0 0 0]{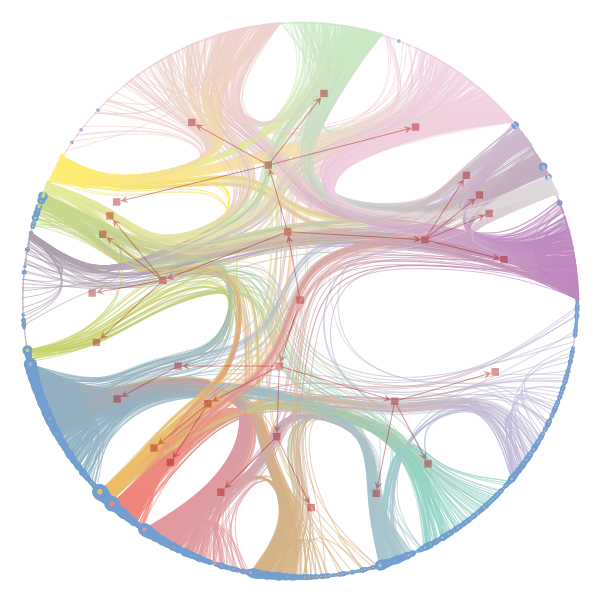}\put(0,0){(c)}\end{overpic}
  \end{tabular}

  \caption{Most likely hierarchical partitions of a network of political
  blogs~\cite{adamic_political_2005}, according to the three model
  variants considered, as well as the inferred number of groups $B_1$ at
  the bottom of the hierarchy, and the description length $\Sigma$:
  (a) NDC-SBM, $B_1=42$, $\Sigma \approx 89,938$ bits,
  (b) DC-SBM, $B_1=23$, $\Sigma \approx 87,162$ bits,
  (c) DC-SBM with the degree prior of
  Eq.~\ref{eq:micro-distributed-degrees}, $B_1=20$, $\Sigma \approx
  84,890$ bits.  The nodes circled in blue were classified as
  ``liberals'' and the remaining ones as ``conservatives'' in
  Ref.~\cite{adamic_political_2005} based on the blog contents. Adapted
  from Ref.~\cite{peixoto_nonparametric_2017}.\label{fig:polblogs}}
\end{figure}

If we apply this nonparametric approach to the same political blog
network of Ref.~\ref{fig:polblogs}, we find a much more detailed picture
of its structure, revealing many more than two groups, as shown in
Fig.~\ref{fig:polblogs}, for three model variants: the nested SBM, the
nested DC-SBM and the nested DC-SBM with the degree prior of
Eq.~\ref{eq:micro-distributed-degrees}. All three model variants are in
fact capable of identifying the same Republican/Democrat division at the
topmost hierarchical level --- showing that the non-degree-corrected SBM
is not as inept in capturing this aspect of the data as the result
obtained by forcing $B=2$ might suggest. However the internal divisions
of both factions that they uncover are very distinct from each other. If
we inspect the obtained values of the description length with each model
we see that the DC-SBM (in particular when using
Eq.~\ref{eq:micro-distributed-degrees}) results in a smaller value,
indicating that it better captures the structure of the data, despite
the increased number of parameters. Indeed, a systematic analysis
carried out in Ref.~\cite{peixoto_nonparametric_2017} showed that the
DC-SBM does in fact yield shorter description lengths for a majority of
empirical datasets, thus ultimately confirming the original intuition
behind the model formulation.

\subsection{Group overlaps}\label{sec:overlap}\index{topic}{stochastic blockmodel!overlapping}\index{topic}{stochastic blockmodel!mixed-membership}

Another way we can change the internal structure of the model is to
allow the groups to overlap, i.e. we allow a node to belong to more than
one group at the same time. The connection patterns of the nodes are
then assumed to be a mixture of the ``pure'' groups, which results in a
richer type of model~\cite{airoldi_mixed_2008}. Following Ball et
al.~\cite{ ball_efficient_2011}, we can adapt the Poisson formulation to
overlapping SBMs in a straightforward manner,
\begin{equation}\label{eq:odc-sbm}
  P(\bm{A}|\bm{\kappa},\bm{\lambda}) = \prod_{i<j}\frac{e^{-\lambda_{ij}}\lambda_{ij}^{A_{ij}}}{A_{ij}!}\prod_i\frac{e^{-\lambda_{ii}/2}(\lambda_{ii}/2)^{A_{ii}/2}}{A_{ii}/2!},
\end{equation}
with
\begin{equation}
  \lambda_{ij} = \sum_{rs}\kappa_{ir}\lambda_{rs}\kappa_{js},
\end{equation}
where $\kappa_{ir}$ is the probability with which node $i$ is chosen
from group $r$, so that $\sum_i\kappa_{ir}=1$, and $\lambda_{rs}$ is
once more the expected number of edges between groups $r$ and $s$. The
parameters $\bm{\kappa}$ replace the disjoint partition $\bb$ we have
been using so far by a ``soft'' clustering into overlapping
categories\footnote{Note that, differently from the non-overlapping
case, here it is possible for a node not to belong to any group, in
which case it will never receive an incident edge.}. Note, however,
that this model is a direct generalization of the non-overlapping DC-SBM
of Eq.~\ref{eq:dc-sbm}, which is recovered simply by choosing
$\kappa_{ir}=\theta_i\delta_{r,b_i}$. The Bayesian formulation can also
be performed by using an uninformative prior for $\bm{\kappa}$,
\begin{equation}
  P(\bm{\kappa}) = \prod_r(N-1)!\delta(\textstyle\sum_i\kappa_{ir}-1),
\end{equation}
in addition to the same prior for $\bm{\lambda}$ in
Eq.~\ref{eq:prior-lambda-dc}. Unfortunately, computing the marginal
likelihood using Eq.~\ref{eq:odc-sbm} directly,
\begin{equation}
  P(\A|\bm{\kappa})=\int P(\bm{A}|\bm{\kappa},\bm{\lambda})P(\bm{\lambda})\,\dd\bm{\lambda},
\end{equation}
is not tractable, which prevents us from obtaining the posterior
$P(\bm{\kappa}|\A)$. Instead, it is more useful to consider the
auxiliary labelled matrix, or tensor, $\bm{G}=\{G_{ij}^{rs}\}$, where
$G_{ij}^{rs}$ is a particular decomposition of $A_{ij}$ where the two
edge endpoints --- or ``half-edges''
--- of an edge $(i,j)$ are labelled with groups $(r,s)$, such that
\begin{equation}\label{eq:oconstraint}
  A_{ij} = \sum_{rs}G_{ij}^{rs}.
\end{equation}
Since a sum of Poisson variables is also distributed according to a
Poisson, we can write Eq.~\ref{eq:odc-sbm} as
\begin{align}
  P(\bm{A}|\bm{\kappa},\bm{\lambda}) &= \sum_{\bm{G}}P(\bm{G}|\bm{\kappa},\bm{\lambda})\prod_{i\le j}\delta_{\sum_{rs}G_{ij}^{rs},A_{ij}},
\end{align}
with each half-edge labelling being generated by
\begin{equation}
  P(\bm{G}|\bm{\kappa},\bm{\lambda}) =
  \prod_{i<j}\prod_{rs}\frac{e^{-\kappa_{ir}\lambda_{rs}\kappa_{js}}(\kappa_{ir}\lambda_{rs}\kappa_{js})^{G_{ij}^{rs}}}{G_{ij}^{rs}!}\times\prod_i\prod_{rs}\frac{e^{-\kappa_{ir}\lambda_{rs}\kappa_{is}/2}(\kappa_{is}\lambda_{rs}\kappa_{is}/2)^{G_{ii}^{rs}/2}}{(G_{ii}^{rs}/2)!}.
\end{equation}
We can now compute the marginal likelihood as
\begin{align}\label{eq:osbm-marginal}
  P(\bm{G}) &= \int P(\bm{G}|\bm{\kappa},\bm{\lambda})P(\bm{\kappa})P(\bm{\lambda}|\bar{\lambda})\; \dd\bm{\kappa}\dd\bm{\lambda},\nonumber\\
  &= \frac{\bar\lambda^E}{(\bar\lambda+1)^{E+B(B+1)/2}}\frac{\prod_{r<s}e_{rs}!\prod_re_{rr}!!}{\prod_{rs}\prod_{i<j}G_{ij}^{rs}!\prod_iG_{ii}^{rs}!!}\times
   \prod_r\frac{(N-1)!}{(e_r+N-1)!}\times\prod_{ir}k_i^r!,
\end{align}
which is very similar to Eq.~\ref{eq:dc-sbm-marginal} for the
DC-SBM. With the above, and knowing from Eq.~\ref{eq:oconstraint} that
there is only one choice of $\A$ that is compatible with any given
$\bm{G}$, i.e.
\begin{equation}
  P(\A|\bm{G}) = \prod_{i\le j}\delta_{\sum_{rs}G_{ij}^{rs},A_{ij}},
\end{equation}
we can sample from (or maximize) the posterior distribution of the
half-edge labels $\bm{G}$, just like we did for the node partition $\bb$
in the nonoverlapping models,
\begin{equation}
  P(\bm{G}|\A) = \frac{P(\A|\bm{G})P(\bm{G})}{P(\A)} \propto P(\bm{G}) \times \prod_{i\le j}\delta_{\sum_{rs}G_{ij}^{rs},A_{ij}},
\end{equation}
where the product in the last term only accounts for choices of $\bm{G}$
which are compatible with $\A$, i.e. fulfill
Eq.~\ref{eq:oconstraint}.
Once more, the model of Eq.~\ref{eq:osbm-marginal} is equivalent to its
microcanonical analogue~\cite{peixoto_model_2015},
\begin{equation}
  P(\bm{G}) = P(\bm{G}|\bm{k},\e)P(\bm{k}|\e)P(\e),
\end{equation}
where
\begin{align}
  P(\bm{G}|\bm{k},\e) &= \frac{\prod_{r<s}e_{rs}!\prod_re_{rr}!!\prod_{ir}k_i^r!}{\prod_{rs}\prod_{i<j}G_{ij}^{rs}!\prod_iG_{ii}^{rs}!!\prod_re_r!},\\
  P(\bm{k}|\e) &= \prod_r\multiset{N}{e_r}^{-1}\label{eq:kflat_prior}
\end{align}
and $P(\e)$ given by Eq.~\ref{eq:prior-edges}. The variables
$\bm{k}=\{k_i^r\}$ are the labelled degrees of the labelled network
$\bm{G}$, where $k_i^r$ is the number of incident edges of type $r$ a
node $i$ has.
The description length becomes likewise
\begin{equation}
  \Sigma = -\log_2P(\bm{G}) = -\log_2P(\bm{G}|\bm{k},\e)-\log_2P(\bm{k}|\e)-\log_2P(\e).
\end{equation}
The nested variant can be once more obtained by replacing $P(\e)$ in the
same manner as before, and $P(\bm{k}|\e)$ in a manner that is
conditioned on the labelled degree frequencies and degree of overlap, as
described in detail in Ref.~\cite{peixoto_model_2015}.

\begin{figure}
  \begin{tabular}{ccc}
    \begin{overpic}[width=.32\textwidth,trim=0 2cm 0 2cm]{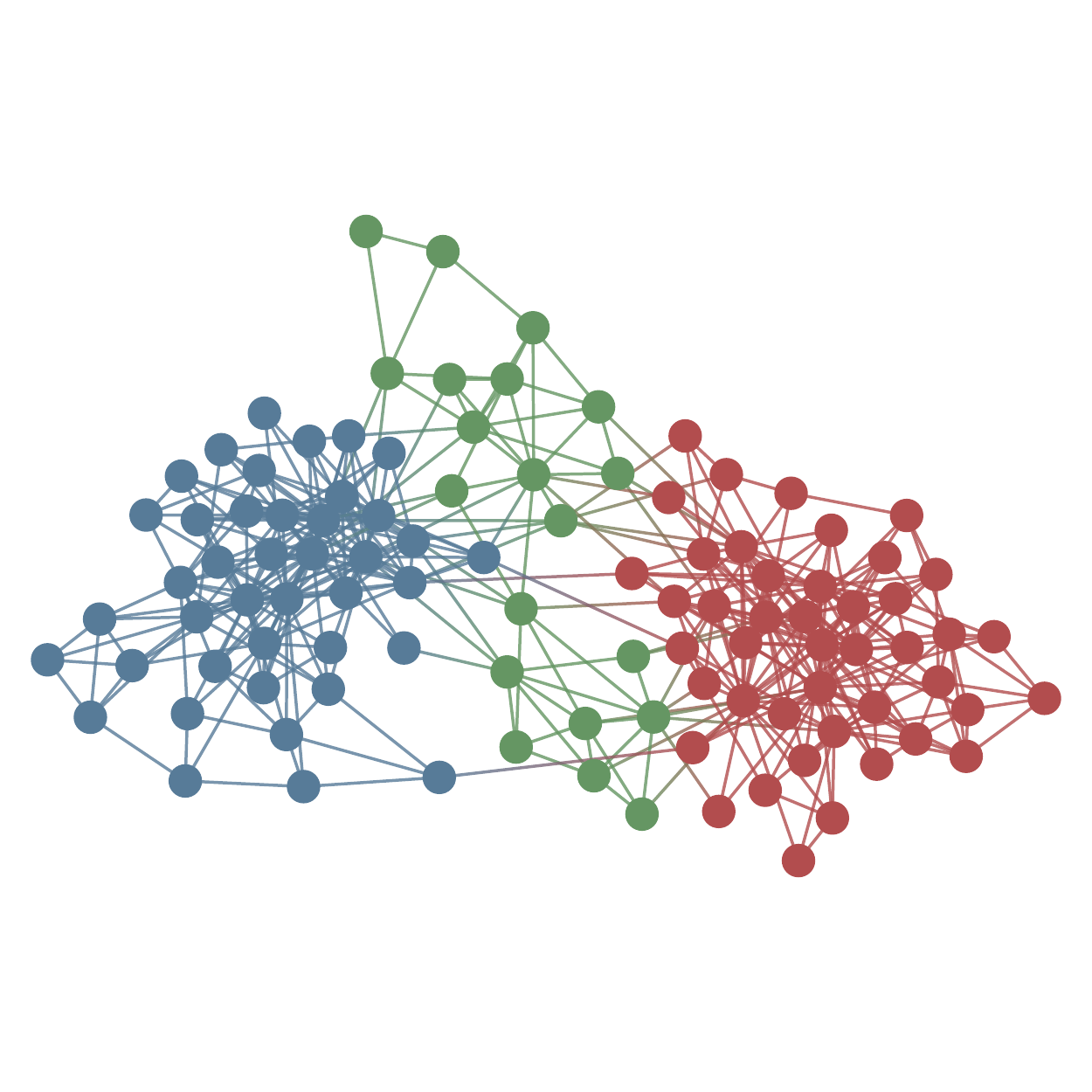}\put(0,0){(a)}\end{overpic}&
    \begin{overpic}[width=.32\textwidth,trim=0 2cm 0 2cm]{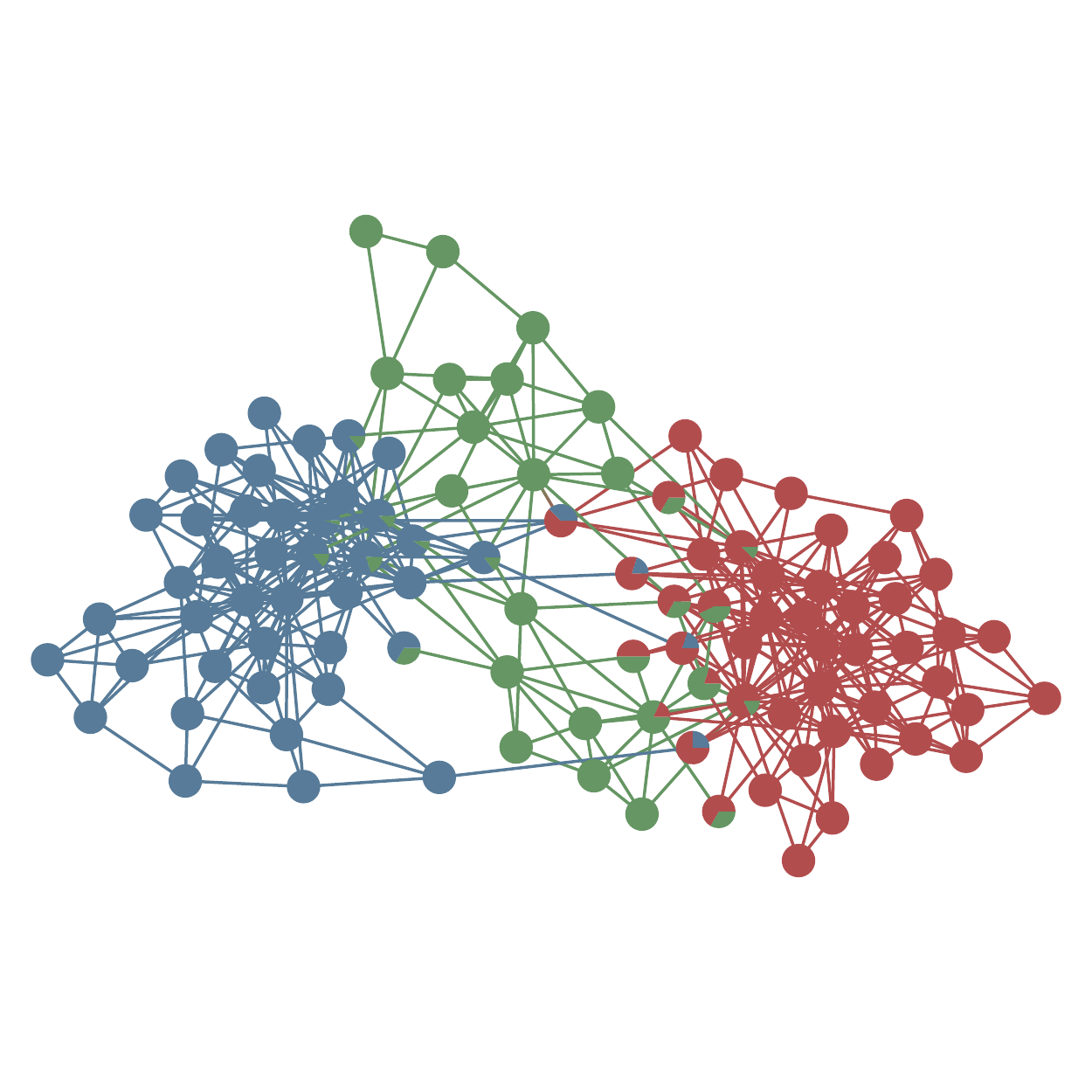}\put(0,0){(b)}\end{overpic}&
    \begin{overpic}[width=.32\textwidth,trim=0 2cm 0 2cm]{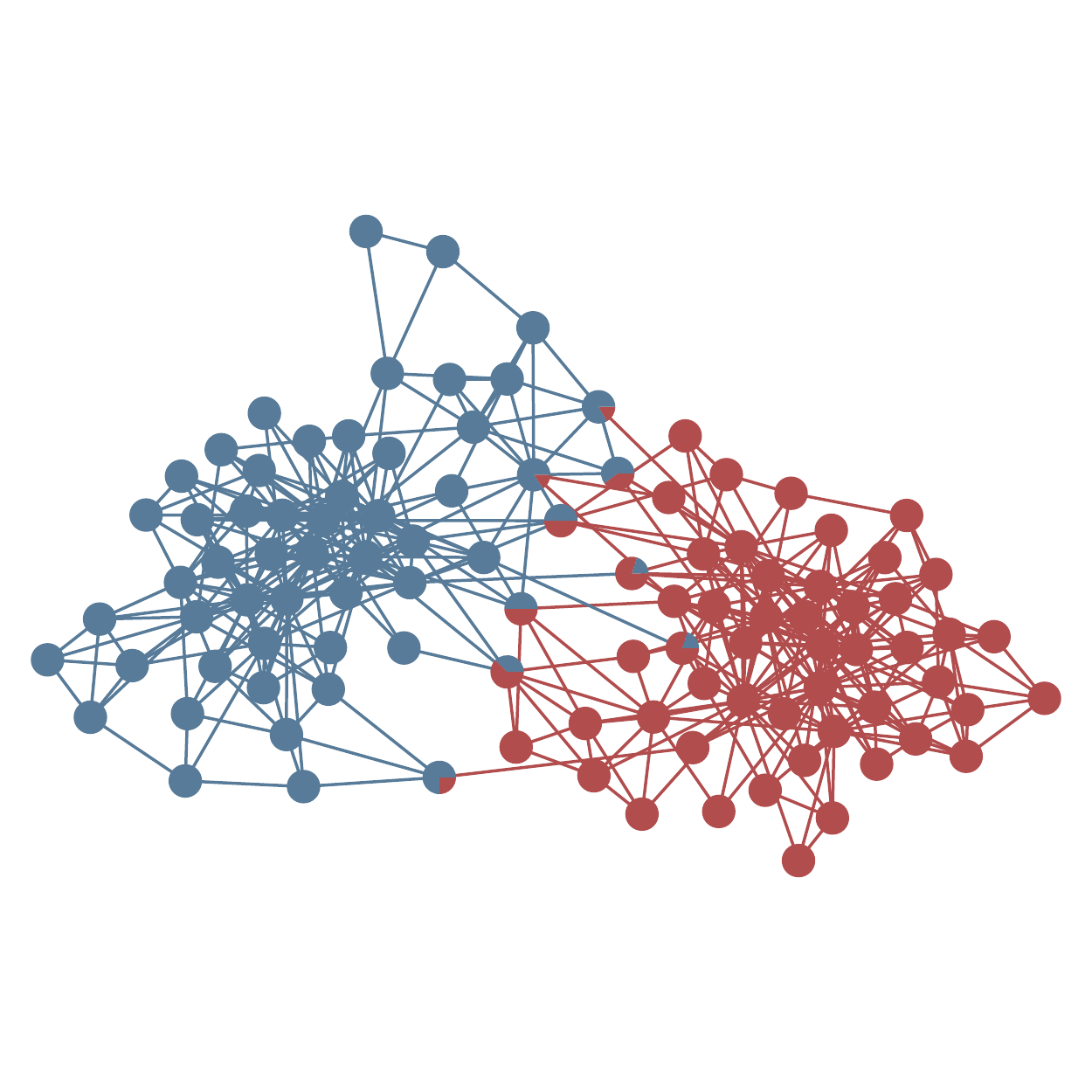}\put(0,0){(c)}\end{overpic}
  \end{tabular}

  \smaller[2]
  \caption{Network of co-purchases of books about US
  politics~\cite{krebs_political_nodate}, with groups inferred using (a)
  the non-overlapping DC-SBM, with description length $\Sigma\approx
  1,938$ bits, (b) the overlapping SBM with description length
  $\Sigma\approx 1,931$ bits and (c) the overlapping SBM forcing only
  $B=2$ groups, with description length $\Sigma\approx 1,946$
  bits.\label{fig:sbm-overlap}}
\end{figure}

Equipped with this more general model, we may ask ourselves again if it
provides a better fit of most networks, like we did for the DC-SBM in
the previous section. Indeed, since the model is more general, we might
conclude that this is a inevitability. However, this could be a fallacy,
since more general models also include more parameters and hence are
more likely to overfit. Indeed, previous claims about the existence of
``pervasive overlap'' in networks, based on nonstatistical
methods~\cite{ahn_link_2010}, seemed to be based to some extent on this
problematic logic. Claims about community overlaps are very different
from, for example, the statement that networks possess heterogeneous
degrees, since community overlap is not something that can be observed
directly; instead it is something that must be \emph{inferred}, which is
precisely what our Bayesian approach is designed to do in a
methodologically correct manner. An example of such a comparison is
shown in Fig~\ref{fig:sbm-overlap}, for a small network of political
books. This network, when analyzed using the nonoverlapping SBM, seems
to be composed of three groups, easily interpreted as ``left wing,''
``right wing'' and ``center,'' as the available metadata
corroborates. If we fit the overlapping SBM, we observe a mixed division
into the same kinds of group. If we force the inference of only two
groups, we see that some of the ``center'' nodes are split between
``right wing'' or ``left wing.'' The latter might seem like a more
pleasing interpretation, but looking at the description length reveals
that it does not improve the description of the data. The best model in
this case does seem to be the overlapping SBM with $B=3$
groups. However, the difference in the description length between all
model variants is not very large, making it difficult to fully reject
any of the three variants. A more systematic analysis done in
Ref.~\cite{peixoto_model_2015} revealed that for most empirical
networks, in particular larger ones, the overlapping models do not
provide the best fits in the majority of cases, and yield larger
description lengths than the nonoverlapping variants. Hence it seems
that the idea of overlapping groups is less pervasive than that of
degree heterogeneity --- at least according to our modeling ansatz.

\begin{figure}
  \begin{tabular}{ccc}
    \begin{overpic}[width=.28\textwidth,trim=0cm 0 0 0]{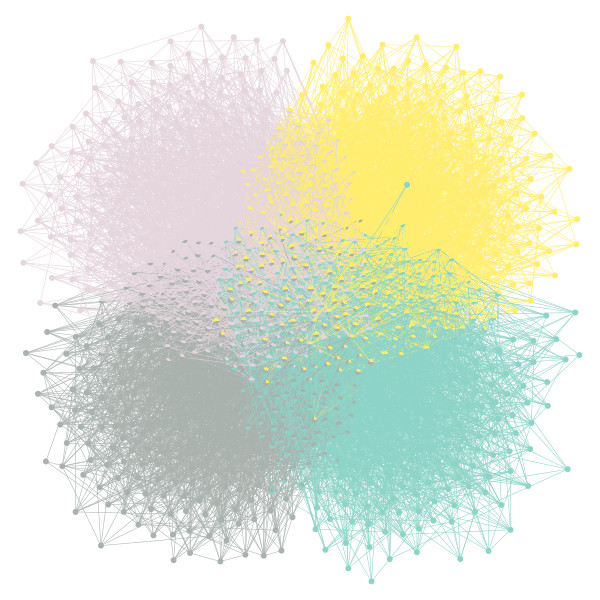}\put(0,0){(a)}\end{overpic}&
    \begin{overpic}[width=.28\textwidth,trim=0cm 0 0 0]{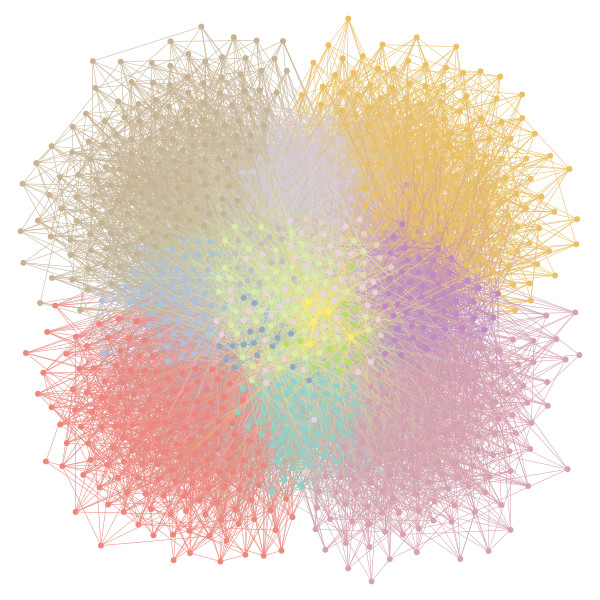}\put(0,0){(b)}\end{overpic}&
    \begin{overpic}[width=.36\textwidth,trim=0cm 0 0 0]{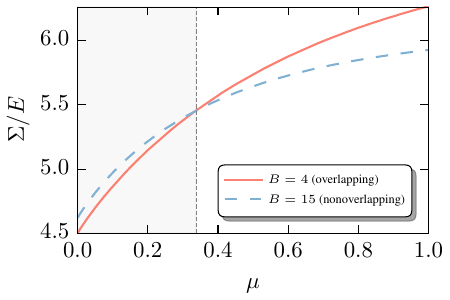}\put(0,0){(c)}\end{overpic}
  \end{tabular} \caption{(a) Artificial network sampled from an assortative
    overlapping SBM with $B=4$ groups and expected mixture sizes given
    by $n_{\vec b}\propto \mu^{|\vec{b}|}$, with $\mu\in[0,1]$
    controlling the degree of overlap (see
    Ref.~\cite{peixoto_inferring_2015} for details). (b) the same
    network as in (a), but generated according to an equivalent
    nonoverlapping SBM with $B=15$ groups. (c) Description length per
    edge $\Sigma/E$ for the same models in (a) and (b), as a function of
    the degree of overlap $\mu$, showing a cross-over where the
    nonoverlapping model is preferred. Adapted from
    Ref.~\cite{peixoto_inferring_2015}.\label{fig:overlap-equiv}}
\end{figure}

It should be emphasized that we can always represent a network generated
by an overlapping SBM by one generated with the nonoverlapping SBM with
a larger number of groups representing the individual types of
mixtures. Although model selection gives us the most parsimonious choice
between the two, it does not remove the equivalence. In
Fig.~\ref{fig:overlap-equiv} we show how networks generated by the
overlapping SBM can be better represented by the nonoverlapping SBM
(i.e. with a smaller description length) as long as the overlapping
regions are sufficiently large.

\subsection{Further model extensions}

The simple and versatile nature of the SBM has spawned a large family of
extensions and generalizations incorporating various types of more
realistic features. This includes, for example, versions of the SBM that
are designed for networks with continuous edge covariates (a.k.a. edge
weights)~\cite{aicher_learning_2014,peixoto_nonparametric_2018}\index{topic}{stochastic
blockmodel!weighted}, multilayer networks that are composed of
different types of edges~\cite{peixoto_inferring_2015,
stanley_clustering_2016, paul_consistent_2016,
valles-catala_multilayer_2016,
de_bacco_community_2017}\index{topic}{stochastic blockmodel!multilayer}, networks that evolve in
time~\cite{fu_dynamic_2009,xu_dynamic_2014,peixoto_modelling_2017,
  peel_detecting_2015,ghasemian_detectability_2016,zhang_random_2017,corneli_exact_2016,matias_catherine_statistical_2016}\index{topic}{stochastic blockmodel!dynamic},
networks that possess node attributes~\cite{peel_active_2015} or are
annotated with
metadata~\cite{newman_structure_2016,hric_network_2016}\index{topic}{stochastic
blockmodel!annotated}, networks with uncertain
structure~\cite{martin_structural_2016}, as well as networks that do not
possess a discrete modular structure at all, and are instead embedded in
generalized continuous
spaces~\cite{newman_generalized_2015}\index{topic}{stochastic blockmodel!generalized communities}. These model variations are too numerous
to be described here in any detail. But it suffices to say that the
general Bayesian approach outlined here, including model selection, also
applicable to these variations, without any conceptual difficulty.

\section{Efficient inference using Markov chain Monte Carlo (MCMC)}\label{sec:mcmc}\index{topic}{Markov chain Monte Carlo (MCMC)}

Although we can write exact expressions for the posterior probability
of Eq.~\ref{eq:b-posterior} (up to a normalization constant) for a
variety of model variants, the resulting distributions are not simple
enough to allow us to sample from them --- much less find their maximum
--- in a direct manner. In fact, fully characterizing the posterior
distribution or finding its maximum is, for most models like the SBM,
typically a NP-hard problem. What we can do, however, is to employ
Markov chain Monte Carlo (MCMC)~\cite{newman_monte_1999}, which can be
done efficiently, and in an asymptotically exact manner, as we now show.
The central idea is to sample from $P(\bb|\A)$ by first starting from
some initial configuration $\bb_0$ (in principle arbitrary), and making
move proposals $\bb\to\bb'$ with a probability $P(\bb'|\bb)$, such that,
after a sufficiently long time, the equilibrium distribution is given
exactly by $P(\bb|\A)$. In particular, given any arbitrary move
proposals $P(\bb'|\bb)$ --- with the only condition that they fulfill
ergodicity,
i.e. that they allow every state to be visited eventually --- we can
guarantee that the desired posterior distribution is eventually reached
by employing the Metropolis-Hastings
criterion~\cite{metropolis_equation_1953, hastings_monte_1970}, which
dictates we should accept a given move proposal $\bb \to \bb'$ with a
probability $a$ given by
\begin{align}\label{eq:metropolis}
  a &= \operatorname{min}\left(1,\frac{P(\bb'|\A)}{P(\bb|\A)}
  \frac{P(\bb|\bb')}{P(\bb'|\bb)}\right),
\end{align}
otherwise the proposal is rejected. The ratio $P(\bb|\bb')/P(\bb'|\bb)$
in Eq.~\ref{eq:metropolis} enforces a property known as \emph{detailed
balance} or \emph{reversibility}, i.e.
\begin{align}\label{eq:detailed}
  T(\bb'|\bb)P(\bb|\A) = T(\bb|\bb')P(\bb'|\A),
\end{align}
where $T(\bb'|\bb)$ are the final transition probabilities after
incorporating the acceptance criterion of Eq.~\ref{eq:metropolis}. The
detailed balance condition of Eq.~\ref{eq:detailed} together with the
ergodicity property guarantee that the Markov chain will converge to the
desired equilibrium distribution $P(\bb|\A)$. Importantly, we note that
when computing the ratio $P(\bb'|\A)/P(\bb|\A)$ in
Eq.~\ref{eq:metropolis}, we do not need to determine the intractable
normalization constant of Eq.~\ref{eq:b-posterior}, since it cancels
out, and thus it can be performed exactly.

The above gives a generic protocol that we can use to sample from the
posterior whenever we can compute the numerator of
Eq.~\ref{eq:b-posterior}. If instead we are interested in maximizing the
posterior, we can introduce an ``inverse temperature'' parameter
$\beta$, by changing $P(\bb|\A)\to P(\bb|\A)^{\beta}$ in the above
equations, and making $\beta\to\infty$ in slow increments; what is known
as \emph{simulated annealing}~\cite{kirkpatrick_optimization_1983}. The
simplest implementation of this protocol for the inference of SBMs is to
start from a random partition $\bb_0$, and use move proposals where a
node $i$ is randomly selected, and then its new group membership $b_i'$
is chosen randomly between all $B+1$ choices (where the remaining choice
means we populate a new group),
\begin{equation}\label{eq:random-move}
  P(b_i'|\bb) = \frac{1}{B+1}.
\end{equation}
By inspecting Eqs.~\ref{eq:sbm-marginal}, \ref{eq:dc-sbm-marginal},
\ref{eq:osbm-marginal} and \ref{eq:partition-prior} for all SBM variants
considered, we notice that the ratio $P(\bb'|\A)/P(\bb|\A)$ can be
computed in time $O(k_i)$, where $k_i$ is the degree of node $i$,
independently of other properties of the model such as the number of
groups $B$. Note that this is not true for all alternative formulations
of the SBM; e.g. for the models in
Refs.~\cite{gopalan_efficient_2013,schmidt_nonparametric_2013,come_model_2015,newman_estimating_2016,riolo_efficient_2017}
computing such an update requires $O(k_i + B)$ time [the heat-bath move
proposals of Ref.~\cite{newman_estimating_2016} increases this even
further to $O(B(k_i + B))$], thus making them very inefficient for large
networks, where the number of groups can reach the order of thousands or
more. Hence, when using these move proposals, a full sweep of all $N$
nodes in the network can be done in time $O(E)$, independent of $B$.

\begin{figure}
  \begin{tabular}{cc}
    \begin{overpic}[width=.49\textwidth,trim=0 0cm 0 0cm]{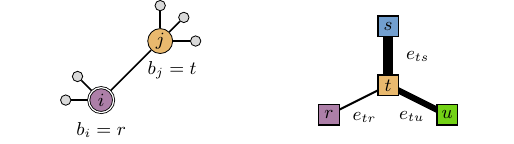}\put(0,0){(a)}\end{overpic}&
    \begin{overpic}[width=.47\textwidth,trim=0 0cm 0 0cm]{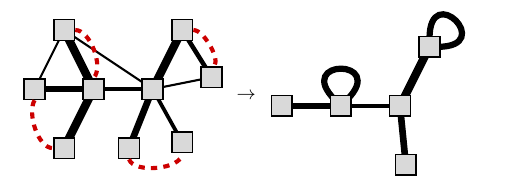}\put(0,0){(b)}\end{overpic}
  \end{tabular}

  \caption{Efficient MCMC strategies: (a) Move proposals are made by
  inspecting the neighborhood of node $i$ and selecting a random
  neighbor $j$. Based on its group membership $t=b_j$, the edge counts
  between groups are inspected (right), and the move proposal $b_i=s$ is
  made with probability proportional to $e_{ts}$. (b) The initial state
  of the MCMC is obtained with an agglomerative heuristic, where groups
  are merged together using the same proposals described in (a).\label{fig:mcmc}}
\end{figure}

Although fairly simple, the above algorithm suffers from some
shortcomings that can seriously degrade its performance in practice. In
fact, it is typical for naive implementations of the Metropolis-Hastings
algorithm to perform very badly, despite its theoretical
guarantees. This is because the asymptotic properties of the Markov
chain may take a very long time to be realized, and the equilibrium
distribution is never observed in practical time. Generally, we should
expect good convergence times only when: 1. The initial state $\bb_0$ is
close enough to the most likely states of the posterior and
2. the move proposals $P(\bb'|\bb)$ resemble the shape of the
posterior. Indeed, it is a trivial (and not very useful) fact that if
the starting state $\bb_o$ is sampled directly from the posterior, and
the move proposals match the posterior exactly, $P(\bb'|\bb) =
P(\bb'|\A)$, the Markov chain would be instantaneously
equilibrated. Hence if we can approach this ideal scenario, we should be
able to improve the inference speeds. Here we describe two simple
strategies in achieving such an improvement which have been shown to
yield a significance performance
impact~\cite{peixoto_efficient_2014}. The first one is to replace the
fully random move proposals of Eq.~\ref{eq:random-move} by a more
informative choice. Namely, we use the current information about the
model being inferred to guide our next move. We do so by selecting the
membership of a node $i$ being moved according to
\begin{equation}\label{eq:smart-move}
  P(b_i=r|\bb) = \sum_sP(s|i)\frac{e_{sr} + \epsilon}{e_s + \epsilon(B+1)},
\end{equation}
where $P(s|i) = \sum_jA_{ij}\delta_{b_j,s}/k_i$ is the fraction of
neighbors of node $i$ that belong to group $s$, and $\epsilon > 0$ is an
arbitrary parameter that enforces ergodicity, but with no other
significant impact in the algorithm, provided it is sufficiently small
(however, if $\epsilon \to \infty$ we recover the fully random moves of
Eq.~\ref{eq:random-move}). What this move proposal means is that we
inspect the local neighborhood of the node $i$, and see which groups
$s$ are connected to this node, and we use the typical neighborhood $r$
of the groups $s$ to guide our placement of node $i$ (see
Fig.~\ref{fig:mcmc}a). The purpose of these move proposals is not to
waste time with attempted moves that will almost surely be rejected, as
will typically happen with the fully random version. We emphasize that
the move proposals of Eq.~\ref{eq:smart-move} do not bias the partitions
toward any specific kind of mixing pattern; in particular they do not
prefer assortative versus non-assortative partitions. Furthermore, these
proposals can be generated efficiently, simply by following three steps:
1. sampling a random neighbor $j$ of node $i$, and inspecting its group
membership $s=b_j$, and then; 2. with probability $\epsilon(B+1)/(e_s
+ \epsilon(B+1))$ sampling a fully random group $r$ (which can be a
new group); 3. or otherwise, sampling a group label $r$ with a
probability proportional to the number of edges leading to it from group
$s$, $e_{sr}$. These steps can be performed in time $O(k_i)$, again
independently of $B$, as long as a continuous book-keeping is made of
the edges which are incident to each group, and therefore it does not
affect the overall $O(E)$ time complexity.

The second strategy is to choose a starting state that lies close to the
mode of the posterior. We do so by performing a Fibonacci
search~\cite{kiefer_sequential_1953} on the number of groups $B$, where
for each value we obtain the best partition from a larger partition with
$B'>B$ using an agglomerative heuristic, composed of the following steps
taken alternatively: 1. We attempt the moves of Eq.~\ref{eq:smart-move}
until no improvement to the posterior is observed, 2. We merge groups
together, achieving a smaller number of groups $B''\in[B,B']$, stopping
when $B''=B$. We do the last step by treating each group as a single
node and using Eq.~\ref{eq:smart-move} as a merge proposal, and
selecting the ones that least decrease the posterior (see
Fig~\ref{fig:mcmc}b). As shown in Ref.~\cite{peixoto_efficient_2014},
the overall complexity of this initialization algorithm is
$O(E\log^2N)$, and thus can be employed for very large networks.

The approach above can be adapted to the overlapping model of
Sec.~\ref{sec:overlap}, where instead of the partition $\bb$, the move
proposals are made with respect to the individual half-edge
labels~\cite{peixoto_model_2015}. For the nested model, we have instead
a hierarchical partition $\{b_l\}$, and we proceed in each step of the
Markov chain by randomly choosing a level $l$ and performing the
proposals of Eq.~\ref{eq:smart-move} on that level, as described in
Ref.~\cite{peixoto_nonparametric_2017}.

The combination of the two strategies described above makes the inference
procedure quite scalable, and has been successfully employed on networks
on the order of $10^7$ to $10^8$ edges, and up to $B=N$ groups. The
MCMC algorithm described in this section, for all model variants
described, is implemented in the \texttt{graph-tool}
library~\cite{peixoto_graph-tool_2014}, freely available under the GPL
license at \url{http://graph-tool.skewed.de}.

\section{To sample or to optimize?}\label{sec:sample}

In the examples so far, we have focused on obtaining the most likely
partition from the posterior distribution, which is the one that
minimizes the description length of the data. But is this in fact the
best approach? In order to answer this, we need first to quantify how
well our inference is doing, by comparing our estimate $\hat\bb$ of the
partition to the true partition that generated the data $\bb^*$, by
defining a so-called \emph{loss function}. For example, if we choose to
be very strict, we may reject any partition that is strictly different
from $\bb^*$ on equal measure, using the indicator function
\begin{equation}
  \Delta(\hat\bb,\bb^*) = \prod_i\delta_{\hat b_i,b_i^*},
\end{equation}
so that $\Delta(\hat\bb,\bb^*)=1$ only if $\hat\bb=\bb^*$, otherwise
$\Delta(\hat\bb,\bb^*)=0$. If the observed data $\A$ and parameters $\bb$ are
truly sampled from the model and priors, respectively, the best
assessment we can make for $\bb^*$ is given by the posterior distribution
$P(\bb|\A)$. Therefore, the average of the indicator over the posterior is
given by
\begin{equation}
  \bar{\Delta}(\hat\bb) = \sum_{\bb}\Delta(\hat\bb,\bb)P(\bb|\A).
\end{equation}
If we maximize $\bar{\Delta}(\hat\bb)$ with respect to $\hat\bb$, we
obtain the so-called maximum \emph{a posteriori} (MAP) estimator
\begin{equation}
  \hat\bb = \underset{\bb}{\operatorname{argmax}}\, P(\bb|\A),
\end{equation}
which is precisely what we have been using so far, and it is equivalent
to employing the MDL principle. However, using this estimator is
arguably overly optimistic, as we are unlikely to find the true
partition with perfect accuracy in any but the most ideal
cases. Instead, we may relax our expectations and consider instead the
overlap function
\begin{equation}
  d(\hat\bb,\bb^*) = \frac{1}{N}\sum_i\delta_{\hat b_i,b_i^*},
\end{equation}
which measures the \emph{fraction} of nodes that are correctly
classified. If we maximize now the average of the overlap over the
posterior distribution
\begin{equation}
  \bar{d}(\hat\bb) = \sum_{\bb}d(\hat\bb,\bb)P(\bb|\A),
\end{equation}
we obtain the \emph{marginal estimator}
\begin{equation}
  \hat{b}_i = \underset{r}{\operatorname{argmax}}\, \pi_i(r),
\end{equation}
where
\begin{equation}
  \pi_i(r) = \sum_{\bb\setminus b_i}P(b_i=r,\bb\setminus b_i|\A)
\end{equation}
is the marginal distribution of the group membership of node $i$, summed
over all remaining nodes.\footnote{The careful reader will notice that
we must have in fact a trivial constant marginal $\pi_i(r)=1/B$ for
every node $i$, since there is a symmetry of the posterior distribution
with respect to re-labelling of the groups, in principle rendering this
estimator useless. In practice, however, our samples from the posterior
distribution (e.g. using MCMC) will not span the whole space of label
permutations in any reasonable amount of time, and instead will
concentrate on a mode around one of the possible permutations. Since the
modes around the label permutations are entirely symmetric, the node
marginals obtained in this manner can be meaningfully used. However, for
networks where some of the groups are not very large, \emph{local}
permutations of individual group labels are statistically possible
during MCMC inference, leading to degeneracies in the marginal
$\pi_i(r)$ of the affected nodes, resulting in artefacts when using the
marginal estimator. This problem is exacerbated when the number of
groups changes during MCMC sampling.} The marginal estimator is notably
different from the MAP estimator in that it leverages information from
the entire posterior distribution to inform the classification of any
single node. If the posterior is tightly concentrated around its
maximum, both estimators will yield compatible answers. In this
situation the structure in the data is very clear, and both estimators
agree. Otherwise, the estimators will yield different aspects of the
data, in particular if the posterior possesses many local maxima. For
example, if the data has indeed been sampled from the model we are
using, the multiplicity of local maxima can be just a reflection of the
randomness in the data, and the marginal estimator will be able to
average over them and provide better
accuracy~\cite{moore_computer_2017,zdeborova_statistical_2016}.

\begin{figure}[t!]
  \begin{tabular}{ccc}
    \begin{overpic}[width=.33\textwidth,trim=0 0 0 0]{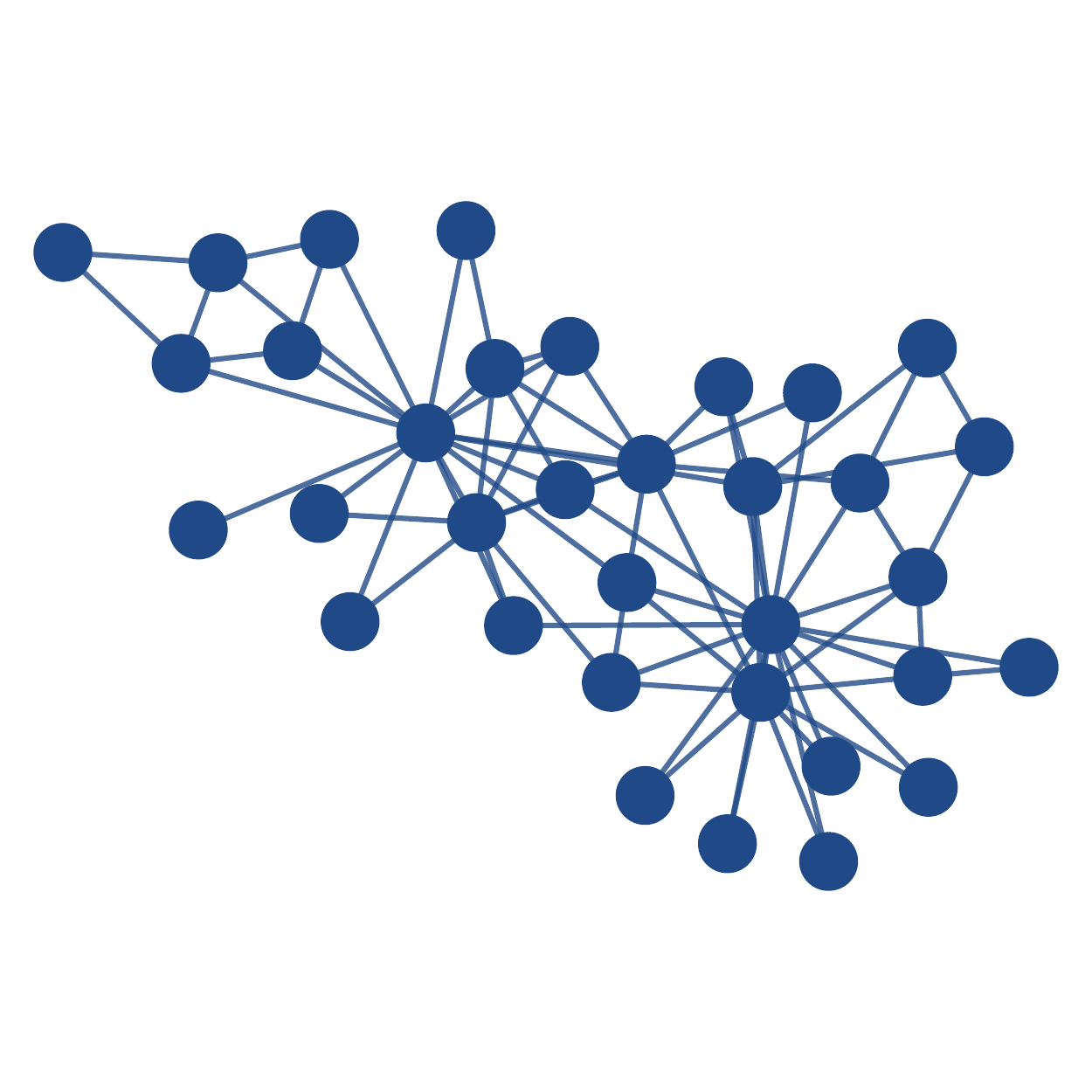}\put(20,0){(a) $\bb_0$, $\Sigma=321.3$ bits}\end{overpic}&
    \begin{overpic}[width=.33\textwidth,trim=0 0 0 0]{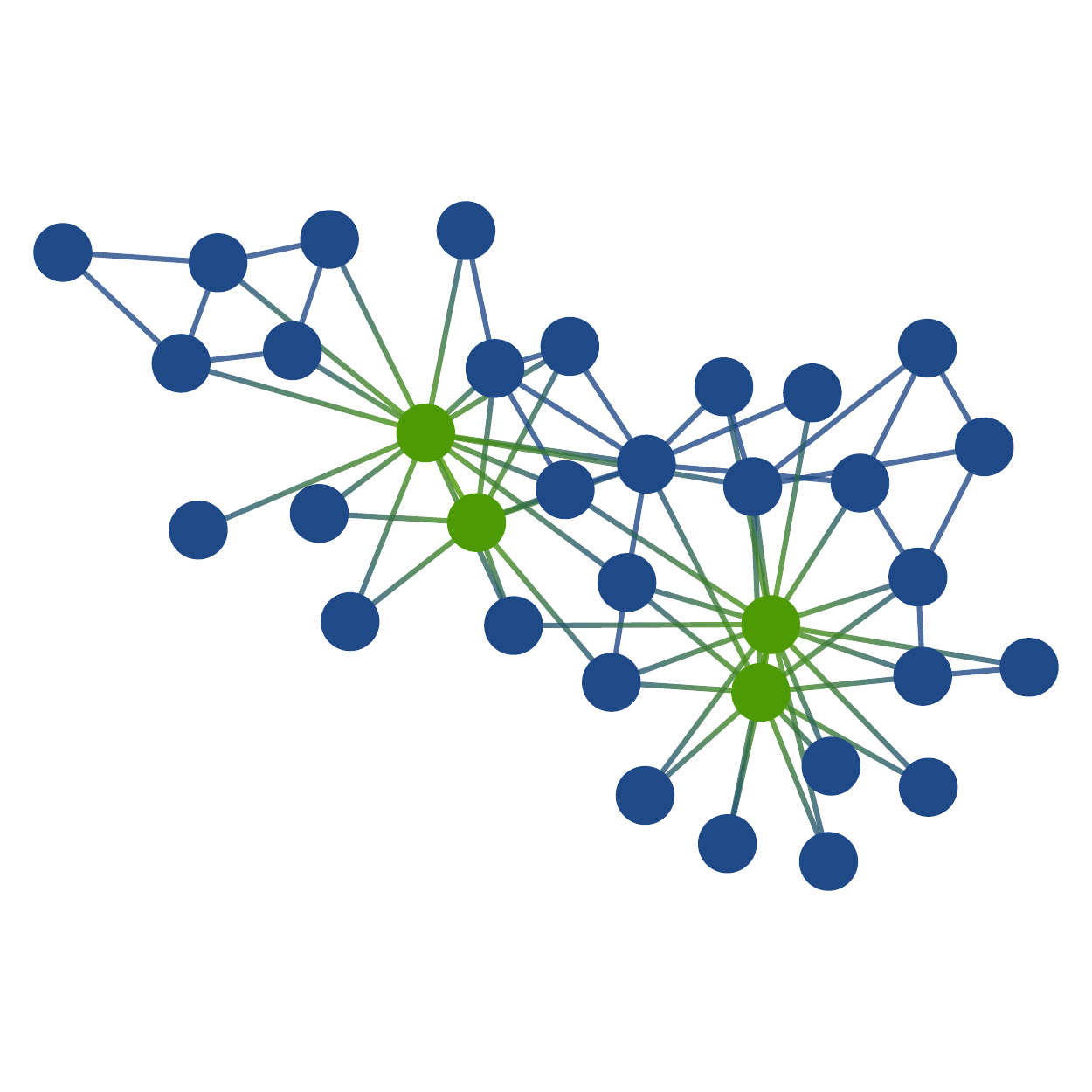}\put(20,0){(b) $\bb_1$, $\Sigma=327.5$ bits}\end{overpic}&
    \begin{overpic}[width=.33\textwidth,trim=0 0 0 0]{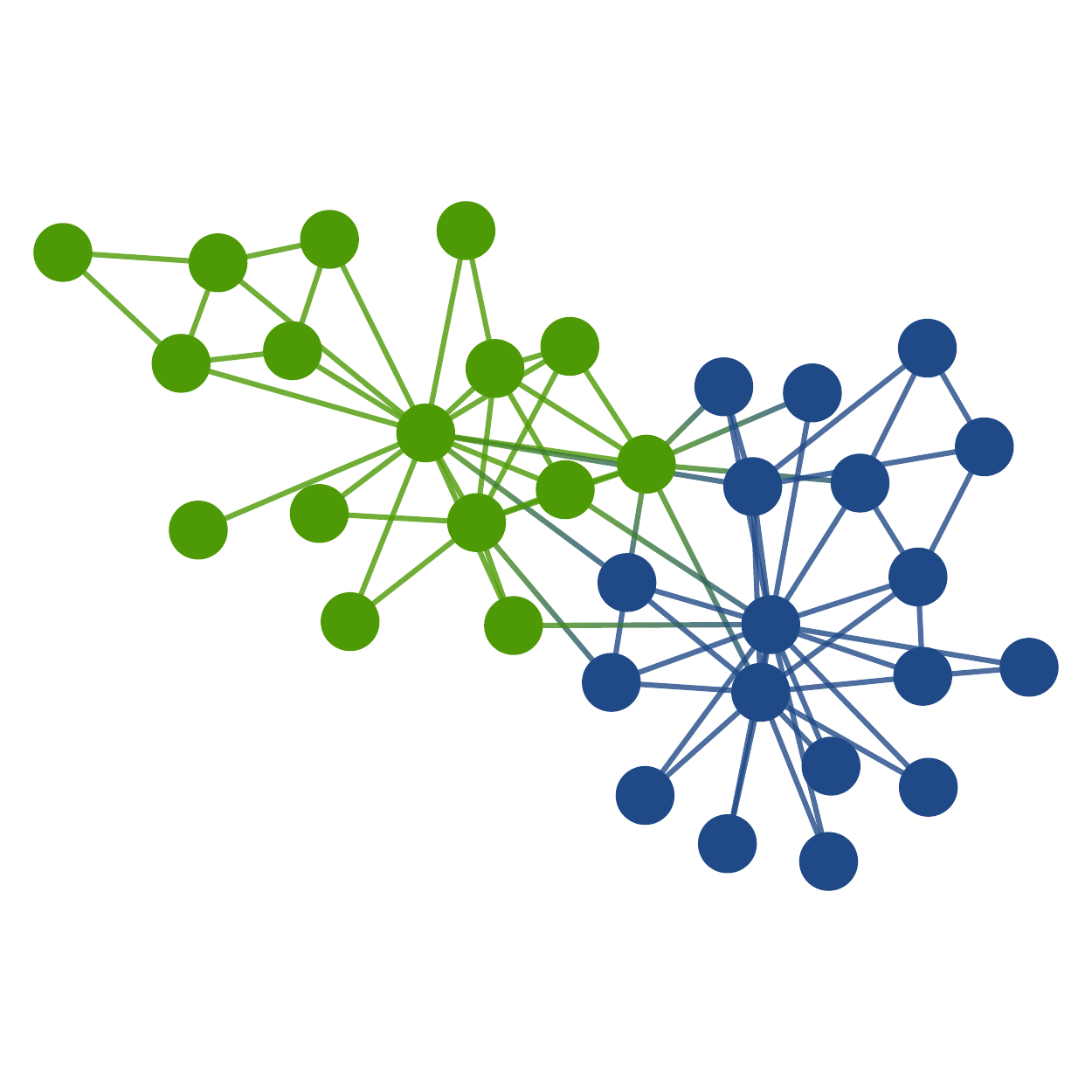}\put(20,0){(c) $\bb_2$, $\Sigma=329.3$ bits}\end{overpic}\\
  \end{tabular}
  \begin{tabular}{cc}
    \begin{overpic}[width=.7\textwidth,trim=.2cm 1.1cm 1.6cm 1.2cm]{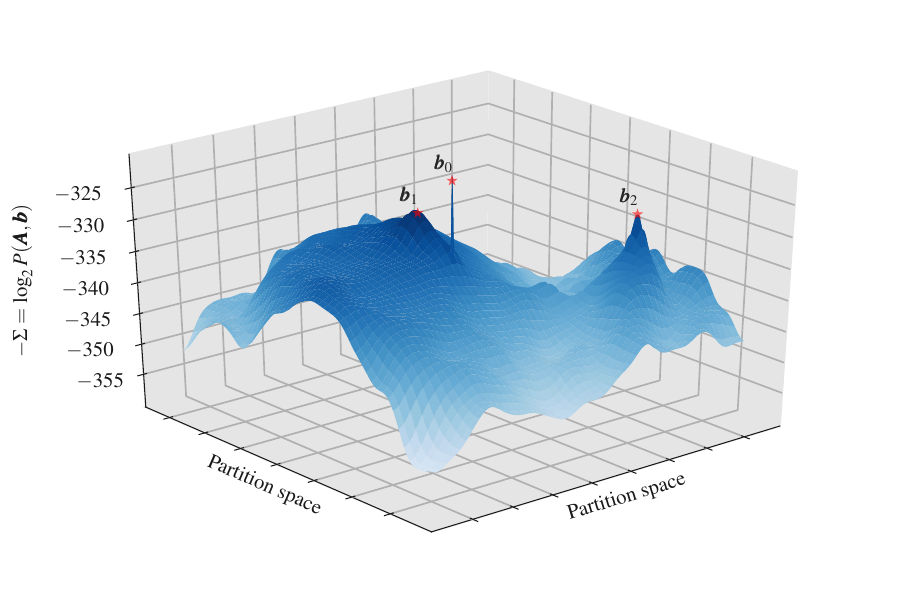}\put(0,0){(d)}\end{overpic}&
    \begin{overpic}[width=.29\textwidth,trim=.3cm .3cm 0 0,clip]{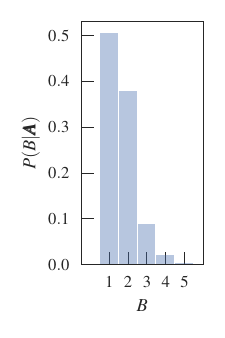}\put(0,0){(e)}\end{overpic}
  \end{tabular}

  \caption{Posterior distribution of partitions of Zachary's karate club
  network using the DC-SBM. Panels (a) to (c) show three modes of the
  distribution and their respective description lengths; (d) 2D
  projection of the posterior obtained using multidimensional
  scaling~\cite{cox_multidimensional_2000}; (e) Marginal posterior
  distribution of the number of groups $B$. \label{fig:karate}}
\end{figure}
In view of the above, one could argue that the marginal estimator should
be generally preferred over MAP. However, the situation is more
complicated for data which are not sampled from model being used for
inference (i.e. the model is \emph{misspecified}). In this situation,
multiple peaks of the distribution can point to very different
partitions that are all statistically significant. These different peaks
function as alternative explanations for the data that must be accepted
on equal footing, according to their posterior probability. The marginal
estimator will in general mix the properties of all peaks into a
consensus classification that is not representative of any single
hypothesis, whereas the MAP estimator will concentrate only on the most
likely one (or an arbitrary choice if they are all equally likely). An
illustration of this is given by the well-known Zachary's karate club
network~\cite{zachary_information_1977}\index{author}{Wayne W. Zachary},
which captures the social interactions between members of a karate club
amidst a conflict between the club's administrator and an instructor,
which lead to a split of the club in two disjoint groups. The
measurement of the network was done before the final split actually
happened, and it is very often used as an example of a network
exhibiting community structure. If we analyze this network with the
DC-SBM, we obtain three partitions that occur with very high probability
from the posterior distribution: A trivial $B=1$ partition,
corresponding to the configuration model without communities
(Fig.~\ref{fig:karate}a), a ``leader-follower'' division into $B=2$
groups, separating the administrator and instructor, together with two
close allies, from the rest of the network (Fig.~\ref{fig:karate}b), and
finally a $B=2$ division into the aforementioned factions that
anticipated the split (Fig.~\ref{fig:karate}c). If we would guide
ourselves strictly by the MDL principle (i.e. using the MAP estimator),
the preferred partition would be the trivial $B=1$ one, indicating that
the most likely explanation of this network is a fully random graph with
a pre-specified degree sequence, and that the observed community
structure emerged spontaneously. However, if we inspect the posterior
distribution more closely, we see that other divisions into $B>1$ groups
amount to around $50\%$ of the posterior probability (see
Fig.~\ref{fig:karate}e). Therefore, if we consider all $B>1$ partitions
\emph{collectively}, they give us little reason to completely discard
the possibility that the network does in fact posses some group
structure. Inspecting the posterior distribution even more closely, as
shown in Fig.~\ref{fig:karate}d, reveals a multimodal structure
clustered around the three aforementioned partitions, giving us three
very different explanations for the data, none of which can be
decisively discarded in favor of the others --- at least not according
to the evidence available in the network structure alone.

The situation encountered for the karate club network is a good example
of the so-called \emph{bias-variance
trade-off}\index{topic}{bias-variance trade-off} that we are often
forced to face: If we choose to single-out a singe partition as a unique
representation of the data, we must invariably \emph{bias} our result
toward any of the three most likely scenarios, discarding the remaining
ones at some loss of useful information. Otherwise, if we choose to
eliminate the bias by incorporating the entire posterior distribution in
our representation, by the same token it will incorporate a larger
variance,
i.e. it will simultaneously encompass diverging explanations of the
data, leaving us without an unambiguous and clear interpretation.  The
only situation where this trade-off is not required is when the model is
a perfect fit to the data, such that the posterior is tightly peaked
around a single partition. Therefore, the variance of the posterior
serves as a good indication of the quality of fit of the model,
providing another reason to include it in the analysis.

\begin{figure}
  \begin{minipage}{.36\textwidth}
    \begin{overpic}[width=\textwidth]{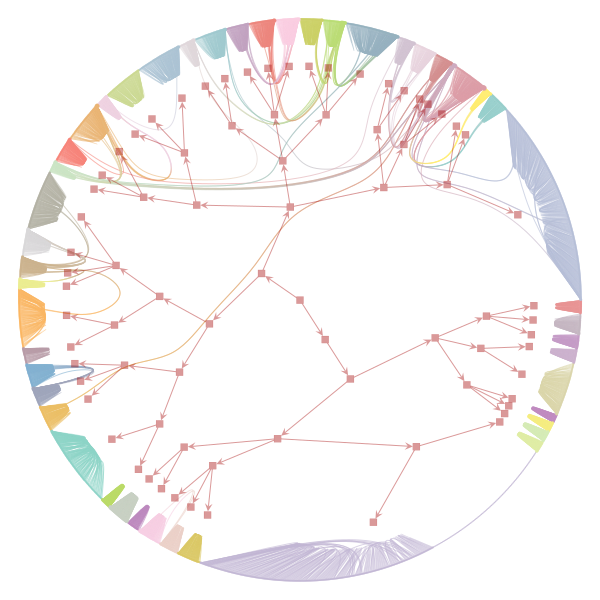}
      \put(0,0){(a)}
    \end{overpic}
  \end{minipage}
  \begin{minipage}{.52\textwidth}
    \begin{tabular}{ccc}
    \begin{overpic}[width=.33\textwidth]{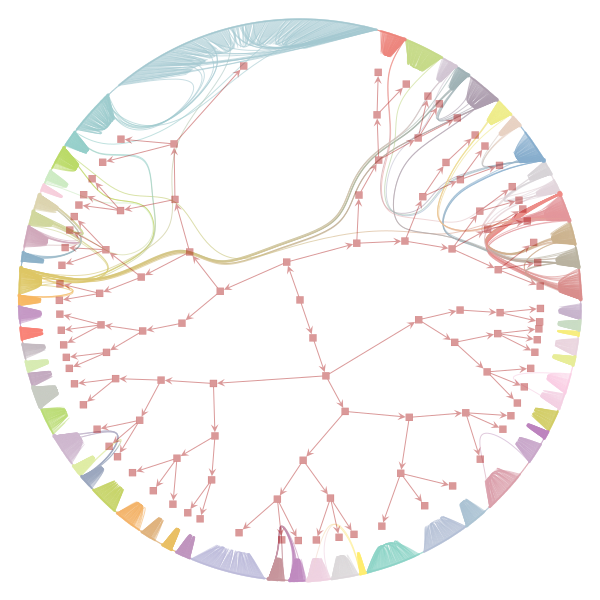}
    \end{overpic}&
    \begin{overpic}[width=.33\textwidth]{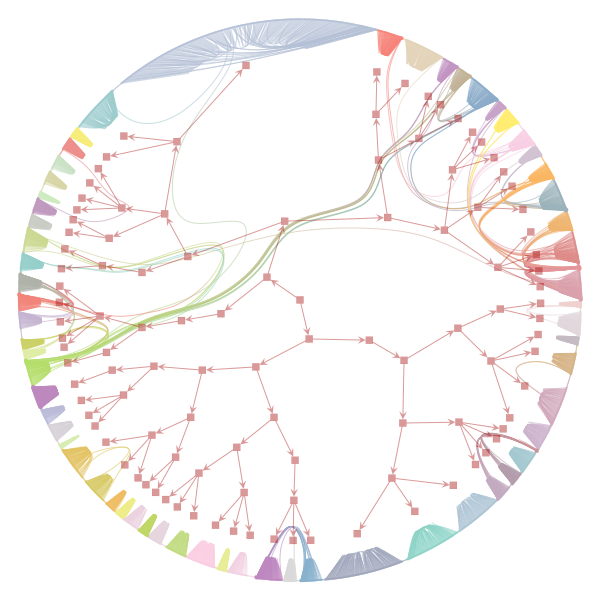}
    \end{overpic}&
    \begin{overpic}[width=.33\textwidth]{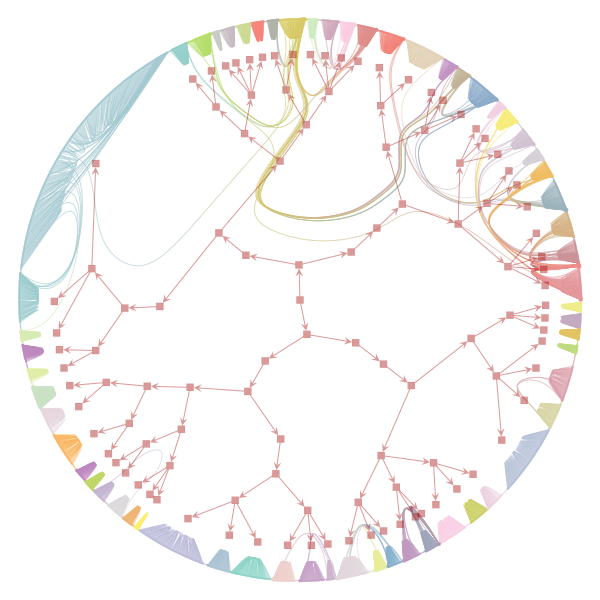}
    \end{overpic}\\
    \begin{overpic}[width=.33\textwidth]{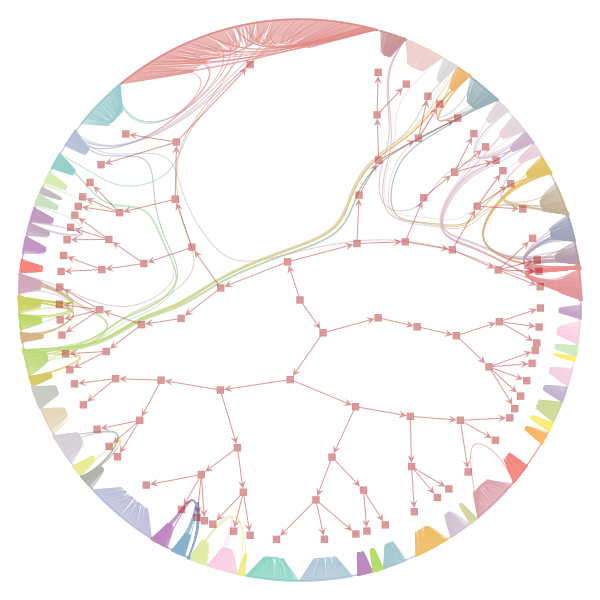}
      \put(0,0){(b)}
    \end{overpic}&
    \begin{overpic}[width=.33\textwidth]{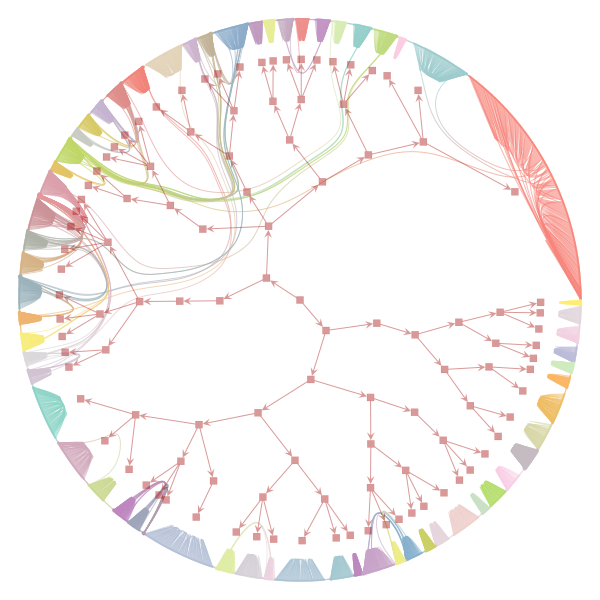}
    \end{overpic}&
    \begin{overpic}[width=.33\textwidth]{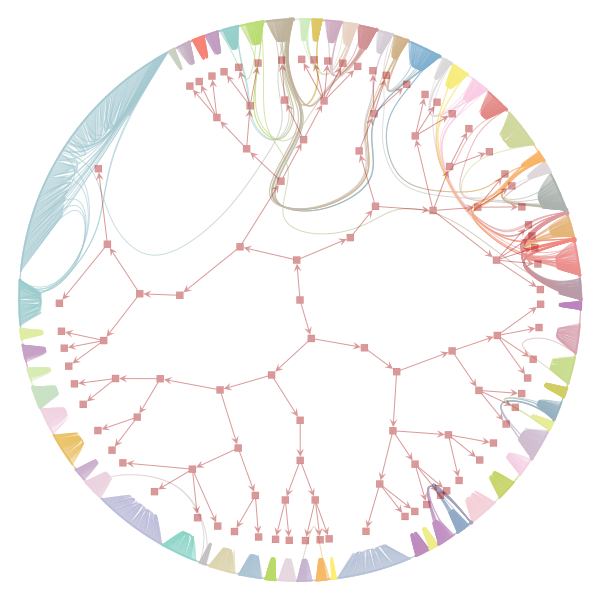}
    \end{overpic}
    \end{tabular}
  \end{minipage}

  \vspace{.8em}
  \begin{minipage}{\textwidth}
    \begin{overpic}[width=\textwidth]{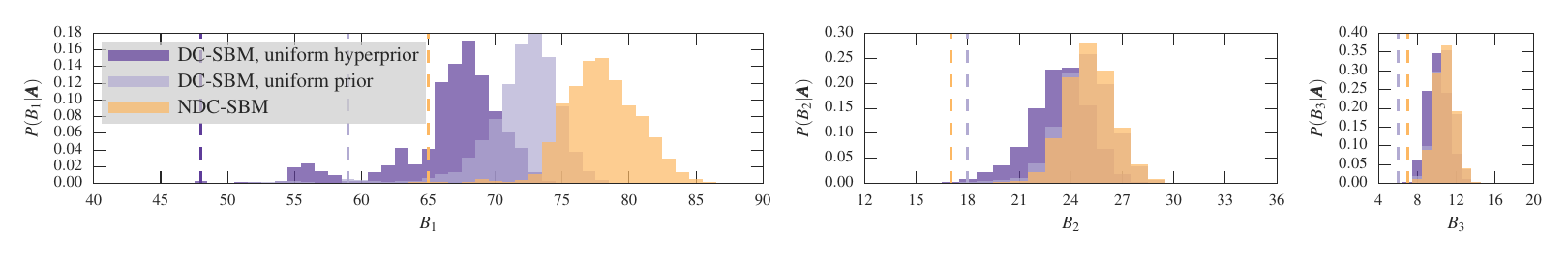}\put(0,0){(c)}\end{overpic}
  \end{minipage}

  \caption{Hierarchical partitions of a network of collaboration between
  scientists~\cite{newman_finding_2006}. (a) Most likely hierarchical
  partition according to the DC-SBM with a uniform hyperprior. (b)
  Uncorrelated samples from the posterior distribution. (c) Marginal
  posterior distribution of the number of groups at the first three
  hierarchical levels, according to the model variants described in the
  legend. The vertical lines mark the value obtained for the most likely
  partition. Adapted from
  Ref.~\cite{peixoto_nonparametric_2017}.\label{fig:netscience}}
\end{figure}

It should also be remarked that when using a nonparametric approach,
where the dimension of the model is also inferred from the posterior
distribution, the potential bias incurred when obtaining only the most
likely partition usually amounts to an \emph{underfit} of the data,
since the uncertainty in the posterior typically translates into the
existence of a more conservative partition with fewer
groups.\footnote{This is different from \emph{parametric} posteriors,
where the dimension of the model is externally imposed in the prior, and
the MAP estimator tends to
overfit~\cite{moore_computer_2017,zdeborova_statistical_2016}.}
Instead, if we sample from the posterior distribution, we will average
over may alternative fits, including those that model the data more
closely with a larger number of groups. However, each individual sample
of the posterior will tend to incorporate more randomness from the data,
which will disappear only if we average over all samples. This means
that single samples will tend to overfit the data, and hence we must
resist looking at them individually. It is only in the aforementioned
limit of a perfect fit that we are guaranteed not to be misled one way
or another. An additional example of this is shown in
Fig.~\ref{fig:netscience} for a network of collaborations among
scientists. If we infer the best nested SBM, we find a specific
hierarchical division of the network. However, if we sample hierarchical
divisions from the posterior distribution, we typically encounter larger
models --- with a larger number of groups and deeper hierarchy. Each
individual sample from the posterior is likely to be an overfit, but
collectively they give a more accurate picture of the network in
comparison with the most likely partition, which probably
over-simplifies it. As already mentioned, this discrepancy, observed for
all three SBM versions, tells us that neither of them is an ideal fit
for this network.

The final decision on which approach to take depends on the actual
objective and resources available. In general, sampling from the
posterior will be more suitable when the objective is to generalize from
observation and make predictions (see next section and
Ref.~\cite{valles-catala_consistency_2017}), and when computational
resources are ample. Conversely, if the objective is to make a precise
statement about the data, e.g. in order to summarize and interpret it,
and the computational resources are scarce, maximizing the posterior
tends to be more adequate.

\section{Generalization and prediction}\index{topic}{prediction!missing links}\index{topic}{prediction!spurious links}

When we fit a model like the SBM to a network, we are doing more than
simply dividing the nodes into statistically equivalent groups; we are
also making a statement about a possible mechanism that generated the
network. This means that, to the extent that the model is a good
representation of the data, we can use it generalize and make
predictions about what has \emph{not} been observed. This has been most
explored for the prediction of missing and spurious
links~\cite{clauset_hierarchical_2008, guimera_missing_2009}. This
represents the situation where we know or stipulate that the observed
data is noisy, and may contain edges that in fact do not exist, or does
not contain edges that do exist. With a generative model like the SBM,
we are able to ascribe probabilities to existing and non-existing edges
of being spurious or missing, respectively, as we now describe.

Following Ref.~\cite{valles-catala_consistency_2017}, the scenario we will
consider is the situation where there exists a complete network $\G$
which is decomposed in two parts,
\begin{equation}\label{eq:G_composition}
  \G = \Ao + \dA
\end{equation}
where $\Ao$ is the network that we observe, and the $\dA$ is the set of
missing and spurious edges that we want to predict, where an entry
$\delta A_{ij} > 0$ represents a missing edge, and $\delta A_{ij}< 0$ a
spurious one. Hence, our task is to obtain the posterior distribution
\begin{equation}\label{eq:dA_posterior}
  P(\dA|\Ao).
\end{equation}
The central assumption we will make is that the complete network $\G$
has been generated using some arbitrary version of the SBM, with a marginal
distribution
\begin{equation}
  P_G(\G|\bb).
\end{equation}
Given a generated network $\G$, we then select $\dA$ from some arbitrary
distribution that models our source of errors
\begin{align}\label{eq:PdA}
  P_{\delta A}(\dA|\G).
\end{align}
With the above model for the generation of the complete network and its
missing and spurious edges, we can proceed to compute the posterior of
Eq.~\ref{eq:dA_posterior}. We start from the joint distribution
\begin{align}
  P(\Ao,\dA|\G) &= P(\Ao|\dA, \G) P_{\delta A}(\dA|\G)\\
  &=\delta(\G - (\Ao + \dA))P_{\delta A}(\dA|\G),
\end{align}
where we have used the fact $P(\Ao|\dA, \G) = \delta(\G - (\Ao + \dA))$
originating from Eq.~\ref{eq:G_composition}.  For the joint distribution
conditioned on the partition, we sum the above over all possible graphs
$\G$, sampled from our original model,
\begin{align}
  P(\Ao,\dA|\bb) &= \sum_{\G}P(\Ao,\dA|\G)P_G(\G|\bb)\\
  &= P_{\delta A}(\dA|\Ao+\dA)P_G(\Ao+\dA|\bb).
\end{align}
The final posterior distribution of Eq.~\ref{eq:dA_posterior} is
therefore
\begin{align}
  P(\dA|\Ao) &= \frac{\sum_{\bb}P(\Ao,\dA|\bb)P(\bb)}{P(\Ao)} \\
  &= \frac{P_{\delta A}(\dA|\Ao+\dA)\sum_{\bb}P_G(\Ao+\dA|\bb)P(\bb)}{P(\Ao)},
\end{align}
with $P(\Ao)$ being a normalization constant, independent of $\dA$. This
expression gives a general recipe to compute the posterior, where one
averages the marginal likelihood $P_G(\Ao+\dA|\bb)$ obtained by sampling
partitions from the prior $P(\bb)$. However, this procedure will
typically take an astronomical time to converge to the correct
asymptotic value, since the largest values of $P_G(\Ao + \dA|\bb)$ will
be far away from most values of $\bb$ sampled from $P(\bb)$. A much
better approach is to perform importance sampling, by rewriting the
posterior as
\begin{align}
  P(\dA|\Ao) &\propto P_{\delta A}(\dA|\Ao+\dA)\sum_{\bb}P_G(\Ao+\dA|\bb)\frac{P_G(\Ao|\bb)}{P_G(\Ao|\bb)}P(\bb)\\
  &\propto P_{\delta A}(\dA|\Ao+\dA)\sum_{\bb}\frac{P_G(\Ao+\dA|\bb)}{P_G(\Ao|\bb)}P_G(\bb|\Ao),\label{eq:dA_posterior_importance}
\end{align}
where $P_G(\bb|\Ao)$ is the posterior of partitions obtained by
pretending that the observed network came directly from the SBM. We can
sample from this posterior using MCMC as described in
Sec.~\ref{sec:mcmc}. As the number of entries in $\dA$ is typically much
smaller than the number of observed edges, this importance sampling
approach will tend to converge much faster. This allows us to compute
$P(\dA|\Ao)$ in practical manner
--- up to a normalization constant. However, if we want to compare the
relative probability between specific sets of missing/spurious edges,
$\{\dA_i\}$, via the ratio
\begin{equation}\label{eq:missing_ratio}
  \lambda_i = \frac{P(\dA_i|\Ao)}{\sum_jP(\dA_j|\Ao)},
\end{equation}
this normalization constant plays no role. The above still depends on
our chosen model for the production of missing and spurious edges, given
by Eq.~\ref{eq:PdA}. In the absence of domain-specific information about
the source of noise, we must consider all alternative choices
$\{\dA_i\}$ to be equally likely a priori, so that the we can simply
replace $P_{\delta A}(\dA|\Ao+\dA) \propto 1$ in
Eq.~\ref{eq:dA_posterior_importance} --- although more realistic choices
can also be included.

\begin{figure}\centering
  \begin{overpic}[width=.6\textwidth]{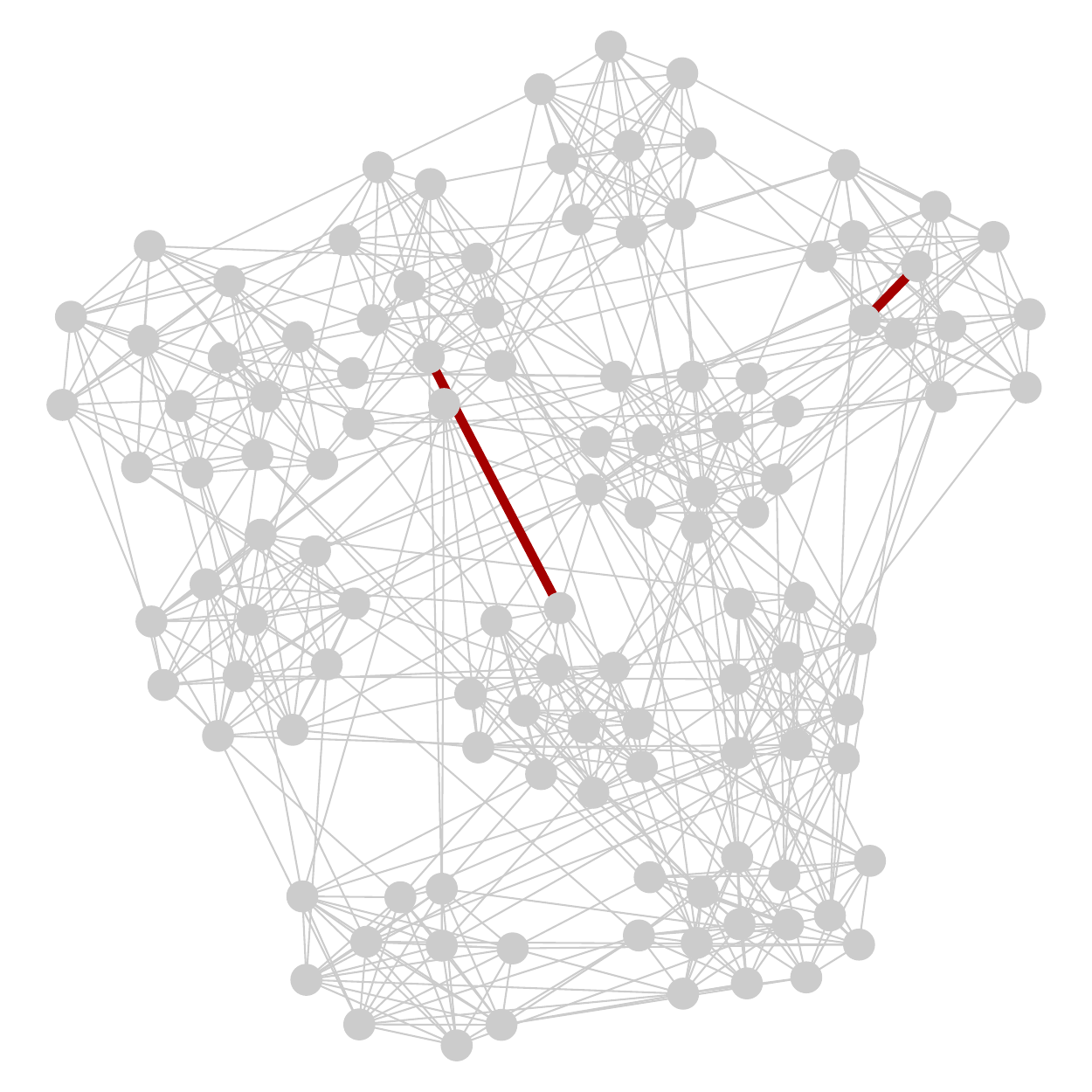}
    \put(37,52){(a)}
    \put(70,70){(b)}
  \end{overpic} \caption{Two hypothetical missing edges in the network
  of American college football teams. The edge $(a)$ connects teams of
  different conferences, whereas $(b)$ connects teams of the same
  conference. According to the nested DC-SBM, their posterior
  probability ratios are $\lambda_a\approx 0.013(1)$ and
  $\lambda_b\approx 0.987(1)$.\label{fig:football-missing}}
\end{figure}

In Fig.~\ref{fig:football-missing} we show the relative probabilities of
two hypothetical missing edges for the American college football
network, obtained with the approach above. We see that a particular
missing edge between teams of the same conference is almost a hundred
times more likely than one between teams of different conference.

The use of the SBM to predict missing and spurious edges has been
employed in a variety of applications, such as the prediction of novel
interactions between drugs~\cite{guimera_network_2013}, conflicts in
social networks~\cite{rovira-asenjo_predicting_2013}, as well to provide
user
recommendations~\cite{guimera_predicting_2012,godoy-lorite_accurate_2016},
and in many cases has outperformed a variety of competing methods.

\section{Fundamental limits of inference: the detectability-indetectability phase transition}

Besides defining useful models and investigating their behavior in data,
there is another line of questioning which deals with how far it is
possible to go when we try to infer the structure of
networks. Naturally, the quality of the inference depends on the
statistical evidence available in the data, and we may therefore ask if
it is possible at all to uncover \emph{planted} structures ---
i.e. structures that we impose ourselves --- with our inference methods,
and if so, what is the best performance we can expect. Research in this
area has exploded in recent
years~\cite{zdeborova_statistical_2016,moore_computer_2017}, after it
was shown by Decelle et
al~\cite{decelle_inference_2011,decelle_asymptotic_2011}\index{author}{Aurélien
Decelle}\index{author}{Florent Krzakala}\index{author}{Cristopher
Moore}\index{author}{Lenka Zdeborová} that not only it may be impossible
to uncover planted structures with the SBM, but the inference undergoes
a ``phase transition'' where it becomes possible only if the structure
is strong enough to cross a non-trivial threshold. This result was
obtained using methods from statistical physics, which we now describe.

The situation we will consider is a ``best case scenario,'' where all
parameters of the model are known, with the exception of the partition
$\bb$ --- this in contrast to our overall approach so far, where we
considered all parameters to be unknown random variables. In particular,
we will consider only the prior
\begin{equation}
  P(\bb|\bm{\gamma}) = \prod_i\gamma_{b_i}.
\end{equation}
where $\gamma_r$ is the probability of a node belonging in group
$r$. Given this, we wish to obtain the posterior distribution of the
node partition, using the SBM of Eq.~\ref{eq:poisson-sbm},
\begin{equation}
  P(\bb|\A,\bm{\lambda},\bm{\gamma}) =
  \frac{P(\A|\bb,\bm{\lambda})P(\bb|\bm{\gamma})}{P(\A|\bm{\lambda},\bm{\gamma}).}
  = \frac{e^{-\mathcal{H}(\bb)}}{Z}
\end{equation}
which was written above in terms of the ``Hamiltonian''
\begin{equation}
  \mathcal{H}(\bb) = -\sum_{i<j}(A_{ij}\ln\lambda_{b_i,b_j}-\lambda_{b_i,b_j}) - \sum_i\ln\gamma_{b_i},
\end{equation}
drawing an analogy with Potts-like models in statistical
physics~\cite{wu_potts_1982}. The normalization constant, called the
``partition function,'' is given by
\begin{equation}
  Z = \sum_{\bb}e^{-\mathcal{H}(\bb)}.
\end{equation}
Far from being an unimportant detail, the partition function can be used
to determine all statistical properties of our inference procedure. For
example, if we wish to obtain the marginal posterior distribution of
node $i$, we can do so by introducing the perturbation
$\mathcal{H}'(\bb)=\mathcal{H}(\bb)-\mu\delta_{b_i,r}$ and computing the
derivative
\begin{equation}
  P(b_i=r|\A,\bm{\lambda},\bm{\gamma})=\left.\frac{\partial\ln Z}{\partial\mu}\right|_{\mu=0}=\sum_{\bb}\delta_{b_i,r}\frac{e^{-\mathcal{H}(\bb)}}{Z}.
\end{equation}
Unfortunately, it does not seem possible to compute the partition
function $Z$ in closed form for an arbitrary graph $\A$. However, there
is a special case for which we \emph{can} compute the partition
function, namely when $\A$ is a \emph{tree}. This is useful for us,
because graphs sampled from the SBM will be ``locally
tree-like''\index{topic}{locally tree-like} if they are sparse (i.e. the
degrees are small compared to the size of the network $k_i\ll N$), and
the group sizes scale with the size of the system,
i.e. $n_r = O(N)$ (which implies $B\ll N$). Locally tree-like means that
typical loops will have length $O(N)$, and hence at the immediate
neighborhood of any given node the graph will look like a
tree. Although being locally tree-like is not quite the same as being a
tree, the graph will become increasing \emph{closer} to being a tree in
the ``thermodynamic limit'' $N\to\infty$. Because of this, many
properties of locally tree-like graphs will become asymptotically
identical to trees in this limit. If we assume that this limit holds, we
can compute the partition function by pretending that the graph is close
enough to being a tree, in which case we can write the so-called Bethe
free energy (we refer to
Refs.~\cite{mezard_information_2009,decelle_asymptotic_2011} for a
detailed derivation)
\begin{equation}
  \mathcal{F} = -\ln Z = -\sum_i \ln Z^i + \sum_{i<j}A_{ij}\ln Z^{ij} - E
\end{equation}
with the auxiliary quantities given by
\begin{align}
Z^{ij} &= N\sum_{r<s}\lambda_{rs}(\psi_r^{i\to j}\psi_s^{j\to i} + \psi_s^{i\to j}\psi_r^{j\to i}) + N\sum_r\lambda_{rr}\psi_r^{i\to j}\psi_r^{j\to i}\\
Z^i &= \sum_rn_re^{-h_r}\prod_{j\in\partial i}\sum_rN\lambda_{rb_i}\psi_r^{j\to i},
\end{align}
where $\partial i$ means the neighbors of node $i$.  In the above
equations, the values $\psi_r^{i\to j}$ are called ``messages,'' and
they must fulfill the self-consistency equations
\begin{equation}\label{eq:bp}
  \psi_r^{i\to j} = \frac{1}{Z^{i\to j}} \gamma_r e^{-h_r}\prod_{k\in\partial i\setminus j}\left(\sum_sN\lambda_{rs}\psi_s^{k\to i}\right)
\end{equation}
where $k\in\partial i\setminus j$ means all neighbors $k$ of $i$
excluding $j$, the value $Z^{i\to j}$ is a normalization constant
enforcing $\sum_r\psi_r^{i\to j}=1$, and $h_r =
\sum_i\sum_r\lambda_{rb_i}\psi_r^i$ is a local auxiliary
field. Eqs.~\ref{eq:bp} are called the {\bf belief-propagation} (BP)
equations~\cite{mezard_information_2009}\index{topic}{belief
propagation}, and the entire approach is also known under the name
``cavity method''~\cite{mezard_spin_1986}\index{topic}{cavity
method}. The values of the messages are typically obtained by iteration,
where we start from some initial configuration (e.g. a random one),
and compute new values from the right-hand side of Eq.~\ref{eq:bp},
until they converge asymptotically. Note that the messages are
only defined on edges of the network, and an update involves inspecting
the values at the neighborhood of the nodes, where the messages can be
interpreted as carrying information about the marginal distribution of a
given node, if the same is removed from the network (hence the names
``belief propagation'' and ``cavity method''). Each iteration of the BP
equations can be done in time $O(EB^2)$, and the convergence is often
obtained only after a few iterations, rendering the whole computation
fairly efficient, provided $B$ is reasonably small. After the messages
have been obtained, they can be used to compute the node marginals,
\begin{equation}\label{eq:bp_marginal}
  P(b_i=r|\A,\bm{\lambda},\bm{\gamma})=\psi_r^i=\frac{1}{Z^i}\gamma_r\prod_{j\in\partial i}\left[\sum_s\left(N\lambda_{rs}\right)^{A_{ij}}e^{-\lambda_{rs}}\psi_s^{j\to i}\right],
\end{equation}
where $Z^i$ is a normalization constant.

\begin{figure}\centering
  \begin{overpic}[width=.75\textwidth]{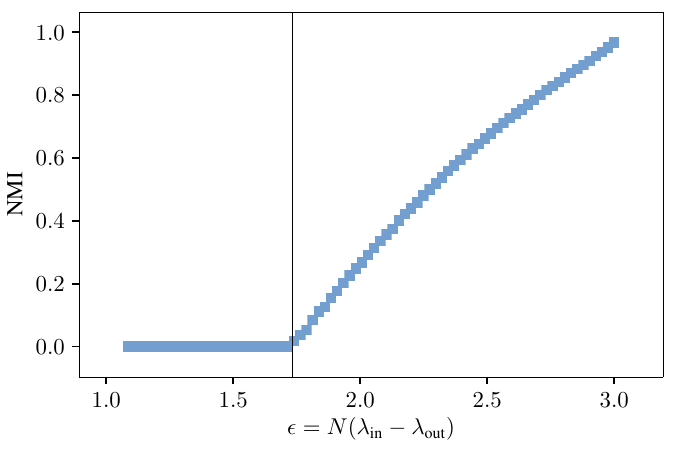}\end{overpic}
  \caption{Normalized mutual information (NMI) between the planted and
  inferred partitions of a PP model with $N=10^5$, $B=3$ and $\avg{k}=3$
  and $\epsilon=N(\lambda_{\text{in}}-\lambda_{\text{out}})$. The
  vertical line marks the detectability threshold $\epsilon^* =
  B\sqrt{\avg{k}}$.\label{fig:bp}}
\end{figure}

This whole procedure gives a way of computing the marginal distribution
$P(b_i=r|\A,\bm{\lambda},\bm{\gamma})$ in a manner that is
asymptotically exact --- if $\A$ is sufficiently large and locally
tree-like. Since networks that are sampled from the SBM fulfill this
property\footnote{Real networks, however, should not be expected to be
locally tree-like. This does not invalidate the results of this section,
which pertain strictly to data sampled from the SBM. However, despite
not being exact, the BP algorithm yields surprisingly accurate results
for real networks, even when the tree-like property is
violated~\cite{decelle_asymptotic_2011}.}, we may proceed with our
original question, and test if we can recover the true value of $\bb$ we
used to generate a network. For the test, we use a simple
parametrization named the planted partition model
(PP)~\cite{dyer_solution_1989,condon_algorithms_2001}, where
$\gamma_r=1/B$ and
\begin{equation}
  \lambda_{rs} = \lambda_{\text{in}}\delta_{rs} + \lambda_{\text{out}}(1-\delta_{rs}),
\end{equation}
with $\lambda_{\text{in}}$ and $\lambda_{\text{out}}$ specifying the
expected number of edges between nodes of the same groups and of
different groups, respectively. If we generate networks from this
ensemble, use the BP equations to compute the posterior marginal
distribution of Eq.~\ref{eq:bp_marginal} and compare its maximum values
with the planted partition, we observe, as shown in Fig.~\ref{fig:bp},
that it is recoverable only up to a certain value of
$\epsilon=N(\lambda_{\text{in}}-\lambda_{\text{out}})$, above which the
posterior distribution is fully uniform. By inspecting the stability of
the fully uniform solution of the BP equations, the exact threshold can
be determined~\cite{decelle_asymptotic_2011},\index{topic}{stochastic
blockmodel!detectability-indetectability transition}
\begin{equation}\label{eq:bp_thresh}
  \epsilon^* = B\sqrt{\avg{k}},
\end{equation}
where $\avg{k}=N\sum_{rs}\lambda_{rs}/B^2$ is the average degree of the
network. The existence of this threshold is remarkable, because the
ensemble is only equivalent to a completely random one if $\epsilon=0$;
yet there is a non-negligible range of values
$\epsilon\in[0,\epsilon^*]$ for which \emph{the planted structure cannot
be recovered even though the model is not random}. This might seem
counter-intuitive, if we argue that making $N$ sufficiently large should
at some point give us enough data to infer the model with arbitrary
precision. The hole in logic lies in the fact that the number of
parameters --- the node partition $\bb$ --- also grows with $N$, and
that we would need the effective sample size, i.e. the number of edges
$E$, to grow \emph{faster} than $N$ to guarantee that the data is
sufficient. Since for sparse graphs we have $E=O(N)$, we are never able
to reach the limit of sufficient data. Thus, we should be able to
achieve asymptotically perfect inference only for dense graphs
(e.g. with $E=O(N^2)$) or by inferring simultaneously from many graphs
independently sampled from the same model. Neither situation, however,
is representative of what we typically encounter when we study networks.

The above result carries important implications into the overall field
of network clustering. The existence of the ``detectable'' phase for
$\epsilon>\epsilon^*$ means that, in this regime, it is possible for
algorithms to discover the planted partition in polynomial time, with
the BP algorithm doing so optimally. Furthermore, for $B>4$ (or $B>3$
for the dissortative case with $\lambda_{\text{in}} <
\lambda_{\text{out}}$) there is another regime in a range
$\epsilon^*<\epsilon<\epsilon^\dagger$, where BP converges to the
planted partition only if the messages are initialized close enough to
the corresponding fixed point. In this regime, the posterior landscape
exhibits a ``glassy'' structure, with exponentially many maxima that are
almost as likely as the planted partition, but are completely
uncorrelated with it. The problem of finding the planted partition in
this case is possible, but conjectured to be NP-hard.

Many systematic comparisons of different community detection algorithms
were done in a manner that was oblivious to these fundamental facts
regarding detectability and
hardness~\cite{lancichinetti_benchmark_2008,lancichinetti_benchmarks_2009},
even though their existence had been conjectured
before~\cite{reichardt_detectable_2008,ronhovde_multiresolution_2009},
and hence should be re-framed with it in mind. Furthermore, we point out
that although the analysis based on the BP equations is mature and
widely accepted in statistical physics, they are not completely rigorous
from a mathematical point of view. Because of this, the result of
Decelle et al~\cite{decelle_asymptotic_2011} leading to the threshold of
Eq.~\ref{eq:bp_thresh} has initiated intense activity from
mathematicians in search of rigorous proofs, which have subsequently
been found for a variety of relaxations of the original statement (see
Ref.~\cite{abbe_community_2017} for a review), and remains an active
area of research.

\section{Conclusion}

In this chapter we gave a description of the basic variants of the
stochastic blockmodel (SBM), and a consistent Bayesian formulation that
allows us to infer them from data. The focus has been on developing a
framework to extract the large-scale structure of networks while
avoiding both overfitting (mistaking randomness for structure) and
underfitting (mistaking structure for randomness), and doing so in a
manner that is analytically tractable and computationally efficient.

The Bayesian inference approach provides a methodologically correct
answer to the very central question in network analysis of whether
patterns of large-scale structure can in fact be supported by
statistical evidence. Besides this practical aspect, it also opens a
window into the fundamental limits of network analysis itself, giving us
a theoretical underpinning we can use to understand more about the
nature of network systems.

Although the methods described here go a long way into allowing us to
understand the structure of networks, some important open problems
remain. From a modeling perspective, we know that for most systems the
SBM is quite simplistic, and falls very short of giving us a mechanistic
explanation for them. We can interpret the SBM as being to network
data what a histogram is to spatial data~\cite{olhede_network_2014}, and
thus while it fulfills the formal requirements of being a generative
model, it will never deplete the modeling requirements of any particular
real system. Although it is naive to expect to achieve such a level of
success with a general model like the SBM, it is yet still unclear how
far we can go. For example, it remains to be seen how tractable it is to
incorporate local structures
--- like densities of subgraphs --- together with the large-scale
structure that the SBM prescribes.

From a methodological perspective, although we can select between the
various SBM flavors given the statistical evidence available, we still
lack good methods to assess the quality of fit of the SBM at an absolute
level. In particular, we do not yet have a systematic understanding of
how well the SBM is able to reproduce properties of empirical systems,
and what would be the most important sources of deficiencies, and how
these could be overcome.

In addition to these outstanding challenges, there are areas of
development that are more likely to undergo continuous
progress. Generalizations and extensions of the SBM to cover specific
cases are essentially open ended, such as the case of dynamic networks,
and we can perhaps expect more realistic models to appear. Furthermore,
since the inference of the SBM is in general a NP-hard problem, and thus
most probably lacks a general solution, the search for more efficient
algorithmic strategies that work in particular cases is also a long term
goal that is likely to attract further attention.


\bibliography{chapter}

\end{document}